\documentclass[sigconf]{acmart}

\usepackage{amsmath}
\usepackage{booktabs}
\usepackage{graphicx}
\usepackage{microtype}
\usepackage{multirow}
\usepackage{xspace}
\usepackage{algorithm}
\usepackage{algpseudocode}
\copyrightyear{2026}
\acmYear{2026}
\setcopyright{none}
\renewcommand\footnotetextcopyrightpermission[1]{}
\acmConference[ICVGIP 2026]{Indian Conference on Computer Vision, Graphics, and Image Processing}
{December 21--24, 2026}{Kolkata, India}
\acmBooktitle{Indian Conference on Computer Vision, Graphics, and Image Processing (ICVGIP 2026),
December 21--24, 2026, Kolkata, India}
\acmPrice{}
\acmDOI{xx.xxxx/xxxxxxx.xxxxxxx}
\acmISBN{xxx-x-xxxx-xxxx-x/2026/12}
\newcommand{\method}{\textsc{CleanGS}\xspace}
\newcommand{\gs}{3DGS\xspace}
\newcommand{\fitcol}[1]{\resizebox{\ifdim\width>\columnwidth\columnwidth\else\width\fi}{!}{#1}}

\newcommand{\qhead}{%
  & \scriptsize GT & \scriptsize 3DGS & \scriptsize GCut & \scriptsize Ours
  & \scriptsize GT 6$\times$ & \scriptsize 3DGS 6$\times$
  & \scriptsize GCut 6$\times$ & \scriptsize Ours 6$\times$ \\}

\begin{document}

\title{\textsc{CleanGS}: Pareto-Guided Post-Hoc Cleanup of
3D Gaussian Splatting Scenes}

\author{Jyotirmaya Shivottam}
\orcid{0000-0002-2688-9421}
\affiliation{%
  \institution{National Institute of Science Education and Research}
  \city{Bhubaneswar}
  \postcode{752050}
  \state{Odisha}
  \country{India}}
\affiliation{%
  \institution{Homi Bhabha National Institute}
  \city{Mumbai}
  \postcode{400094}
  \country{India}}
\email{jyotirmaya.shivottam@niser.ac.in}

\author{Om Kumar}
\orcid{0009-0001-6754-995X}
\affiliation{%
  \institution{Manipal Institute of Technology}
  \city{Manipal}
  \state{Karnataka}
  \country{India}}
\email{om.mitmpl2023@learner.manipal.edu}

\author{Subhankar Mishra}
\orcid{0000-0002-9910-7291}
\affiliation{%
  \institution{National Institute of Science Education and Research}
  \city{Bhubaneswar}
  \postcode{752050}
  \state{Odisha}
  \country{India}}
\affiliation{%
  \institution{Homi Bhabha National Institute}
  \city{Mumbai}
  \postcode{400094}
  \country{India}}
\email{smishra@niser.ac.in}

\renewcommand{\shortauthors}{Shivottam et al.}

\begin{abstract}
  A pretrained 3D Gaussian Splatting (\gs) reconstruction can reproduce its
capture views well while retaining off-object structure that becomes visible
under reframing. We present \method, a post-hoc cleanup method that removes
such structure without retraining the underlying representation. Given
calibrated views and binary target masks, \method scores candidate three-view
subsets using mask reliability, camera baseline, viewing-direction diversity,
and complementary sparse 3D support. Rather than collapsing these objectives
a priori, it computes the exact Pareto front over subsets of the reduced
candidate pool and selects a balanced non-dominated subset using weighted
Chebyshev scalarization. The selected
views then drive rasterization-aware two-of-three voting to retain a subset of
the original Gaussians. We evaluate the exported asset directly and introduce
a fixed-pose widened-field protocol that exposes structure outside the original
framing. Across 18 scenes, \method achieves $19.03\,\textrm{dB}$ standard and
$17.83\,\textrm{dB}$ widened-field PSNR, compared with
$16.44/16.29\,\textrm{dB}$ for FlashSplat and
$15.28/15.10\,\textrm{dB}$ for GaussianCut with the same three masks. \method leads five of six photometric metrics and remains within
$0.32\,\textrm{dB}$ of the uncleaned reconstruction in standard PSNR. Its
filtering-and-export stage takes a median of 20 seconds and
$1.9\,\textrm{GB}$ peak memory, compared with 3.1 minutes and
$41.6\,\textrm{GB}$ for FlashSplat and 4.2 minutes and
$34.9\,\textrm{GB}$ for GaussianCut. Code is available at
\url{https://github.com/smlab-niser/clean-gs/}.
\end{abstract}


\begin{teaserfigure}
  \centering
  \includegraphics[width=0.9\textwidth,height=0.43965\textwidth]{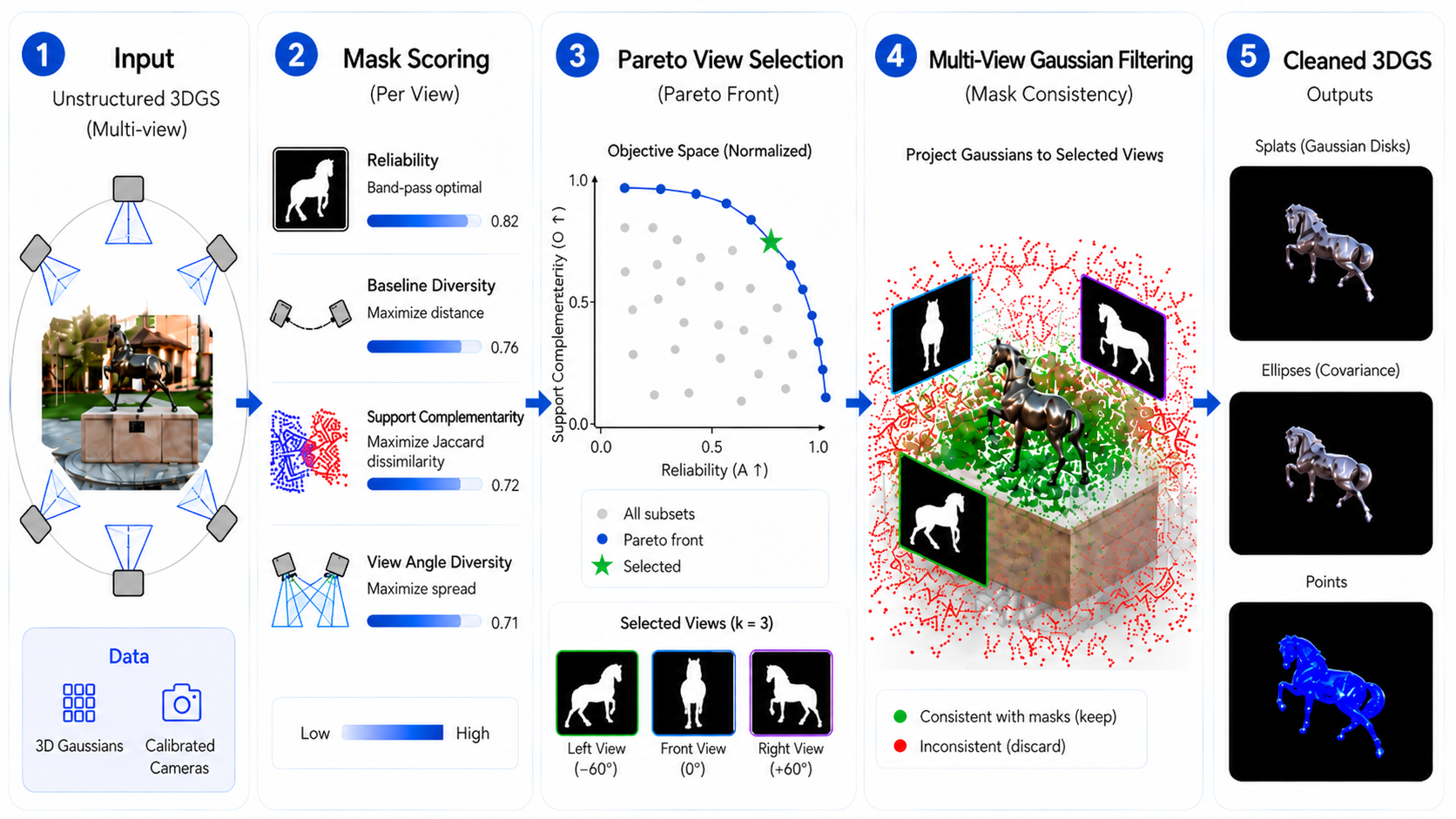}
  \caption{\textbf{Overview of \method.}
  Candidate masks are scored for reliability, and three-view subsets for camera
  baseline, viewing-angle diversity, and complementary masked 3D support. We
  compute the exact Pareto front over the reduced candidate pool, select one
  balanced subset from it, then apply rasterization-aware two-of-three voting.
  Retained Gaussians are exported unchanged as a subset of the input scene.}
  \Description{Five stage cards. Stage one is a pretrained Gaussian scene with
  calibrated cameras. Stage two scores each view for reliability, baseline
  diversity, support diversity and viewing-angle diversity. Stage three plots
  all subsets in objective space, highlights the Pareto front and marks the
  selected subset, with the three selected masks below. Stage four projects
  Gaussians into the selected views and marks consistent ones as kept and
  inconsistent ones as discarded. Stage five shows the cleaned asset rendered
  as splats, as covariance ellipses and as points.}
  \label{fig:overview}
\end{teaserfigure}

\maketitle

\begin{figure*}[t]
  \centering
  \setlength{\tabcolsep}{2pt}
  \begin{tabular}{@{}ccccc@{}}
    \scriptsize GT & \scriptsize Uncleaned \gs & \scriptsize FlashSplat & \scriptsize GaussianCut$^\dag$ & \scriptsize Ours \\
    \includegraphics[width=.19\textwidth]{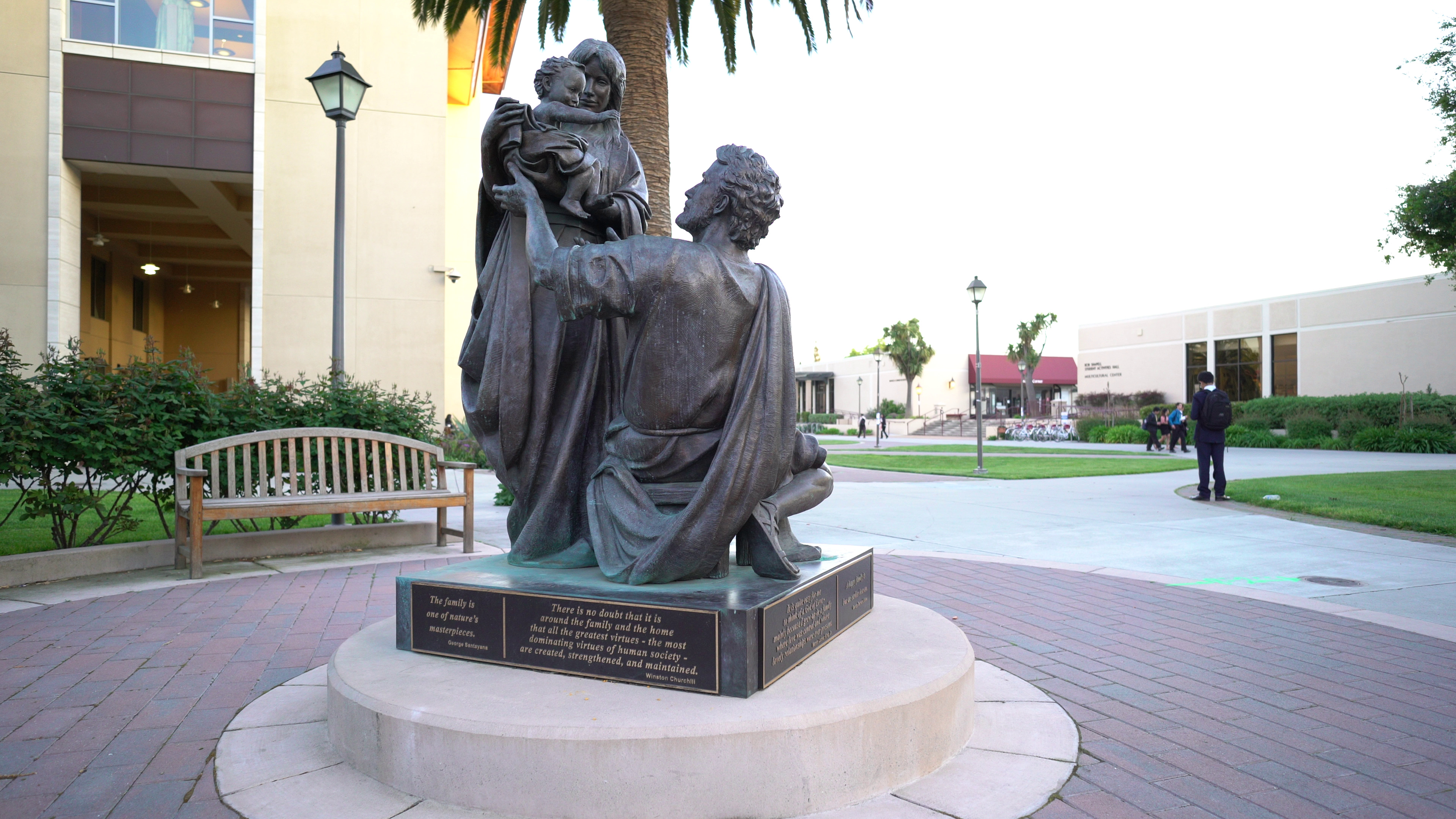} &
    \includegraphics[width=.19\textwidth]{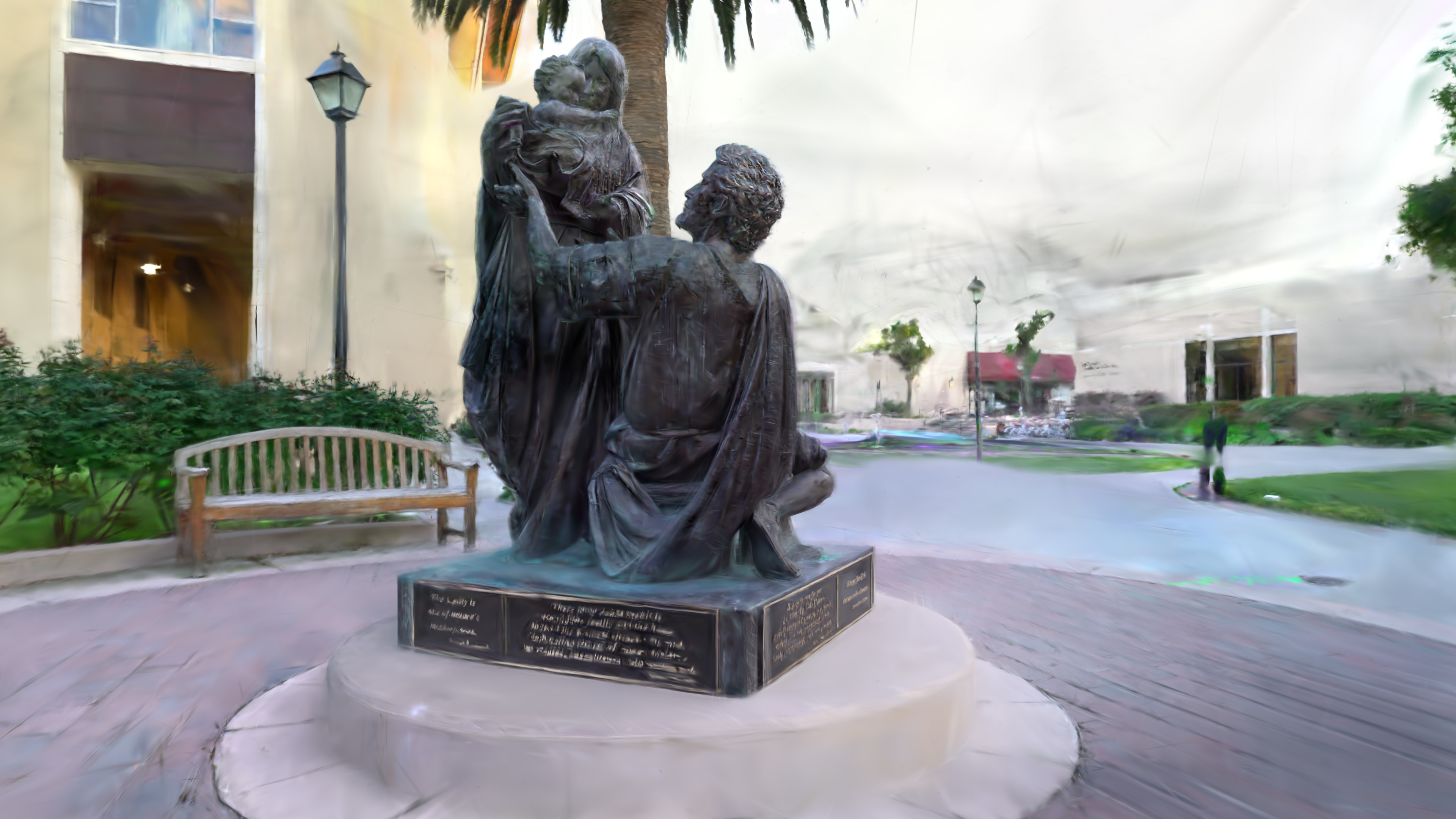} &
    \includegraphics[width=.19\textwidth]{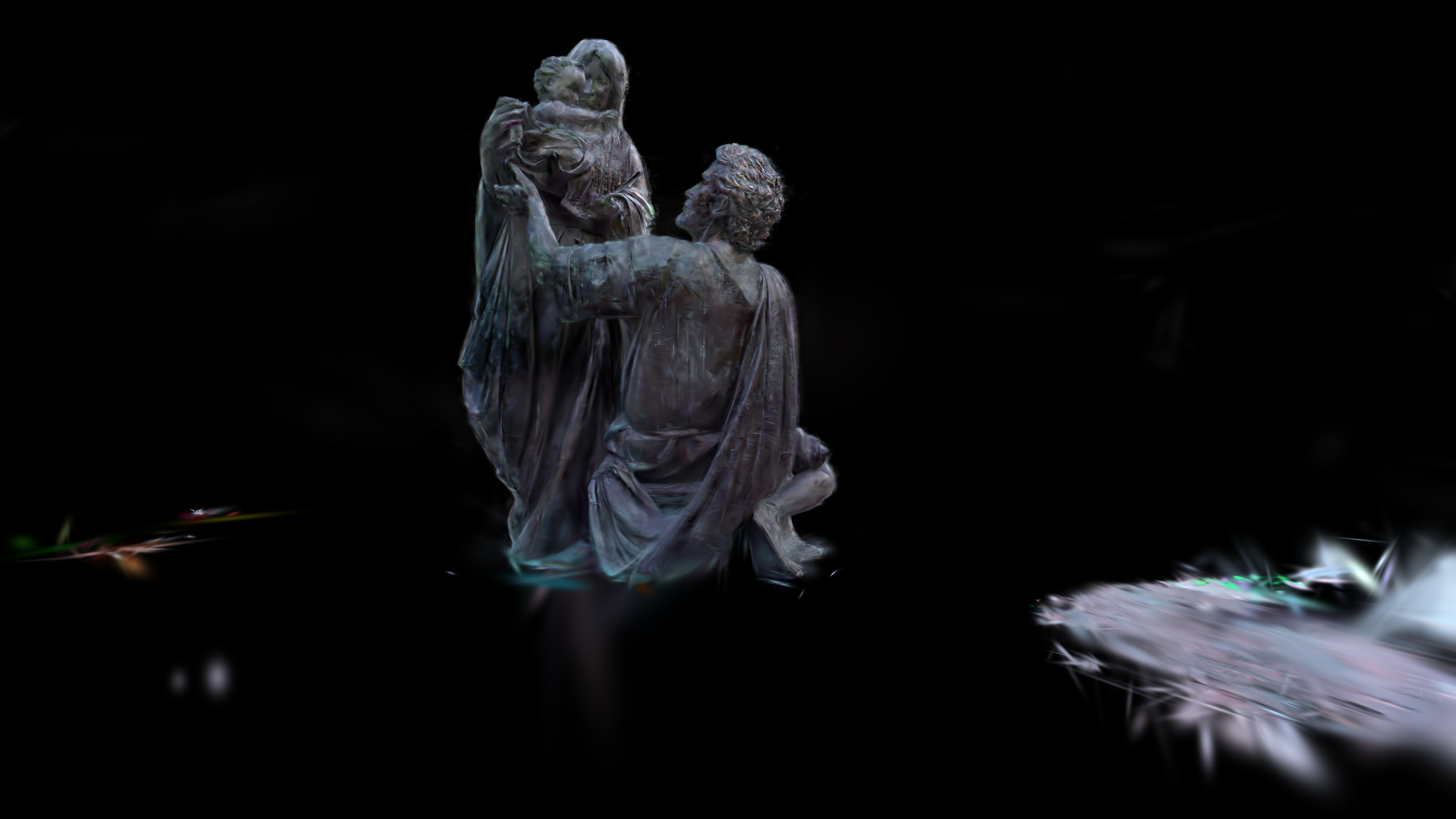} &
    \includegraphics[width=.19\textwidth]{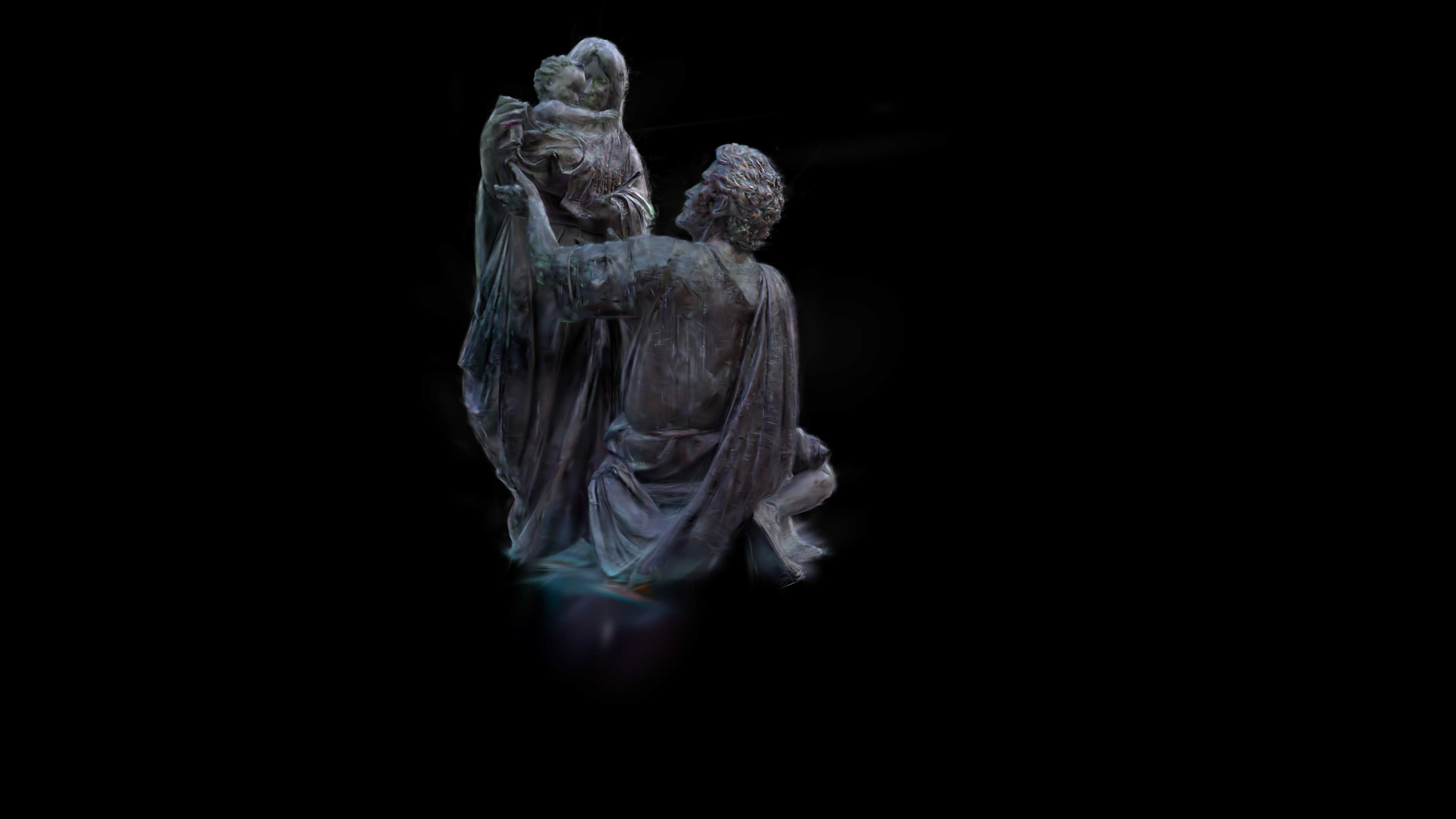} &
    \includegraphics[width=.19\textwidth]{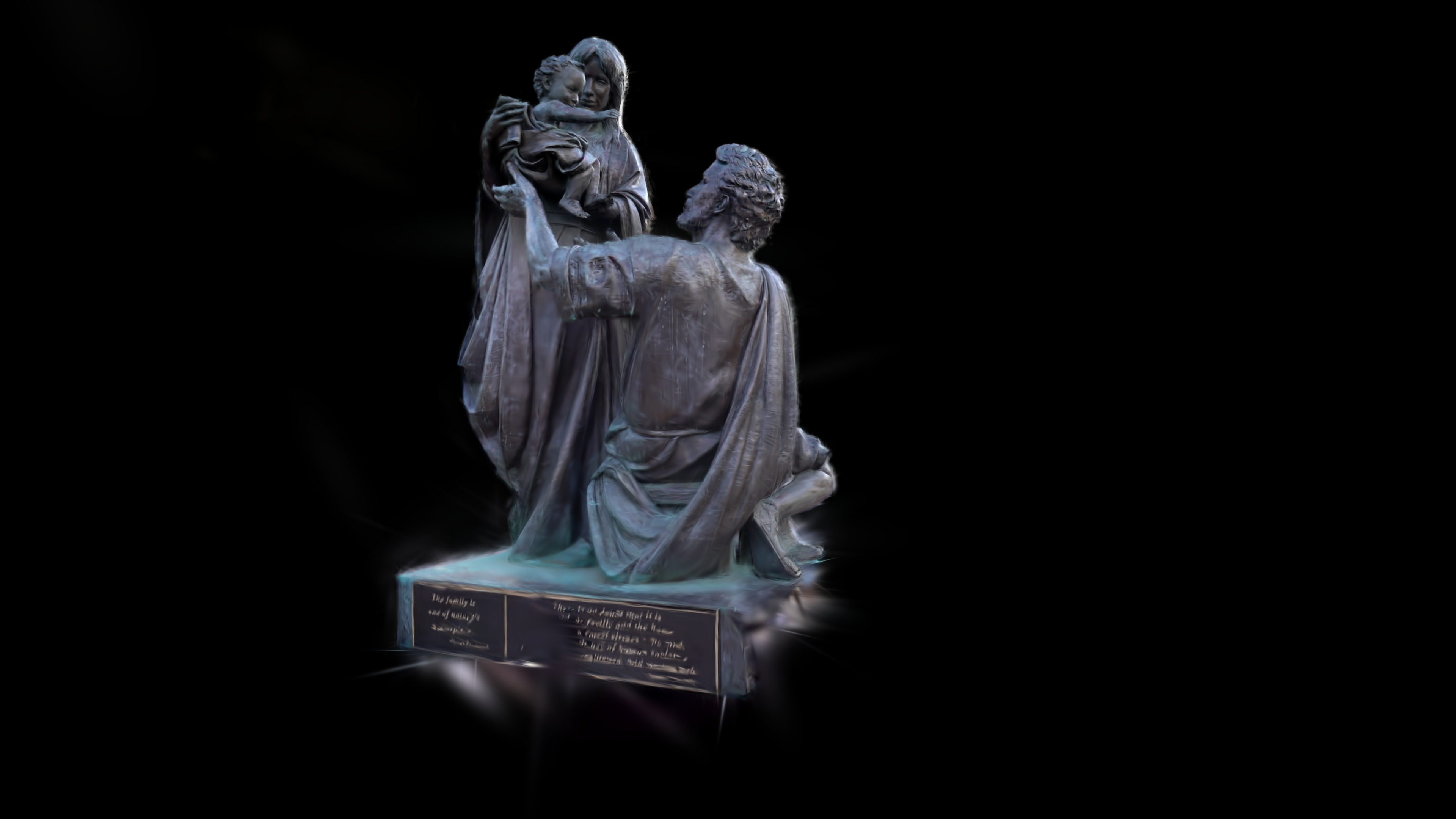} \\
    \includegraphics[width=.19\textwidth]{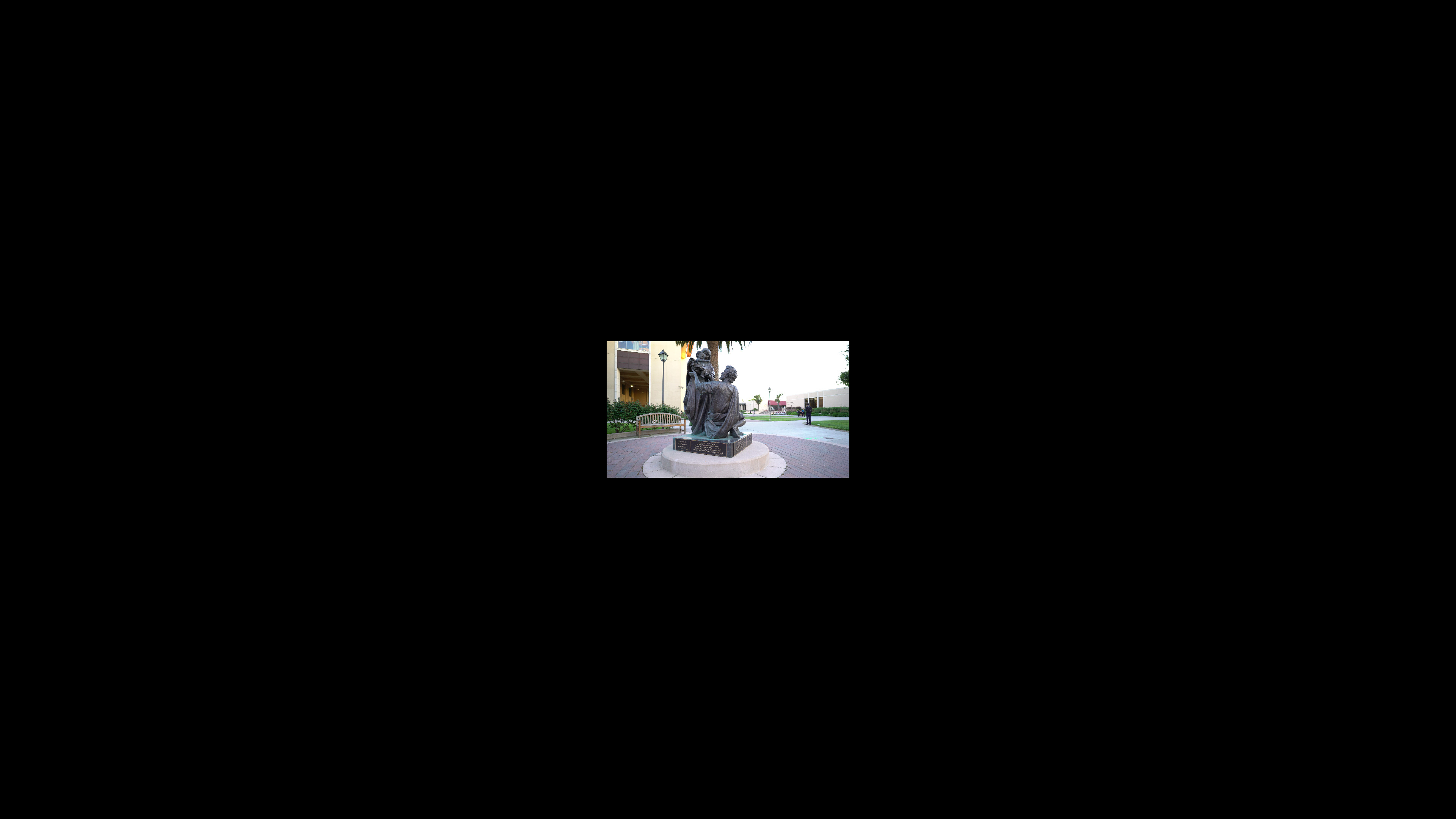} &
    \includegraphics[width=.19\textwidth]{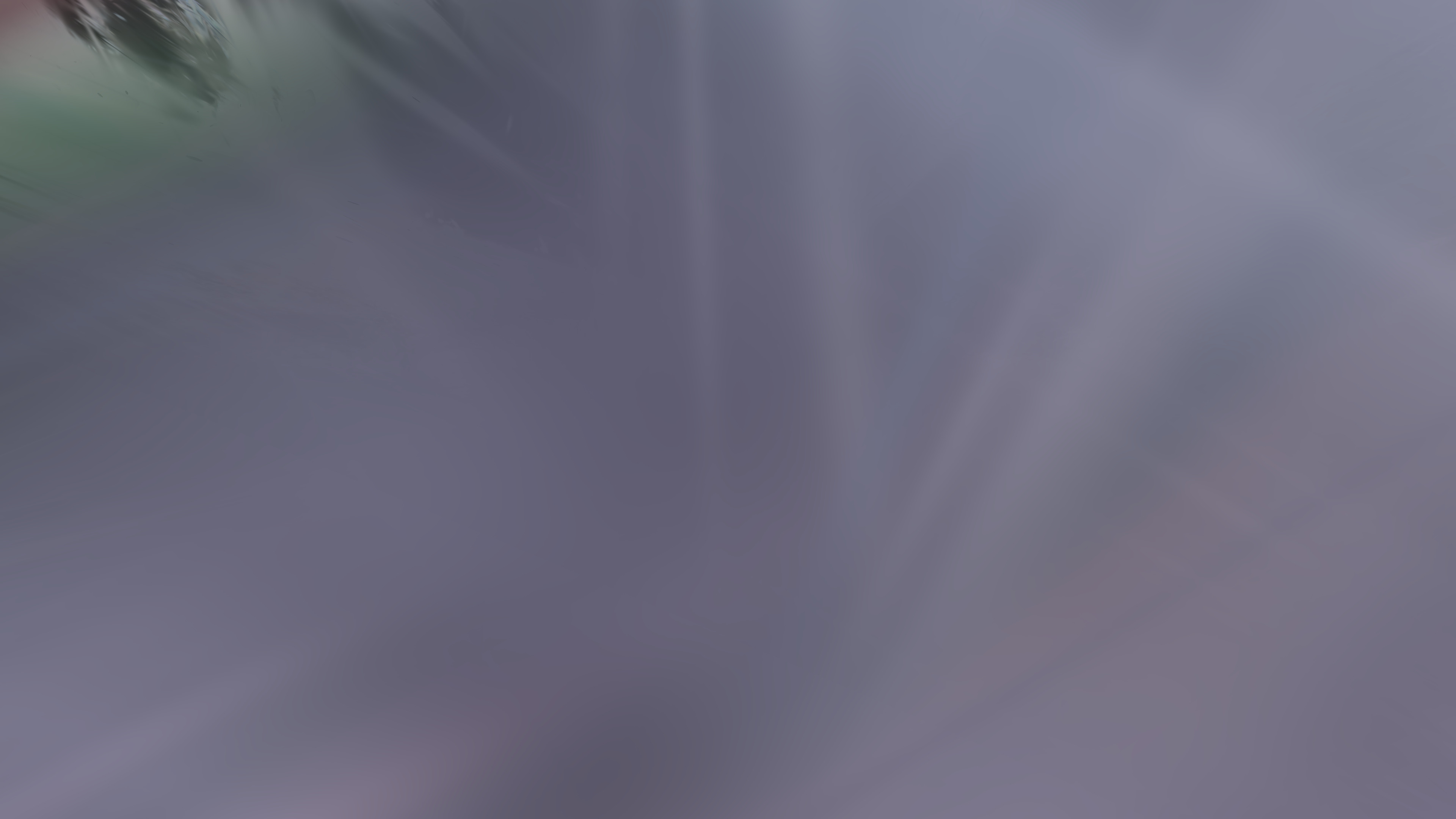} &
    \includegraphics[width=.19\textwidth]{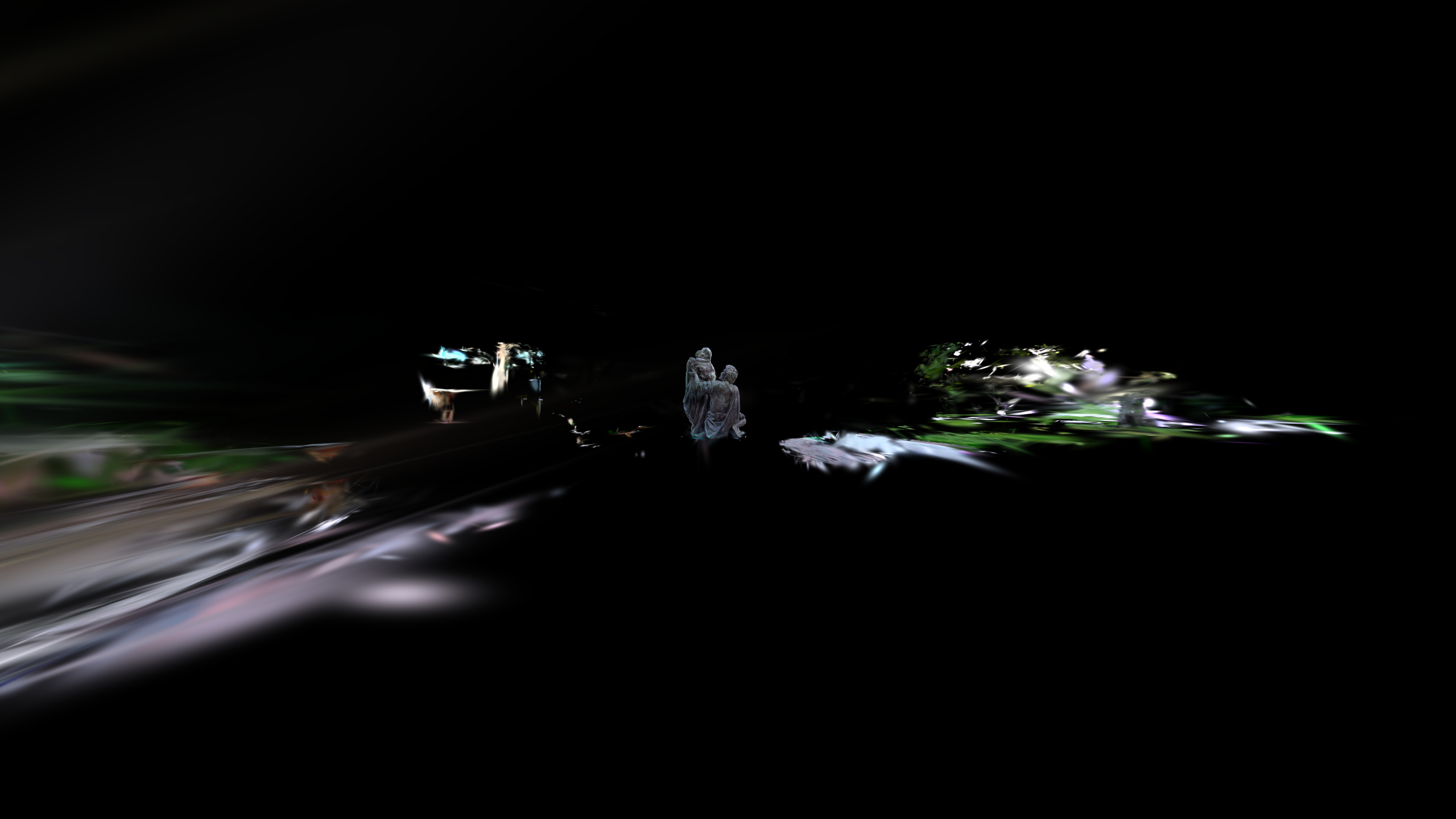} &
    \includegraphics[width=.19\textwidth]{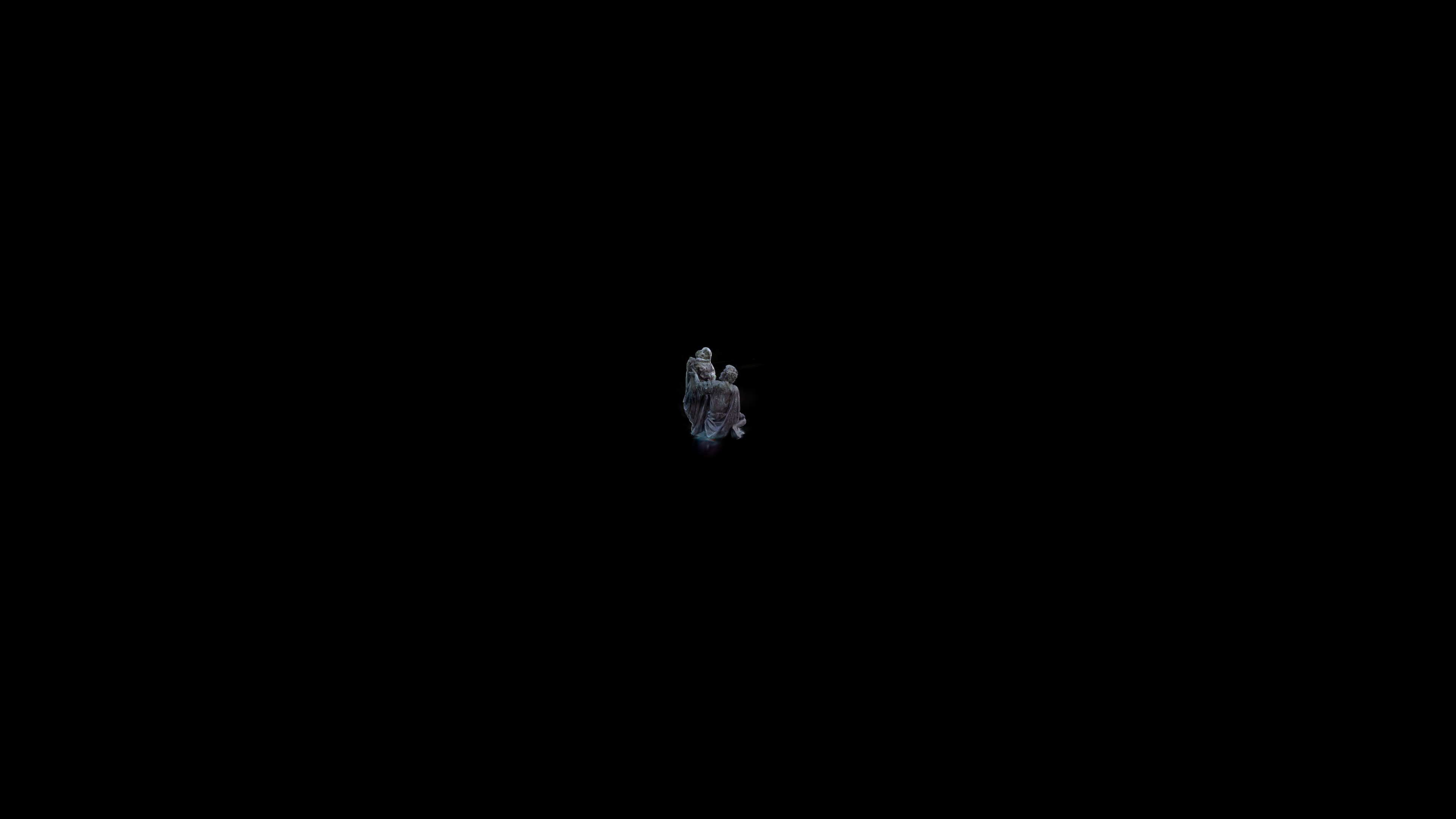} &
    \includegraphics[width=.19\textwidth]{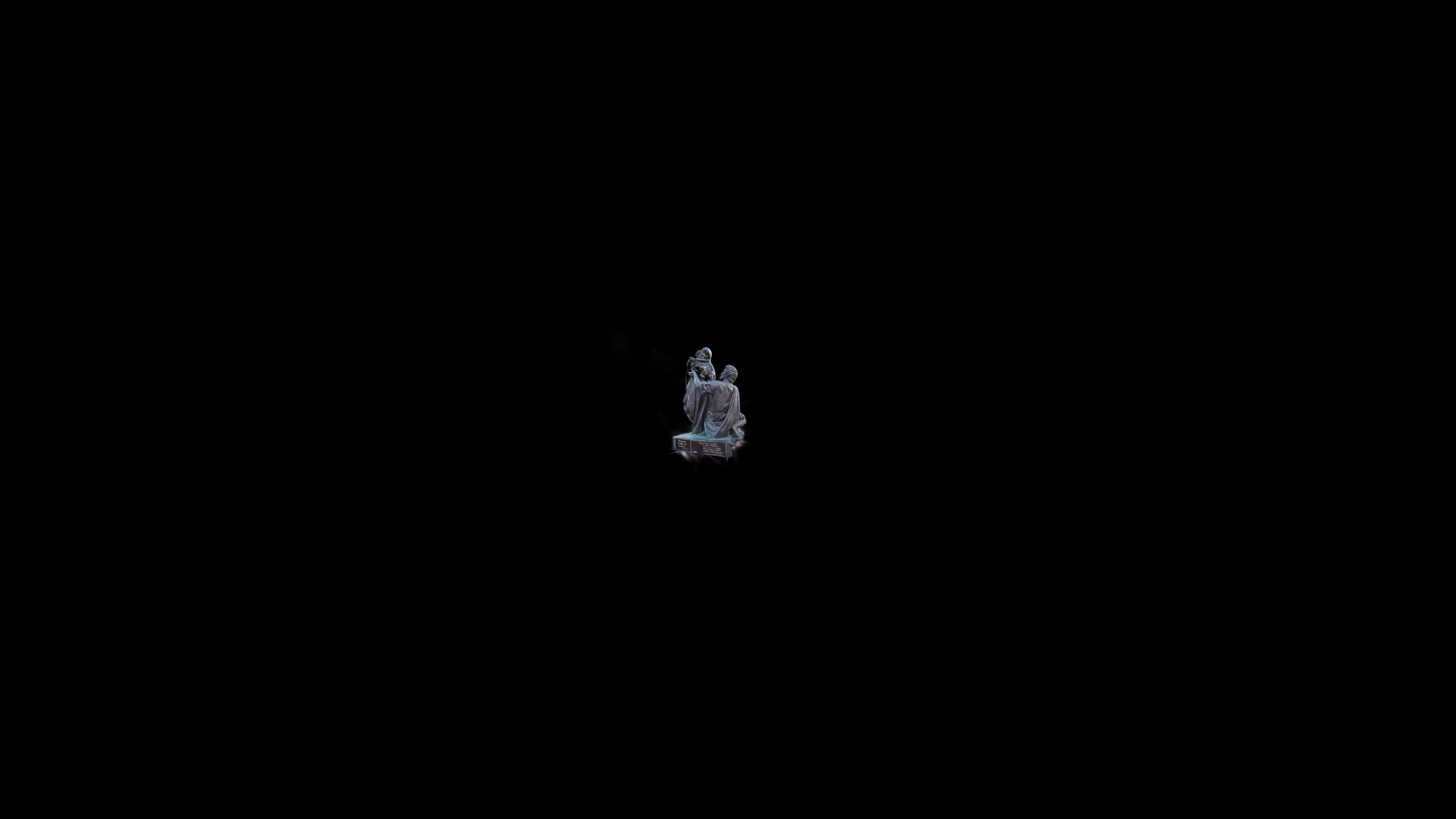}
  \end{tabular}
  \caption{\textbf{Ordinary framing hides retained clutter.}
  Top: captured view. Bottom: $6\times$ wider field of view at the same pose.
  The uncleaned asset reveals substantial off-object structure, while
  FlashSplat and GaussianCut remove the monument's plinth. \method preserves
  the plinth while remaining clean after widening. GaussianCut$^\dag$ denotes
  the all-mask reference.}
  \Description{Captured and widened renderings of a statue. Widening reveals
  off-object structure in the uncleaned asset; CleanGS preserves the plinth
  removed by the baselines.}
  \label{fig:framing}
\end{figure*}

\section{Introduction}

3D Gaussian Splatting (\gs) represents a scene as explicit anisotropic Gaussian
primitives that support efficient, high-quality novel-view synthesis~\cite{3dgs}.
Because individual primitives are directly accessible for editing and
removal~\cite{gsedit}, a trained \gs scene is also attractive as a reusable 3D
asset. However, accurate reconstruction of captured views does not necessarily
produce a clean object representation. Photometric optimization can retain
primitives outside the object of interest, producing translucent sheets,
disconnected fragments, duplicated surfaces, or background structure. We refer
to this unwanted retained structure as \emph{off-object clutter}.

Training-time regularization can reduce such artifacts when reconstruction
itself can be modified~\cite{tidigs,stabgs,efags}. We instead consider a
pretrained scene that must be cleaned post hoc. Binary object masks provide a
natural semantic specification of what should remain, but lifting them to
Gaussians is ambiguous: a primitive projected inside a target silhouette may
belong either to the target or to background along the same viewing ray.
Reliable cleanup therefore requires multiview evidence.

The quality of this evidence depends strongly on the chosen views. Some masks
provide clear, well-contained observations, while others are truncated,
close-up, weakly covering, or geometrically redundant. A small set of reliable
and complementary views can therefore provide stronger evidence than treating
all observations equally. This motivates our central question:
\textbf{which views should provide the evidence used to decide which Gaussians
belong to the target?}

Our \method couples view selection with Gaussian filtering, as summarized in
Figure~\ref{fig:overview}. Given calibrated views and target masks, it scores
three-view subsets using mask reliability, camera baseline, viewing-direction
diversity, and complementary sparse 3D support. After reducing the candidate
pool, we compute the exact Pareto front over these objectives and select a
balanced non-dominated subset using a weighted Chebyshev criterion. The selected masks then drive rasterization-aware
two-of-three voting, and the surviving Gaussians are exported without modifying
their geometry, opacity, or appearance. The three selected views define the
filtering budget; the selector may examine a larger pool of candidate masks.

This formulation makes the quality of multiview evidence explicit. Three views
provide redundancy for a two-of-three decision, while the selection objectives
discourage repeated semantic and geometric evidence. Holding the filtering rule
fixed, Pareto-selected views improve both standard and widened-field
photometric fidelity over random three-view subsets.

Evaluation presents a second challenge: standard capture views measure target
preservation but may hide retained structure outside the original framing. We
therefore evaluate the exported Gaussian asset directly and introduce a
\emph{fixed-pose widened-field evaluation}. The focal length is reduced while
camera pose and image resolution remain fixed, and the same image-plane
transformation is applied to the reference image and target mask. This exposes
previously out-of-frame structure without requiring ground-truth images from
arbitrary novel poses. We report masked PSNR, SSIM, and LPIPS together with
silhouette leakage and Gaussian retention.

Across 18 scenes, \method achieves $19.03\,\textrm{dB}$ PSNR at the captured
framing and $17.83\,\textrm{dB}$ after widening. At the same three-mask
filtering budget, FlashSplat~\cite{flashsplat} reaches $16.44$ and
$16.29\,\textrm{dB}$, while GaussianCut reaches $15.28$ and
$15.10\,\textrm{dB}$. An all-mask GaussianCut$^\dag$ reference reaches $17.27$
and $16.71\,\textrm{dB}$. Overall, \method obtains the best result on five of
six reported photometric metrics, showing that complementary view selection can
preserve target appearance while removing off-object structure without
retraining. Its filtering-and-export stage takes 20 seconds (41 seconds
end to end, including view selection) and
$1.9\,\textrm{GB}$ peak memory, against 3.1 and 4.2 minutes and $41.6$ and
$34.9\,\textrm{GB}$ for FlashSplat and GaussianCut.

Our contributions are:
\begin{enumerate}
\item a retraining-free cleanup pipeline combining Pareto-guided view selection
with rasterization-aware multiview filtering to export a reusable subset of the
original Gaussian scene;

\item a Pareto-guided multi-objective view-selection formulation that exposes
the trade-offs among mask reliability, camera baseline, viewing-direction
diversity, and complementary sparse 3D support before selecting three views;

\item a retained-asset evaluation protocol measuring target fidelity at both
captured and widened framing together with silhouette leakage and Gaussian
retention; and

\item an 18-scene evaluation showing improved photometric fidelity over
retained-asset baselines and random view selection, with analyses of view
budget, voting thresholds, and geometric refinement.
\end{enumerate}

\section{Related Work}

\paragraph{Artifact suppression and pruning.}
Floating and transient structures can arise when photometric optimization explains the capture images using geometry that is poorly constrained away from the observed views. Recent methods address this problem during reconstruction by introducing visibility, depth, frequency, or robustness priors \cite{tidigs,stabgs,efags}. These approaches are complementary to our setting, where the Gaussian scene is already trained and cleanup is performed post hoc. Related work on \gs compression, including LightGaussian and Compact3DGS, removes primitives according to rate--distortion objectives \cite{lg,c3dgs}. Such criteria are effective for reducing representation size, but global reconstruction importance does not by itself distinguish the object of interest from surrounding structure that remains useful for reproducing the captured views. Our setting therefore combines post-hoc pruning with an explicit semantic target.

\paragraph{Mask lifting and object extraction.}
Recent surveys organize Gaussian scene understanding methods into feature distillation, 2D mask lifting, and feed-forward prediction \cite{3dgs_survey,gsedit}. Feature-distillation approaches embed semantic information into Gaussian attributes to support tasks such as open-vocabulary querying \cite{f3dgs,langsplatv2}, while Gaussian Grouping learns per-Gaussian identity features for in\-stance-level interaction \cite{gg}. A second family directly lifts multiview 2D masks to the Gaussian representation \cite{sa3dgs,clickgaussian,unifiedlift}, and \method is most closely related to this line of work.

FlashSplat formulates mask-to-Gaussian assignment as a global linear program \cite{flashsplat}, while Lifting by Gaussians and LUDVIG study training-free mechanisms for lifting and consolidating 2D evidence in Gaussian scenes \cite{lbg,ludvig}. GaussianCut constructs an adjacency graph over Gaussians, converts multiview mask evidence into unary terms, and obtains the object through graph-cut optimization \cite{gaussiancut}. Our focus is specifically on producing a cleaned object asset whose retained primitives can be exported and rendered directly. For this reason, we evaluate the resulting Gaussian subset rather than a method-specific semantic or label rendering. GaussianCut and FlashSplat provide our primary retained-asset comparisons because both ultimately produce a pruned Gaussian set that can be evaluated under the same rendering protocol as \method. In contrast to applying the lifting procedure to a fixed collection of observations, \method first selects a small set of reliable and complementary views and then uses only their evidence for multiview filtering.

\paragraph{View selection.}
View selection is well established in multiview geometry, where camera baseline and viewing angle influence triangulation conditioning, and geometrically diverse observations often provide more useful constraints than highly redundant ones \cite{hartley2004multiple,schoenberger2016mvs}. Our setting differs in that the cameras are already calibrated and each candidate view additionally provides a semantic mask and sparse Structure-from-Motion support \cite{schoenberger2016sfm}. We therefore treat view selection as a multi-objective problem that considers both geometric diversity and the quality of the semantic evidence available in each view.

Rather than combining these heterogeneous criteria into a single score at the outset, we evaluate candidate subsets using separate objectives and retain their non-dominated solutions \cite{miettinen1999nonlinear,deb2001multiobjective}. This formulation makes the trade-offs between semantic reliability, camera separation, viewing direction, and complementary masked 3D support explicit before selecting the subset used for cleanup.

\section{Method}
\label{sec:method}

\subsection{Problem Definition}

Let $\mathcal{G}=\{g_i\}_{i=1}^{N}$ be a pretrained \gs scene, where each
Gaussian $g_i$ contains position, rotation, scale, opacity, and
spherical-harmonic appearance parameters. The candidate view set
$\mathcal{V}$ contains calibrated cameras and binary target masks
$\{M_v\}_{v\in\mathcal{V}}$. Camera $v$ has center $p_v$ and unit viewing
direction $d_v$. Our goal is to select a subset
$S\subseteq\mathcal{V}$ with $|S|=k=3$ and use its mask evidence to construct
a keep-set $\mathcal{K}\subseteq\{1,\ldots,N\}$. The cleaned asset is obtained
by exporting only the Gaussians indexed by $\mathcal{K}$.
Figure~\ref{fig:overview} summarizes the complete pipeline.

We intentionally use a small view budget. Although two calibrated views are
sufficient to triangulate visible structure when their viewing rays are not
parallel~\cite{hartley2004multiple,longuet1981computer}, semantic cleanup
poses a different problem. A mask provides uncertain evidence about whether a
projected Gaussian belongs to the target, and a two-view decision has little
redundancy against an erroneous or incomplete mask. We therefore use three
views with a two-of-three voting rule. This introduces one redundant
observation, allowing a Gaussian to remain supported when one selected mask
fails to capture it correctly.

The choice $k=3$ also provides a favorable balance between filtering quality
and exact subset selection. Proposition~\ref{prop:binomial} shows that, under
conditional independence, two-of-three voting suppresses background support
quadratically in the per-view false-support probability. This motivates
selecting views whose evidence is as complementary as possible. The appendix
evaluates view budget and vote threshold jointly. The unanimous $k=2$
configuration is substantially weaker than the main two-of-three setting, while
the strongest larger-budget setting, three-of-five voting, matches standard
PSNR over all 18 scenes while improving widened PSNR by $0.22\,\textrm{dB}$
and lowering leakage from $0.206$ to $0.180$.

\subsection{Candidate Reliability}

We first estimate how suitable each candidate view is for semantic lifting.
For view $v$, let the foreground occupancy be
\begin{equation}
  r_v=\frac{|M_v|}{|\Omega_v|}.
\end{equation}
View quality is not monotonic in occupancy: very small masks provide weak
target support, while near-full-frame masks often correspond to close or
truncated views. We therefore use the band-pass reliability score
\begin{equation}
  a_v=\exp\!\left[-\frac{1}{2}
  \left(\frac{r_v-r_0}{\sigma}\right)^2\right],
\end{equation}
with $r_0=0.30$ and $\sigma=0.15$ for all scenes, favoring intermediate target
occupancy. Sensitivity to the occupancy target and to the shape of the
reliability score is evaluated in the appendix.

We further reject poorly contained masks using occupancy, border-contact, and
bounding-box tests. When fewer than three candidates survive the subsequent
reduction, we fall back to retaining all eligible masks; if selection still
fails, we disable the containment tests and cap the pool deterministically at
60 views. The exact thresholds and the capping rule are given in the
appendix.

Each candidate is also associated with sparse 3D support. Let $Q_v$ denote the
SfM tracks whose observation in view $v$ lies inside $M_v$; these provide an
inexpensive approximation to the reconstructed support covered by the mask.
Before subset enumeration, we remove candidates dominated in reliability,
border margin, and truncation, and write $\mathcal{V}_r\subseteq\mathcal{V}$
for the remaining pool. This reduces clearly inferior views while retaining
meaningful trade-offs for subsequent subset selection.

\subsection{Pareto-Guided Subset Selection}

After candidate reduction, we evaluate each feasible subset
$S\subseteq\mathcal{V}_r$ using both semantic quality and geometric
complementarity. For an unordered pair set
$P(S)=\{(u,v):u,v\in S,\;u<v\}$, we define
\begin{align}
  A(S) &= \min_{v\in S} a_v, \\
  B(S) &= \frac{1}{|P(S)|}\sum_{(u,v)\in P(S)}
  \lVert p_u-p_v\rVert_2, \\
  O(S) &= \frac{1}{|P(S)|}\sum_{(u,v)\in P(S)}
  \left(1-\frac{|Q_u\cap Q_v|}{|Q_u\cup Q_v|}\right), \\
  T(S) &= \frac{1}{|P(S)|}\sum_{(u,v)\in P(S)}
  \arccos(\langle d_u,d_v\rangle).
\end{align}
The four objectives capture distinct properties of a useful subset. $A(S)$
measures the reliability of its weakest view, since one poor mask can
systematically affect the subsequent vote. $B(S)$ favors separation between camera centers. $O(S)$ is a Jaccard
\emph{dissimilarity} over masked SfM tracks and is therefore maximized. It
discourages subsets whose views encode nearly identical sparse support. It carries a low weight, is
balanced against $A$ and the geometric terms, and the final two-of-three vote
still requires cross-view agreement before a Gaussian is retained. $T(S)$ favors
diversity in viewing direction, and we retain it alongside $B(S)$ because a
large camera displacement does not necessarily imply a large change in viewing
direction.

These criteria have different scales and encode different notions of
desirable evidence, so we do not combine them into a single score before
enumerating their trade-offs. We instead evaluate all feasible $k$-subsets
$S\subseteq\mathcal{V}_r$ and compute their exact Pareto front under
$f(S)=(A,B,O,T)$. A subset is dominated when another subset is at least as
good in every objective and strictly better in at least one. Dominated subsets
cannot achieve a strictly larger value under any utility that is
non-decreasing in all four objectives.

We min--max normalize the objectives and select a single point from the exact
front using weighted Chebyshev distance to the fixed ideal
$z=(1,1,1,1)$:
\begin{equation}
  S^\star = \arg\min_{S\in\mathcal{F}}
  \max_{\ell} w_\ell\bigl(z_\ell-\widetilde f_\ell(S)\bigr),
\end{equation}
where $\mathcal{F}$ denotes the exact Pareto front. Chebyshev scalarization
penalizes solutions whose weakest normalized objective remains far from the
ideal \cite{cheb}. Since the scalarization is monotone in the objectives,
Lemma~\ref{lem:dominance} guarantees that its optimal value is attained on
$\mathcal{F}$. The front also keeps the non-dominated trade-offs explicit,
which allows the same objective space to be examined under different
weightings, as we do in the appendix.

We use weights $(0.45,0.30,0.15,0.10)$ for $(A,B,O,T)$ in every scene. The appendix provides further sensitivity experiments and a discussion on weight choices. The weighting
emphasizes $A$ and $B$, which account for most selection changes at this
operating point: reliability controls the semantic evidence entering the vote,
and baseline provides a strong geometric cue against redundant observations. The lower-weight $O$ and $T$ terms make smaller but still measurable contributions: zeroing either reduces standard PSNR by $0.05$--$0.08\,\textrm{dB}$ and widened-field PSNR by $0.16$--$0.22\,\textrm{dB}$, while also increasing leakage. Their effect is more pronounced under widened framing, where redundant view selection is more likely to leave residual structure exposed. Across the tested settings, standard PSNR varies by at most $0.14\,\textrm{dB}$, while placing greater emphasis on the diversity terms can further improve the widened-field operating point.

\begin{figure}[t]
\centering
\setlength{\tabcolsep}{1pt}
\begin{tabular}{@{}ccc@{}}
\includegraphics[width=.32\columnwidth]{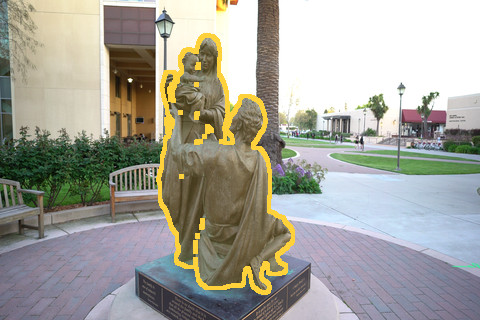} &
\includegraphics[width=.32\columnwidth]{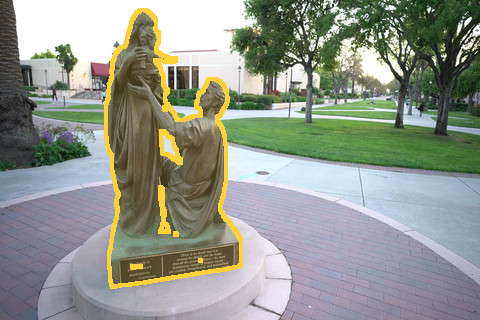} &
\includegraphics[width=.32\columnwidth]{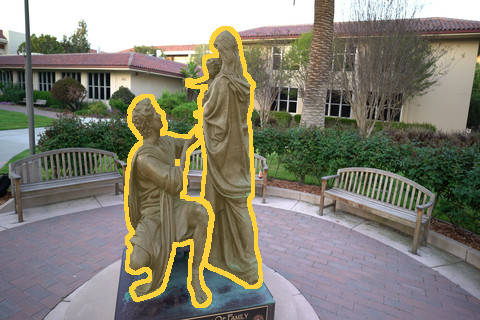} \\[1pt]
\includegraphics[width=.32\columnwidth]{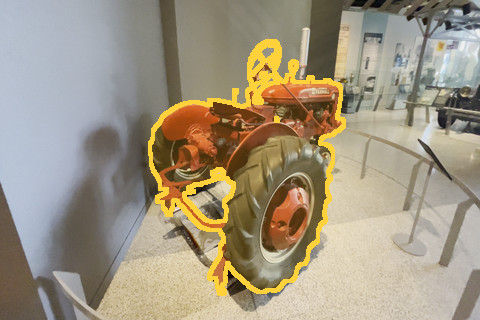} &
\includegraphics[width=.32\columnwidth]{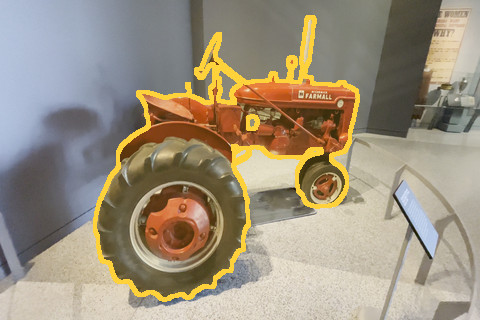} &
\includegraphics[width=.32\columnwidth]{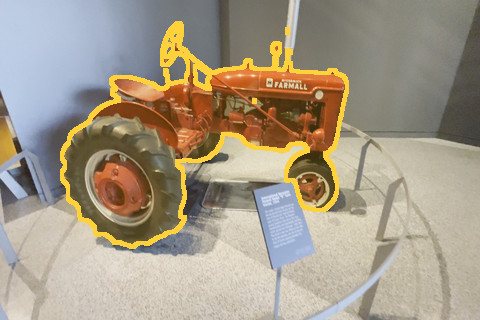} \\[1pt]
\includegraphics[width=.32\columnwidth]{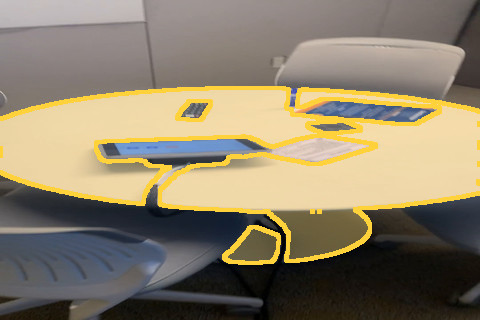} &
\includegraphics[width=.32\columnwidth]{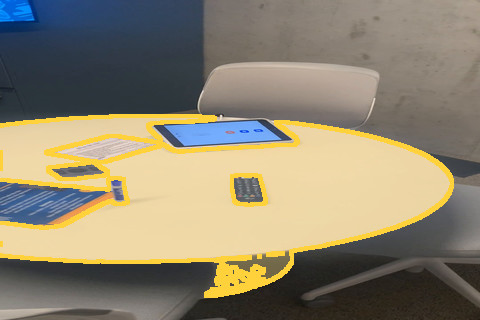} &
\includegraphics[width=.32\columnwidth]{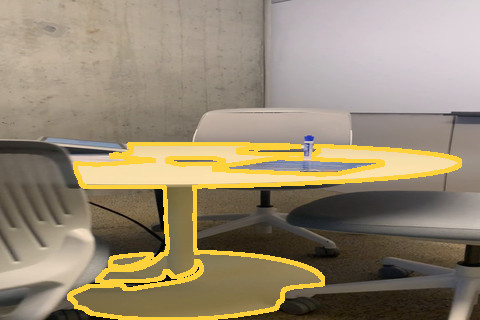}
\end{tabular}
\caption{Selected three-view subsets used for Gaussian filtering $S^\star$ for \emph{Family}, \emph{tractor}, and \emph{table}, with masks outlined. The selected views balance mask reliability with geometric and support diversity.}
\Description{Three selected masked views for each of the three scenes.}
\label{fig:selected}
\end{figure}

Figure~\ref{fig:selected} shows the selected subsets for three
representative scenes.

\begin{lemma}[Dominance is pref\-er\-ence-free]
\label{lem:dominance}
\setlength{\emergencystretch}{2em}
Let $U(S)=\Phi(f(S))$ for any $\Phi$ that is non-decreasing in each of its
four arguments. If $S_2$ dominates $S_1$, then $U(S_2)\ge U(S_1)$.
Consequently $\max_S U(S)$ is attained on $\mathcal{F}$.
\end{lemma}

The proof and a discussion of each objective are in the appendix.
The Chebyshev weights therefore only choose among mutually non-dominated
alternatives.

\subsection{Multiview Filtering and Export}

The selected subset $S^\star$ provides the mask evidence used to filter the
Gaussian scene. We rasterize $\mathcal{G}$ from each selected camera and use
renderer visibility to exclude Gaussians that cannot contribute to the image.
With projection $\pi_v$ and visibility $\operatorname{vis}_v$, we define
\begin{equation}
  X_v(g_i)=
  \mathbf{1}[\operatorname{vis}_v(g_i)]\,
  \mathbf{1}[\pi_v(\mu_i)\in M_v],
  \quad
  \nu_i=\sum_{v\in S^\star}X_v(g_i).
\end{equation}
For $k=3$, we retain $g_i$ when $\nu_i\ge2$. This two-of-three rule requires
multiview agreement while tolerating one incomplete or inconsistent mask. The benefit of voting depends on the selected views providing complementary evidence. The following proposition characterizes the
idealized case of conditionally independent support.

\begin{proposition}[Independent support suppresses clutter]
\label{prop:binomial}
Let $p_f=\Pr[X_v(g)=1\mid g\text{ foreground}]$ and
$p_b=\Pr[X_v(g)=1\mid g\text{ background}]$ with $p_f>p_b$, and suppose
the $k$ indicators $\{X_v(g)\}_{v\in S^\star}$ are conditionally i.i.d.
given the class of $g$. Under the keep rule $\nu(g)\ge\tau$,
\begin{equation}
\Pr[\nu(g)\ge\tau\mid g]
=\sum_{j=\tau}^{k}\binom{k}{j}
p^{\,j}(1-p)^{k-j},
\end{equation}
with $p=p_f$ for foreground and $p=p_b$ for background. For our
$k=3$ and $\tau=2$,
\begin{equation}
\Pr[\nu(g)\ge2\mid g\text{ background}]
=3p_b^2(1-p_b)+p_b^3
\le 3p_b^2 .
\end{equation}
\end{proposition}

The proof is in the appendix. Under conditional independence, background
survival is quadratic in $p_b$ for small $p_b$, motivating $B$, $T$, and $O$
to discourage repeated support, while $A$ favors reliable foreground evidence.
Real views are correlated: on our scenes, two-of-three agreement is about twice
as frequent as independence predicts (median; appendix), so the
proposition is an idealization. Increasing the number of votes alone therefore does not
necessarily provide stronger evidence; the selector instead seeks a small set
of reliable and complementary observations.

Voting is most effective for clutter that is separated from the target and
supported by few masks. Target-adjacent structure is harder, because it can
stay aligned with the target across several views; Section~\ref{sec:ablation}
evaluates geometric refinement for this regime.

\begin{algorithm}[t]
\caption{\method}
\label{alg:cleangs}
\begin{algorithmic}[1]
\Require Gaussians $\mathcal{G}$, calibrated views $\mathcal{V}$ with masks
$\{M_v\}$, sparse tracks, subset size $k=3$, threshold $\tau=2$
\Ensure Retained Gaussians $\mathcal{G}[\mathcal{K}]$
\State \textbf{score} each $v$: occupancy $r_v$, band-pass $a_v$, masked
track set $Q_v$, and reject truncated candidates
\State \textbf{reduce} the pool by mask-level dominance, then cap it, giving
$\mathcal{V}_r$
\ForAll{$S\subseteq\mathcal{V}_r$ with $|S|=k$}
\State evaluate $f(S)=(A,B,O,T)$
\EndFor
\State $\mathcal{F}\gets$ exact Pareto front of $\{f(S)\}$
\Comment{Lemma~\ref{lem:dominance}}
\State $S^\star\gets\arg\min_{S\in\mathcal{F}}\max_\ell
w_\ell(z_\ell-\widetilde f_\ell(S))$
\ForAll{$v\in S^\star$}
\State rasterize $\mathcal{G}$ and record visibility and projected centers
\EndFor
\State $\nu_i\gets\sum_{v\in S^\star}X_v(g_i)$
\State $\mathcal{K}\gets\{i:\nu_i\ge\tau\}$
\Comment{Prop.~\ref{prop:binomial}}
\State \Return $\mathcal{G}[\mathcal{K}]$, attributes unchanged
\end{algorithmic}
\end{algorithm}

Algorithm~\ref{alg:cleangs} summarizes the complete procedure. Cleanup is
purely subtractive: the output contains only Gaussians from the original scene,
with their geometry, opacity, and appearance unchanged, and requires no
gradient-based optimization.

For $n$ candidates after reduction, subset scoring considers
$\binom{n}{3}$ triples, while all-pairs construction of the exact Pareto front
costs $O(\binom{n}{3}^{2})$ and dominates selection. Candidate reduction is
therefore important for keeping exact search practical. By
Lemma~\ref{lem:dominance}, minimizing the Chebyshev criterion directly over all
triples returns the same $S^\star$ in $O(\binom{n}{3})$ time; we build
$\mathcal{F}$ only to expose the trade-offs. Once $S^\star$ is
selected, filtering requires only three rasterization passes and is linear in
scene size. The appendix gives the full per-stage complexity analysis and
empirical subset counts.

\section{Retained-Asset Evaluation}
\label{sec:evaluation}

\subsection{Why Widen the Field of View?}

A cleaned Gaussian scene should preserve the appearance of the target while removing structure that should not remain in the exported asset. Standard capture views measure the first property well, but provide only limited evidence about the second. Gaussians outside the original image boundaries do not contribute to image-space metrics even when they remain part of the scene. Evaluating arbitrary novel viewpoints would expose more of this structure, but existing datasets generally do not provide corresponding ground-truth RGB observations.

We therefore introduce a controlled fixed-pose reframing test. For each evaluation camera, we keep the camera pose and image resolution unchanged and divide the focal length by six. We apply the corresponding image-plane transformation to the reference RGB image and target mask, and evaluate every method using the same transformed camera. Photometric metrics use only target pixels that have captured reference RGB, while leakage also counts newly exposed off-target pixels. The resulting view exposes a larger portion of the retained scene while preserving the original camera center and orientation.

This transformation is used only for evaluation. It does not modify the Gaussian representation or introduce new geometry, but changes which parts of the fixed scene fall inside the image. We refer to this setting as the \emph{widened-field} or \emph{zoom-$6$} evaluation. The results are stable to the magnitude of the focal-length change: at $2\times$ and $4\times$ widening, \method remains the strongest cleanup method in PSNR and the ordering among the three cleanup approaches is unchanged. The appendix reports all four focal-length settings.

\subsection{Metrics}

Let $I_v$ denote the reference image, $\widehat I_v$ the rendered image, $M_v$ the target mask, and $A_v$ the rendered opacity for camera $v$. We first define the masked squared error and the number of evaluated target pixels as
\begin{equation}
  E_v=\sum_p M_v(p)\lVert I_v(p)-\widehat I_v(p)\rVert_2^2,
\qquad Z_v=\sum_p M_v(p).
\end{equation}
We report masked
\begin{equation}
  \operatorname{MSE}_M(v)=\frac{E_v}{3Z_v},\qquad
  \operatorname{PSNR}_M(v)=10\log_{10}\!\left(1/\operatorname{MSE}_M(v)\right),
\end{equation}
together with masked SSIM \cite{ssim} and LPIPS \cite{lpips}. The factor of
three averages over the colour channels. These metrics are computed over the
target region only, so they are not directly comparable to full-image PSNR
reported on object-centric benchmarks; $Z_v$ falls from a mean of $5.8\times10^{5}$ pixels at the captured framing to
$2.3\times10^{4}$ at $6\times$, a factor of $24.6$ rather than the $36$ implied
by area alone, because some targets are already clipped by the captured frame.

Photometric fidelity alone does not measure how much off-target structure remains. We therefore additionally measure silhouette leakage,
\begin{equation}
\mathcal{L}_{\mathrm{leak}}(v)=
\frac{\sum_p(1-M_v(p))\,A_v(p)}
{\sum_p(1-M_v(p))},
\end{equation}
the mean rendered opacity outside the target region. The appendix also reports a thresholded variant that counts only pixels with $A_v(p)>0.1$. We also report the fraction of the original Gaussians retained by each method as a measure of asset compactness.

Leakage and retention are both compactness measures and therefore reward
more aggressive pruning: a method that deleted everything outside the target
could achieve zero leakage even if it damaged the target, while the uncleaned
scene scores $0.983$ because it retains the complete reconstruction. We
therefore report compactness together with photometric fidelity so that lower
leakage is interpreted alongside preservation of the target appearance. Unless
stated otherwise, aggregate leakage is reported at the captured ($1\times$)
framing using the full evaluation masks.

\subsection{Data, Protocol, and Comparators}

All scenes use pretrained \texttt{gsplat} checkpoints~\cite{gsplat}. We evaluate across a diverse scene set: 8 DL3DV~\cite{dl3dv}, 7 NeRFBusters~\cite{nfb}, and 3
Tanks and Temples~\cite{tnt} scenes, for 18 in total. These standard benchmarks
span indoor and outdoor settings, close-up and wider compositions, and casual
and more controlled captures. We report per-scene averages using the same
$6\times$ fixed-pose widening throughout. Target masks are obtained once from SAM~3~\cite{sam3} using manual prompting, checked only for target
coverage, and reused unchanged for every method. \method selects only from training
cameras, keeping evaluation masks unavailable to the selector. Matched
GaussianCut uses the same three selected training-view masks and therefore
follows the same separation between filtering and scored views.

We compare against the uncleaned \gs scene, FlashSplat, and GaussianCut by
rendering each exported Gaussian asset directly. FlashSplat uses removal
slackness $0$ and the same three masks as \method. GaussianCut is evaluated
both with these masks and with all available
capture masks; the latter, denoted GaussianCut$^\dag$, serves as a
denser-supervision reference. Further baseline details are given in the
appendix. Quantitative comparison is restricted to methods that export a directly
renderable Gaussian subset. Semantic-label methods such as Gaussian Grouping
and Gaga~\cite{gg,gaga} are discussed further in the appendix.

\begin{table*}[t]
  \centering
  \caption{\textbf{Retained-asset comparison on 18 scenes.}
  Standard: captured framing; Widened: $6\times$ fixed-pose widened field of
  view. All methods use the same scored frames and evaluation masks;
  FlashSplat, GaussianCut, and \method use the same three selected masks, and
  GaussianCut$^\dag$ uses all capture masks. Best photometric result is in
  \textbf{bold}; best cleanup statistic is \underline{underlined}.}
  \label{tab:main}
  \setlength{\tabcolsep}{3.2pt}
  \resizebox{\textwidth}{!}{%
  \begin{tabular}{l c ccc ccc cc}
    \toprule
    & Masks used
    & \multicolumn{3}{c}{Standard view}
    & \multicolumn{3}{c}{Widened field of view}
    & Leakage ($1\times$) $\downarrow$ & Retained $\downarrow$ \\
    \cmidrule(lr){3-5}\cmidrule(lr){6-8}
    Method
    & for cleanup
    & PSNR $\uparrow$ & SSIM $\uparrow$ & LPIPS $\downarrow$
    & PSNR $\uparrow$ & SSIM $\uparrow$ & LPIPS $\downarrow$
    & & \\
    \midrule
    Uncleaned 3DGS
      & 0
      & \textbf{19.35} & 0.620 & 0.194
      & 12.84 & 0.300 & 0.327
      & 0.983 & 100.0\% \\
    FlashSplat~\cite{flashsplat}
      & selected 3
      & 16.44 & 0.557 & 0.257
      & 16.29 & 0.411 & 0.205
      & 0.141 & \underline{14.9\%} \\
    GaussianCut~\cite{gaussiancut}
      & selected 3
      & 15.28 & 0.533 & 0.279
      & 15.10 & 0.381 & 0.227
      & 0.089 & 17.2\% \\
    GaussianCut$^\dag$~\cite{gaussiancut}
      & all capture
      & 17.27 & 0.572 & 0.234
      & 16.71 & 0.421 & 0.194
      & \underline{0.055} & 20.2\% \\
    \textbf{\method (ours)}
      & selected 3
      & 19.03 & \textbf{0.643} & \textbf{0.188}
      & \textbf{17.83} & \textbf{0.501} & \textbf{0.167}
      & 0.206 & 22.8\% \\
    \bottomrule
  \end{tabular}}
\end{table*}

\section{Results}
\label{sec:results}

\subsection{Appearance--Cleanup Trade-off}

Table~\ref{tab:main} summarizes the main retained-asset comparison. \method achieves the best result on five of the six photometric metrics while retaining $22.8\%$ of the original Gaussians. With the same three masks, \method improves over FlashSplat and GaussianCut by $2.59$ and $3.75\,\textrm{dB}$ in standard PSNR, and by $1.54$ and $2.73\,\textrm{dB}$ under widened framing. Compared with all-mask GaussianCut$^\dag$, the corresponding margins are $1.76\,\textrm{dB}$ and $1.12\,\textrm{dB}$.

The only photometric metric led by the uncleaned scene is standard-view PSNR, where the difference is $0.32\,\textrm{dB}$. \method nevertheless improves over the uncleaned scene in standard SSIM and LPIPS and in all three widened-field metrics. The scene-level results show a similar pattern. \method achieves higher PSNR at both framings on 16 of 18 scenes compared with FlashSplat and on 15 of 18 compared with all-mask GaussianCut$^\dag$, and on all 18 scenes compared with GaussianCut under matched three-mask supervision. Median PSNR gives the same ranking (appendix). Standard-view PSNR varies across scenes with a standard deviation of approximately $2.8\,\textrm{dB}$, and the appendix reports the complete per-scene results.

The cleanup statistics capture the complementary effect of more aggressive pruning. GaussianCut attains the lowest leakage, with $0.089$ under matched supervision and $0.055$ when given all capture masks, compared with $0.141$ for FlashSplat and $0.206$ for \method. FlashSplat is the most compact at $14.9\%$ retention. These gains in silhouette compactness are accompanied by lower target photometric fidelity, particularly under matched supervision. The all-mask GaussianCut result additionally uses masks at evaluation cameras, so we report it as an all-mask reference. Overall, \method occupies a favorable operating point that preserves target appearance while removing more than three quarters of the original Gaussian scene.

\subsection{What Widened Framing Reveals}

The widened-field evaluation highlights differences that are less apparent at
the captured framing. Widening also resamples the reference target, so we
compare methods within each framing rather than interpret absolute changes
between $1\times$ and $6\times$. The mechanism behind the widened ranking is
nevertheless clear. Reducing the focal length enlarges the view frustum, so
Gaussians that were previously culled now enter it. Those lying between the
camera and the target project with large screen-space footprints and composite
over target pixels, which degrades fidelity inside the mask even though the
offending structure is outside it. The uncleaned scene admits the most such
structure and moves from first at the captured framing to last at $6\times$,
while cleaned assets have little left to admit. Figure~\ref{fig:framing}
illustrates the same behavior qualitatively.

Standard and widened PSNR measure target fidelity under different framings and are therefore complementary, while widened leakage separately measures off-target structure exposed by the larger field of view. Within the captured framing, \method stays within $0.32\,\textrm{dB}$ of the uncleaned reconstruction, whereas GaussianCut$^\dag$ gives up $2.08\,\textrm{dB}$. Within the widened framing, \method exceeds the uncleaned scene by $4.99\,\textrm{dB}$ and GaussianCut$^\dag$ by $3.87\,\textrm{dB}$, and \method achieves the strongest PSNR among the cleanup methods. \method therefore preserves much of the target fidelity of the original reconstruction while producing an asset that is substantially more stable under reframing.

\subsection{Selector Coverage}

We next isolate the contribution of view selection. Train-only candidate pools
are available for all 18 scenes, and exact selection completes for every scene
at the $k=3$ operating point used in Table~\ref{tab:main}.

We hold the voting procedure and evaluator fixed and replace the selected
subset with three uniformly sampled training views, using five random seeds per
scene and averaging within scene. The paired photometric comparison covers 17
scenes because, on \emph{super markets}, three random subsets retain no
Gaussians and therefore have undefined photometrics. Pareto-selected views
improve all six photometric metrics. Standard PSNR increases by
$1.56\,\textrm{dB}$, with selection favored on 15 of 17 scenes, while widened
PSNR improves by $1.00\,\textrm{dB}$, favored on 14 of 17. Two-sided Wilcoxon
signed-rank tests give $p{=}0.0005$ and $p{=}0.015$, respectively. The appendix reports the corresponding SSIM, LPIPS,
leakage, and retention results.

Random subsets retain fewer Gaussians and achieve lower leakage on average, but
also preserve less target appearance and can fail on difficult scenes. On
\emph{super markets}, three of five random subsets remove every Gaussian,
whereas \method produces a non-empty output on all 18 scenes.

\begin{figure*}[t]
\centering
\setlength{\tabcolsep}{1pt}
\newcommand{\rl}[1]{\rotatebox{90}{\makebox[.068\textwidth][c]{\tiny #1}}}
\begin{tabular}{@{}c@{\hspace{1pt}}ccccccccc@{}}
& \scriptsize GT & \scriptsize \gs & \scriptsize \gs ($6\times$) & \scriptsize FlashSplat & \scriptsize FS ($6\times$) & \scriptsize GaussianCut & \scriptsize GCut ($6\times$) & \scriptsize Ours & \scriptsize Ours ($6\times$) \\
\rl{Family} &
\includegraphics[width=.104\textwidth]{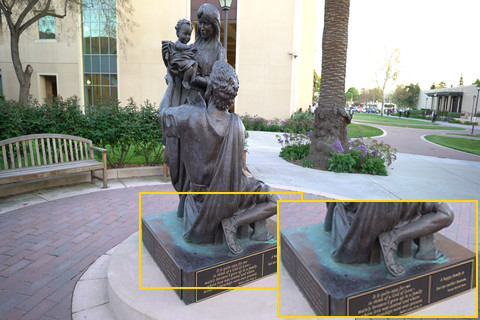} &
\includegraphics[width=.104\textwidth]{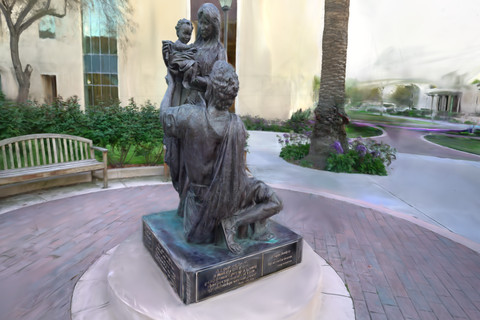} &
\includegraphics[width=.104\textwidth]{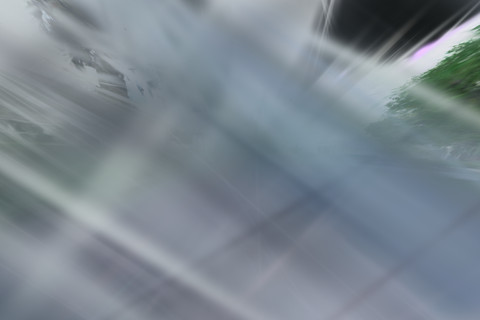} &
\includegraphics[width=.104\textwidth]{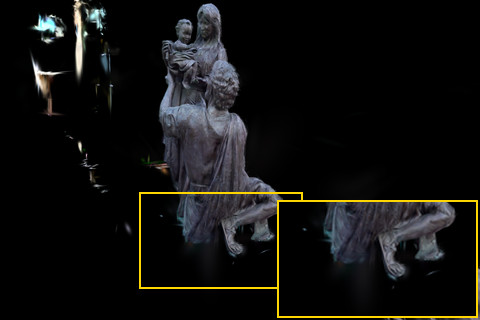} &
\includegraphics[width=.104\textwidth]{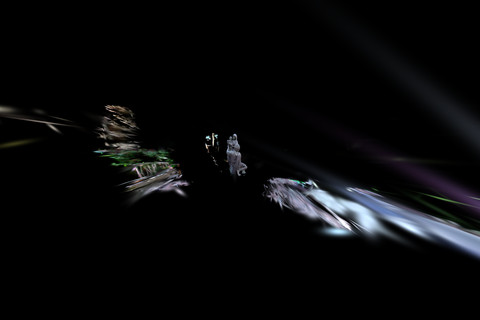} &
\includegraphics[width=.104\textwidth]{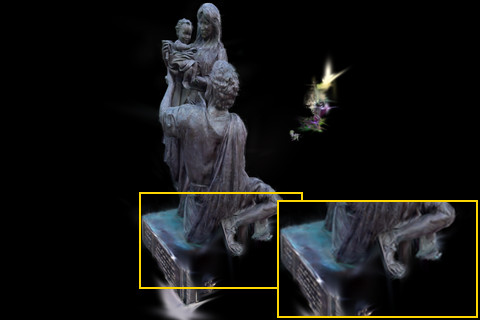} &
\includegraphics[width=.104\textwidth]{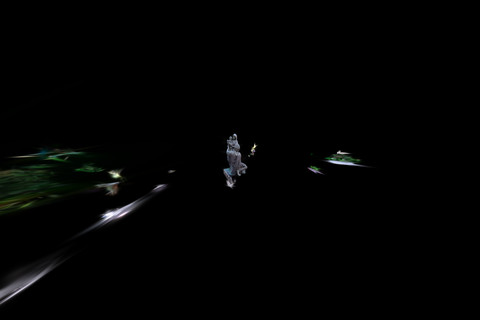} &
\includegraphics[width=.104\textwidth]{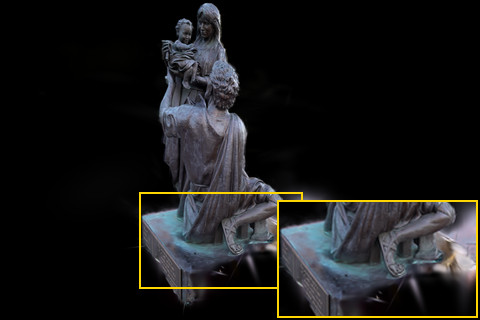} &
\includegraphics[width=.104\textwidth]{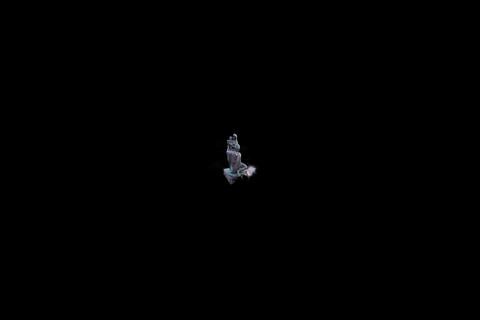} \\
\rl{courthouses} &
\includegraphics[width=.104\textwidth]{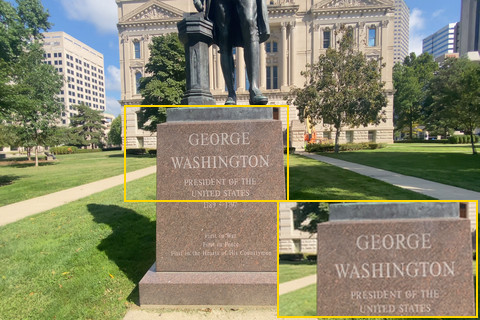} &
\includegraphics[width=.104\textwidth]{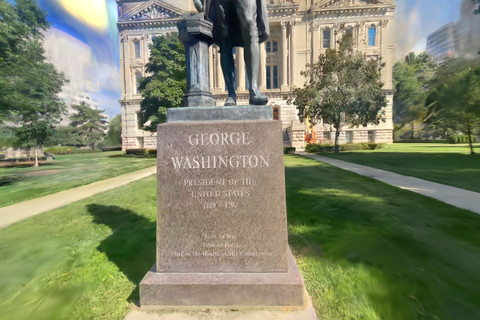} &
\includegraphics[width=.104\textwidth]{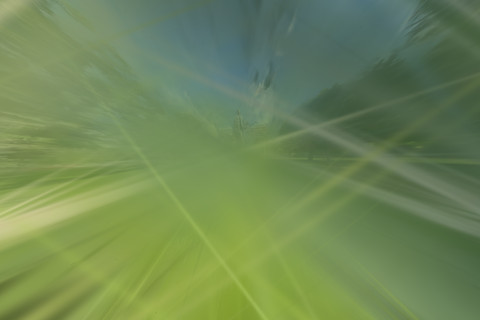} &
\includegraphics[width=.104\textwidth]{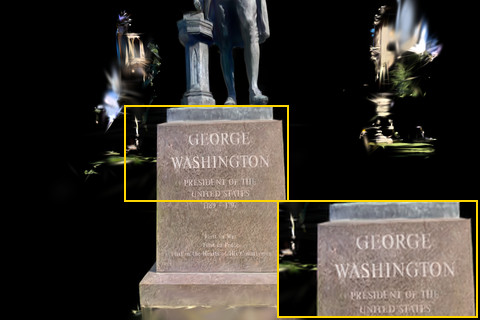} &
\includegraphics[width=.104\textwidth]{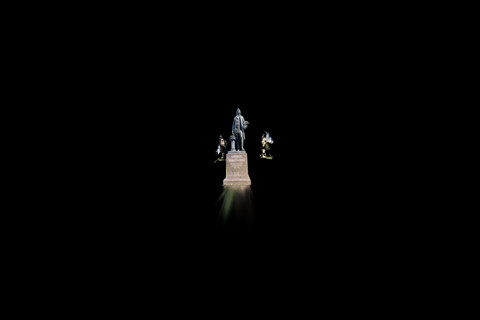} &
\includegraphics[width=.104\textwidth]{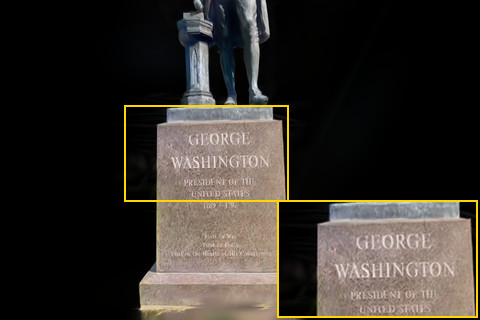} &
\includegraphics[width=.104\textwidth]{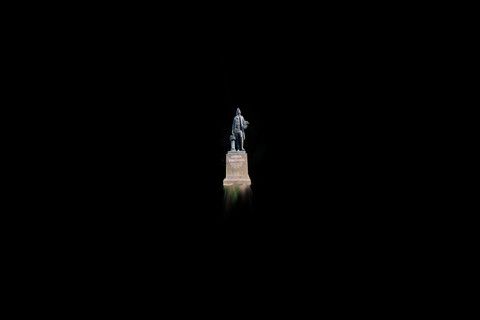} &
\includegraphics[width=.104\textwidth]{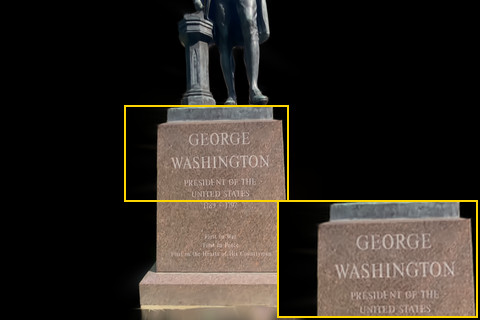} &
\includegraphics[width=.104\textwidth]{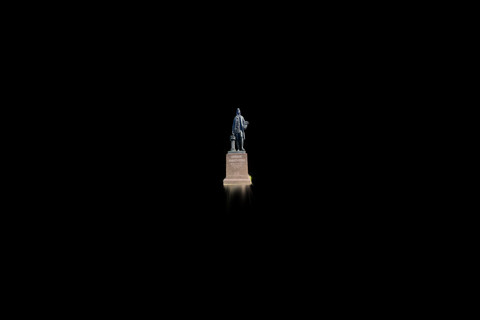} \\
\rl{garbage} &
\includegraphics[width=.104\textwidth]{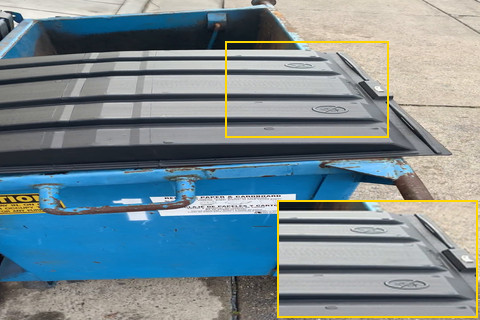} &
\includegraphics[width=.104\textwidth]{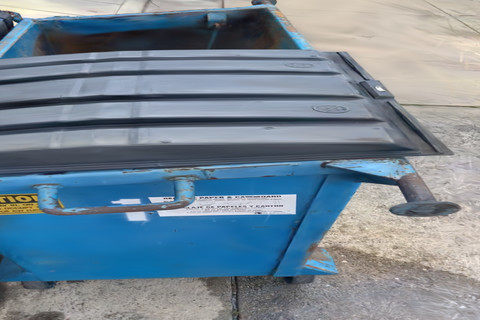} &
\includegraphics[width=.104\textwidth]{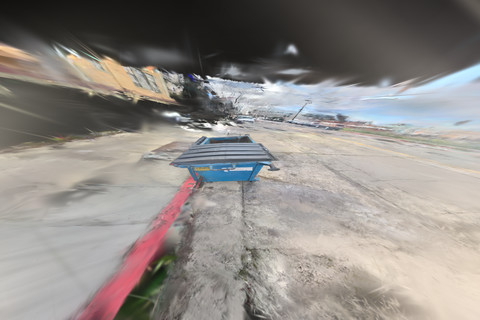} &
\includegraphics[width=.104\textwidth]{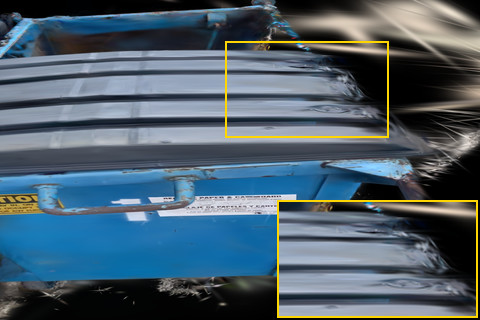} &
\includegraphics[width=.104\textwidth]{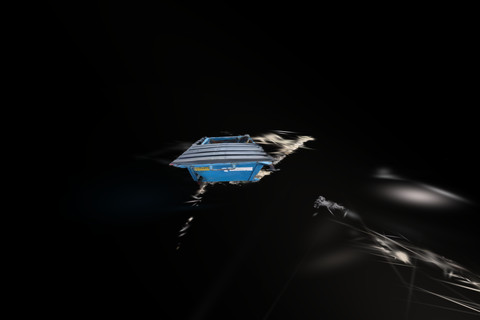} &
\includegraphics[width=.104\textwidth]{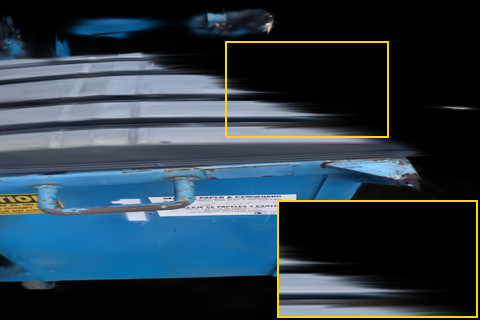} &
\includegraphics[width=.104\textwidth]{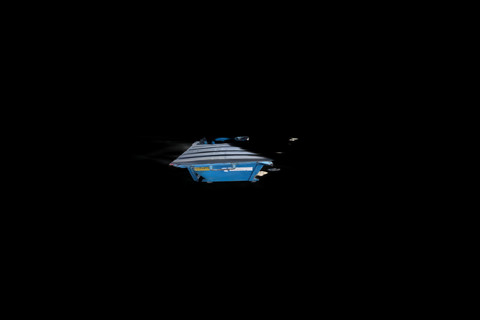} &
\includegraphics[width=.104\textwidth]{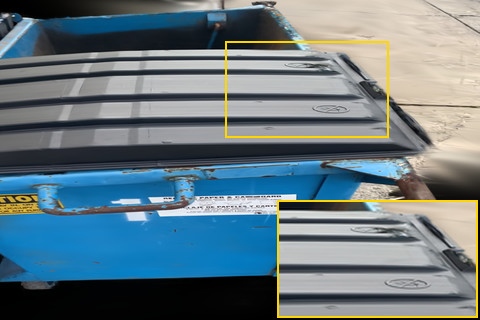} &
\includegraphics[width=.104\textwidth]{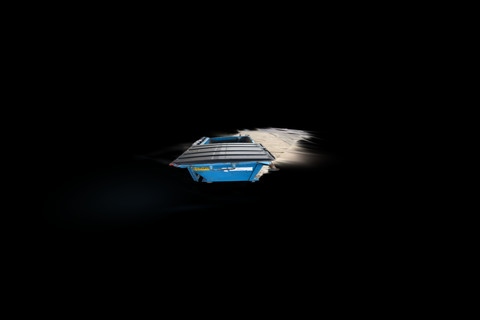} \\
\rl{fountain} &
\includegraphics[width=.104\textwidth]{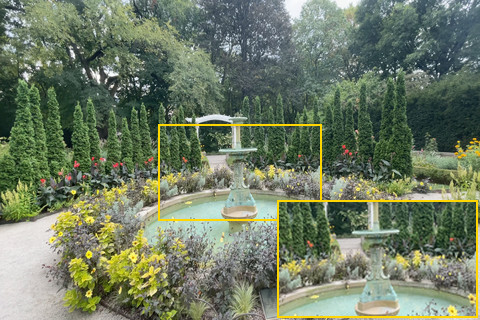} &
\includegraphics[width=.104\textwidth]{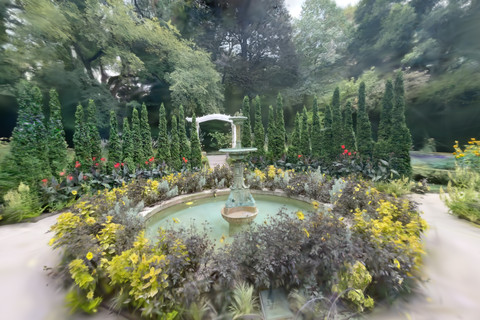} &
\includegraphics[width=.104\textwidth]{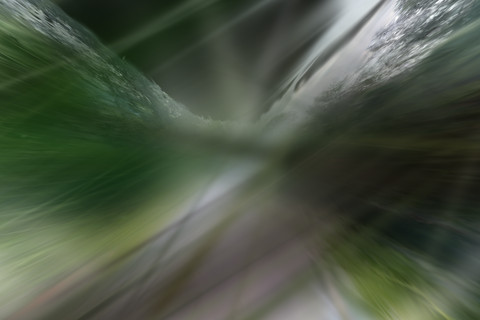} &
\includegraphics[width=.104\textwidth]{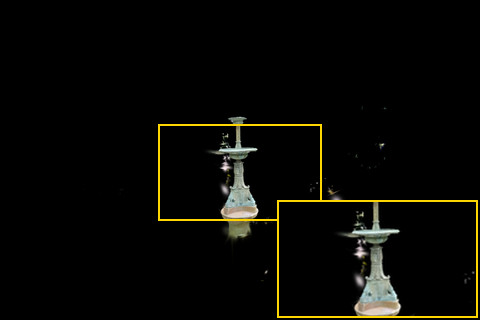} &
\includegraphics[width=.104\textwidth]{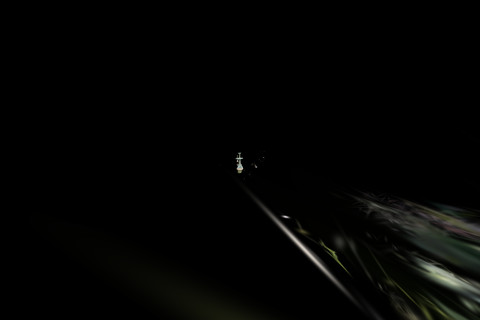} &
\includegraphics[width=.104\textwidth]{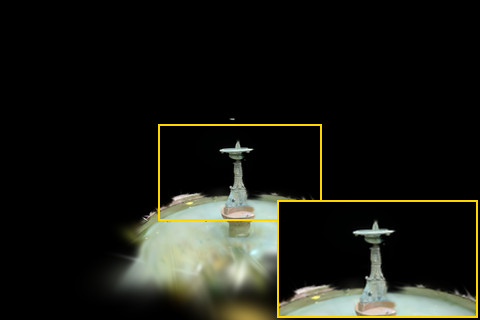} &
\includegraphics[width=.104\textwidth]{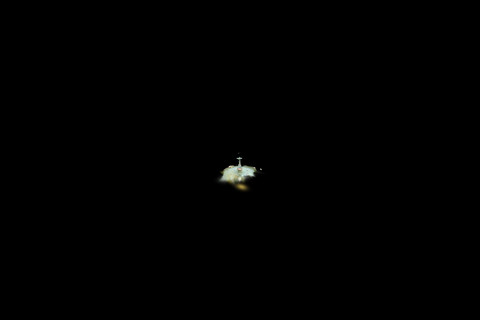} &
\includegraphics[width=.104\textwidth]{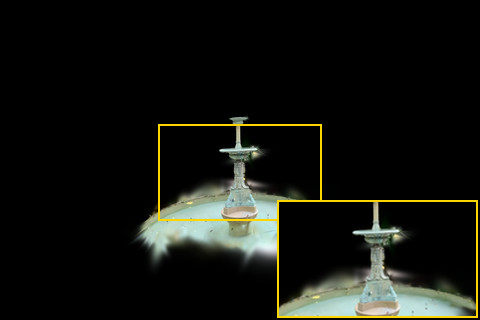} &
\includegraphics[width=.104\textwidth]{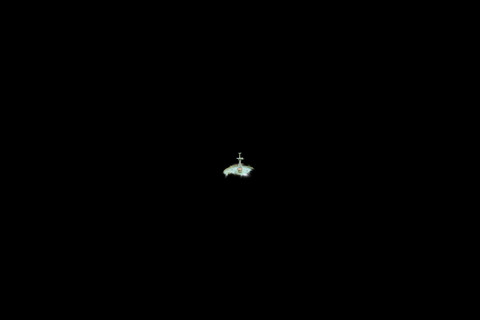}
\end{tabular}
\caption{\textbf{Retained target appearance.} Captured and $6\times$ widened renderings use the same camera pose, so widening reveals content already retained in the exported asset. FlashSplat introduces visible target artifacts, while GaussianCut
often produces tighter silhouettes at the cost of darker target regions.
\method better preserves details such as the plinth, inscription, lettering,
and fountain basin.}
\Description{Captured and widened renderings of four scenes comparing the uncleaned scene, FlashSplat, matched three-mask GaussianCut, and CleanGS.}
\label{fig:qualitative}
\end{figure*}

\subsection{Cleanup Cost}
\label{sec:efficiency}

\begin{table}[t]
  \centering
  \caption{\textbf{Filtering and export cost per scene} over 18 scenes at the matched three-mask budget. Times and peak memory exclude view selection.}
  \label{tab:efficiency}
  \setlength{\tabcolsep}{5pt}
  \fitcol{%
  \begin{tabular}{l r r r}
    \toprule
    Method & Time $\downarrow$ & Range & Peak memory $\downarrow$ \\
    \midrule
    FlashSplat~\cite{flashsplat} & 3.1\,min & 0.7--6.2\,min & 41.6\,GB \\
    GaussianCut~\cite{gaussiancut} & 4.2\,min & 1.0--14.4\,min & 34.9\,GB \\
    \textbf{CleanGS (ours)} & \textbf{20\,s} & 12--58\,s & \textbf{1.9\,GB} \\
    \bottomrule
  \end{tabular}}
\end{table}

Table~\ref{tab:efficiency} compares filtering-and-export cost at the same
three-mask filtering budget. Filtering and export take a median of 20 seconds
for \method, compared with 3.1 minutes for FlashSplat and 4.2 minutes for
GaussianCut, approximately $9\times$ and $13\times$ lower runtime. Peak memory
is 1.9\,GB for \method, compared with 41.6\,GB and 34.9\,GB, an approximately
$18$--$22\times$ reduction. View selection is a separate CPU stage and is
excluded from these numbers; including it, the end-to-end median is
41 seconds (appendix).

The efficiency follows directly from the structure of the pipeline. After view selection, \method requires three rasterization passes and a vote over the resulting support indicators, with no per-Gaussian semantic feature field, no Gaussian adjacency graph, and no global optimization over the full scene. GaussianCut instead builds a graph whose size grows with the Gaussian count and solves a max-flow problem. All runs use an A100-80GB GPU.

\subsection{Does Geometric Refinement Help?}
\label{sec:ablation}

The multiview vote removes structure that is inconsistent with the selected
masks, but nearby surfaces that remain mask-consistent can survive. We therefore
test a simple geometric refinement by applying HDBSCAN~\cite{hdbscan} to the
retained Gaussians and keeping only the largest density component.

Under the eroded evaluation masks used for this paired ablation, clustering
reduces leakage from $0.221$ to $0.115$, showing that spatial connectivity
provides useful cleanup signal, but standard and widened PSNR drop by
$3.84\,\textrm{dB}$ and $3.28\,\textrm{dB}$, respectively. Both endpoints use
the same eroded masks, so the comparison isolates the clustering stage; its
absolute values are therefore not directly comparable with
Table~\ref{tab:main}. Complete aggregate and per-scene results are given in
the appendix.

The fidelity loss is concentrated in scenes where the target spans multiple
density components. Six scenes lose more than $3\,\textrm{dB}$ in standard
PSNR, while the remaining twelve change only slightly. On \emph{tractor}, for
example, the largest of 92 clusters contains only $9\%$ of the retained
Gaussians, so keeping a single component removes substantial target structure,
indicating that a valid target can occupy several density modes. We therefore use the vote-only output as the \method endpoint.

\section{Limitations}

\begin{figure}[t]
\centering
\setlength{\tabcolsep}{1pt}
\begin{tabular}{@{}cccc@{}}
\scriptsize GT & \scriptsize GCut ($6\times$) & \scriptsize Ours & \scriptsize Ours ($6\times$) \\
\includegraphics[width=.235\columnwidth]{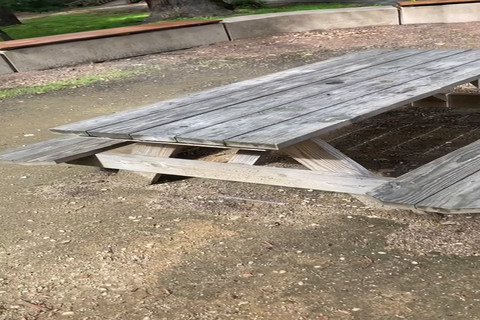} &
\includegraphics[width=.235\columnwidth]{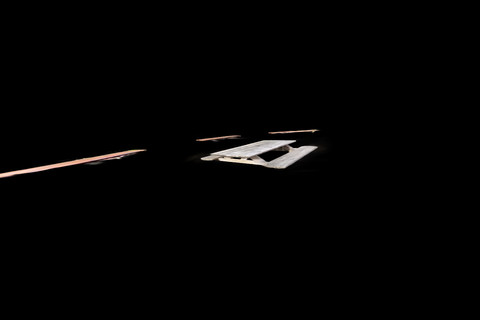} &
\includegraphics[width=.235\columnwidth]{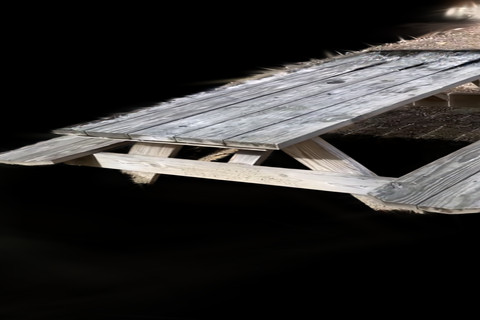} &
\includegraphics[width=.235\columnwidth]{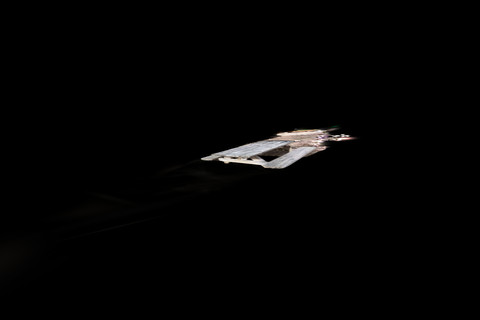} \\
\includegraphics[width=.235\columnwidth]{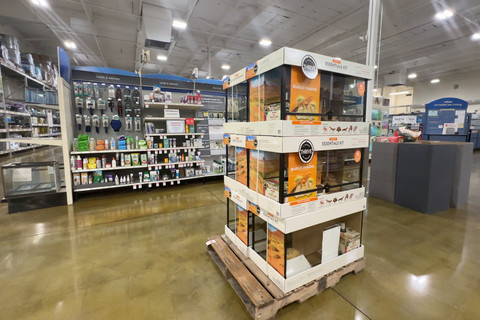} &
\includegraphics[width=.235\columnwidth]{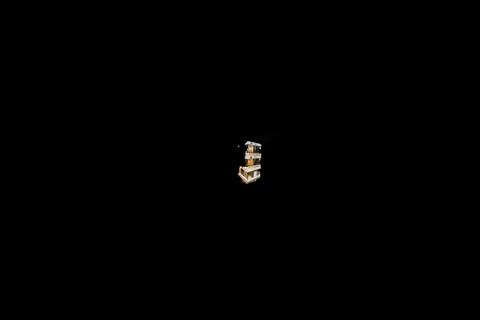} &
\includegraphics[width=.235\columnwidth]{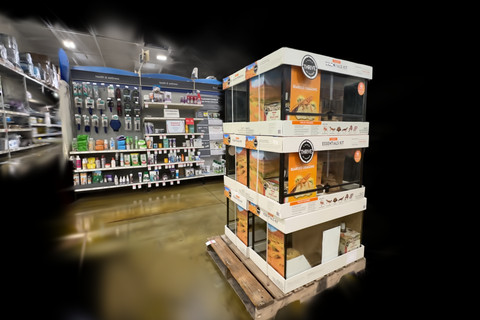} &
\includegraphics[width=.235\columnwidth]{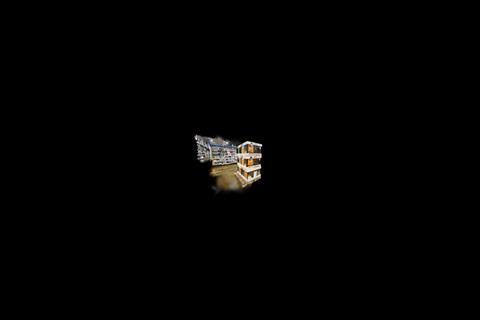}
\end{tabular}
\caption{\textbf{Residual target-adjacent structure.}
Nearby surfaces in \emph{super markets} and \emph{picnic} remain supported
across selected views. Their widened leakage is $0.221$ and $0.099$, against a
benchmark mean of $0.066$. GaussianCut$^\dag$ removes these regions with
all-mask
supervision; on \emph{picnic}, it is $1.49\,\textrm{dB}$ better in widened
PSNR.}
\Description{Two rows showing ground truth, GaussianCut widened, and
CleanGS at both fields of view, for the picnic and super markets scenes,
where CleanGS retains a large attached surface.}
\label{fig:limits}
\end{figure}

\method preserves target appearance well, but nearby surfaces can survive when
they remain consistently aligned with the target across selected views (appendix). Figure~\ref{fig:limits} shows two representative failures. In
\emph{super markets}, a shelving surface behind the target remains supported
across views; in \emph{picnic}, an attached surface survives for the same
reason. These cases illustrate the main limitation of semantic voting when
target and background remain consistently aligned across viewpoints.

Renderer visibility indicates whether a Gaussian can contribute to an image,
but not whether it is geometrically unoccluded by the target. Background along
the same viewing ray can therefore receive repeated support. Neither a
largest-component prior (Section~\ref{sec:ablation}) nor simple mask erosion
reliably resolves this case: 4-pixel erosion reduces leakage from $0.206$ to
$0.196$ but costs $0.66\,\textrm{dB}$ in standard PSNR. Mask membership is
also tested only at the projected Gaussian center, making thin boundaries
sensitive to mask error (a footprint-aware variant is evaluated in the
appendix), while transparent and reflective targets may genuinely
depend on nearby radiance.

The selector additionally depends on diversity and reconstructed support in
the available views. Sparse tracks are less informative for poorly reconstructed
targets, limited camera variation weakens the geometric objectives, and
selection cannot recover evidence absent from the candidate masks. Finally,
fixed-pose widening tests controlled reframing rather than arbitrary camera
motion and cannot expose permanently occluded clutter.

\section{Conclusion}

We presented \method, a retraining-free cleanup pipeline for pretrained \gs
scenes. It selects reliable and complementary views using semantic and
geometric objectives, chooses a balanced three-view subset from the exact
Pareto front over the reduced candidate pool, and applies rasterization-aware
two-of-three filtering while leaving the parameters of retained Gaussians
unchanged. Across 18 scenes, \method achieves the best result on five of six reported
photometric metrics, with its advantage particularly clear under widened
framing, and its filtering-and-export stage runs $9$--$13\times$ faster with
$18$--$22\times$ lower peak memory than the retained-asset baselines. The
random-view comparison further shows that selecting the evidence
used for filtering is essential. Future work can combine multiview semantic evidence with stronger occlusion and geometry-aware reasoning for target-adjacent structure and explore sequential selection under an explicit supervision budget.

\section*{Code Availability}
Code is available at \url{https://github.com/smlab-niser/clean-gs/tree/main}.

\begin{acks}
Computational resources for this work were partly funded by the Department of Atomic Energy, Government of India, under project code: RIN4001. The authors acknowledge the support of the National Institute of Science Education and Research (NISER), Bhu\-baneswar, and Homi Bhabha National Institute (HBNI), Mumbai.
\end{acks}

\bibliographystyle{ACM-Reference-Format}
\bibliography{references}

\appendix
\setlength{\emergencystretch}{2em}
\paragraph{Notation.} In this appendix, \emph{standard} (or \emph{captured}, $1\times$) refers to
evaluation at the original camera framing, and \emph{widened} (or \emph{zoomed},
$6\times$) to the fixed-pose evaluation with the focal length divided by six.
In tables, the subscript $6\times$ and the label ``Zoom'' refer to the widened
setting.

\section{Proofs}

\subsection{Dominance is preference-free}

We restate the dominance lemma, Lemma~\ref{lem:dominance}, here. Throughout
this analysis the feasible set consists of the $k$-subsets of the reduced
candidate pool $\mathcal{V}_r$ that the selector actually enumerates.
Let $f(S)=(A(S),B(S),O(S),T(S))$ and
let $S_2$ \emph{dominate} $S_1$ when $f(S_2)\ge f(S_1)$ coordinatewise
with at least one strict inequality.  Let $\mathcal{F}$ be the set of
non-dominated subsets.

\begin{lemma}\label{lem:supp_dominance}
Let $U(S)=\Phi(f(S))$ for any $\Phi\colon\mathbb{R}^4\to\mathbb{R}$ that is
non-decreasing in each argument.  If $S_2$ dominates $S_1$ then
$U(S_2)\ge U(S_1)$, and $\max_S U(S)$ is attained on $\mathcal{F}$.
\end{lemma}

\begin{proof}
Write $f(S_1)=(x_1,\ldots,x_4)$ and $f(S_2)=(y_1,\ldots,y_4)$ with
$y_\ell\ge x_\ell$ for every $\ell$.  Replace the coordinates one at a
time.  Monotonicity of $\Phi$ in its $\ell$-th argument gives
$$
 \Phi(y_1,\ldots,y_{\ell-1},x_\ell,\ldots,x_4)
 \le
 \Phi(y_1,\ldots,y_{\ell-1},y_\ell,x_{\ell+1},\ldots,x_4)
$$
for each $\ell$, and chaining the four inequalities yields
$\Phi(f(S_1))\le\Phi(f(S_2))$.

For the second claim, let $S$ be any subset.  If $S\in\mathcal{F}$ there is
nothing to prove.  Otherwise $S$ is dominated, and since the number of
feasible subsets is finite, following a chain of dominating subsets terminates
at some $S'\in\mathcal{F}$.  Each step is non-decreasing in $U$, so
$U(S')\ge U(S)$.  Hence no subset outside $\mathcal{F}$ attains a value
larger than the maximum over $\mathcal{F}$.
\end{proof}

The weighted Chebyshev rule used in the main text is one such $\Phi$ after sign reversal, being non-increasing in each normalized shortfall $z_\ell-\widetilde f_\ell(S)$ and hence non-decreasing in each objective.  The
weights therefore express a preference among mutually incomparable subsets,
and they cannot select a subset that another subset beats on all four
objectives simultaneously.

\subsection{Independent support suppresses clutter}

The support proposition, Proposition~\ref{prop:binomial}, is restated and proved here.

\begin{proposition}\label{prop:supp_binomial}
Let
\begin{align*}
 p_f&=\Pr[X_v(g)=1\mid g\text{ foreground}],\\
 p_b&=\Pr[X_v(g)=1\mid g\text{ background}],
\end{align*}
with $p_f>p_b$, and assume the $k$ support indicators
$\{X_v(g)\}_{v\in S^\star}$ are conditionally independent and identically
distributed given the class of $g$.  Under the keep rule $\nu(g)\ge\tau$,
$$
 \Pr[\nu(g)\ge\tau\mid g]
 =\sum_{j=\tau}^{k}\binom{k}{j}
   p^{\,j}(1-p)^{k-j},
$$
with $p=p_f$ for foreground and $p=p_b$ for background.  For
$k=3$ and $\tau=2$,
$$
 \Pr[\nu(g)\ge2\mid g\text{ background}]=3p_b^2(1-p_b)+p_b^3\le 3p_b^2 .
$$
\end{proposition}

\begin{proof}
Under the stated assumptions $\nu(g)$ is a sum of $k$
conditionally i.i.d.\ Bernoulli indicators, hence binomial with parameters
$(k,p_f)$ for foreground and $(k,p_b)$ for
background.  The retention probabilities are the corresponding upper tails,
which gives the displayed sum.

For the bound, set $k=3$ and $\tau=2$ and expand:
$$
 \sum_{j=2}^{3}\binom{3}{j}p_b^{\,j}(1-p_b)^{3-j}
 =3p_b^2(1-p_b)+p_b^3=3p_b^2-2p_b^3 .
$$
Since $p_b\ge0$, $2p_b^3\ge0$, and therefore $3p_b^2-2p_b^3\le 3p_b^2$ for
all $p_b\in[0,1]$.
\end{proof}

\paragraph{Dependence between views.}
Proposition~\ref{prop:supp_binomial} is an idealized reference model rather than an assumption that
real capture views can satisfy exactly. Two views with similar camera centers,
viewing directions, and masked support can produce strongly correlated support
indicators. For example, if $X_1=X_2$ and $X_3$ is independent, a
two-of-three rule fires whenever the duplicated observation is positive, with
probability $p$, rather than the $p^2$ predicted by a two-trial binomial at
the same threshold. The diversity objectives $B$, $T$, and $O$ are
therefore intended to discourage repeated evidence, while $A$ favors views
whose semantic support is reliable.

\paragraph{Measured dependence.}
To measure this dependence, we recorded the per-view support indicator
$X_v(g)$ of every Gaussian during filtering; recording does not change the
output. Table~\ref{tab:supp_dependence} compares the selected views with
random three-view subsets (seeds 0--2, averaged within scene). Support is
strongly positively dependent: over co-visible Gaussians, the mean pairwise
$\phi$ coefficient is $0.61$, and two-of-three agreement occurs at a median
of $1.99\times$ the rate predicted under independence. Selected views also
share more support than random views: $27\%$ of supported Gaussians are
supported by exactly one selected view, against $45\%$ for random views
($p=0.001$, lower on 15 of 18 scenes). This shared support is consistent with
the higher retention and target fidelity of selected views
(Section~\ref{sec:supp_random}). Proposition~\ref{prop:supp_binomial} should therefore be read as an
idealized motivation for the vote rather than as a quantitative prediction.

\begin{table}[t]
  \centering
  \caption{\textbf{Empirical dependence of per-view support.}
  Support indicators $X_v(g)$ of the three filtering views over 18 scenes.
  Random subsets use seeds 0--2, averaged within scene; two \emph{table} runs
  with an unregistered camera are excluded. $p$ is a two-sided Wilcoxon
  signed-rank test.}
  \label{tab:supp_dependence}
  \setlength{\tabcolsep}{4pt}
  \fitcol{%
  \begin{tabular}{lccc}
    \toprule
    Measure & Selected & Random & $p$ \\
    \midrule
    Pairwise $\phi$, co-visible Gaussians (mean) & 0.61 & 0.51 & 0.32 \\
    $\Pr[\nu\ge2]$ / independence prediction (mean) & 3.3 & 2.6 & 0.09 \\
    Supported by exactly one view (fraction) & 0.27 & 0.45 & 0.001 \\
    \bottomrule
  \end{tabular}}
\end{table}

The vote threshold also matters when $k$ changes. At a fixed absolute
threshold,
$\Pr[\mathrm{Bin}(3,p)\ge2]=3p^2-2p^3>p^2=\Pr[\mathrm{Bin}(2,p)\ge2]$,
so simply adding votes can increase retention of both foreground and
background. This is the behavior seen in the budget sweep for a fixed
two-of-$k$ rule and motivates considering view budget and threshold together.

\paragraph{Regime captured by the bound.}
The quadratic term is most informative when $p_b$ is small, as for
free-space floaters or clutter whose projection falls inside the target mask
from only a limited set of directions. Target-adjacent structure is a harder
regime: when background remains aligned with the target across views,
$p_b$ approaches $p_f$ and semantic support alone provides less class
separation. This explains the residual structures highlighted in the main
text and motivates complementary geometric cues. The clustering study in
Section~\ref{sec:supp_cluster} shows that spatial connectivity can substantially
reduce leakage, while a global largest-component rule becomes unreliable when
the retained target is distributed across multiple density modes. Renderer
visibility introduces a related ambiguity, because it identifies potential 
contribution to a render rather than target-side geometric occlusion.

\section{View-Selection Details}

\subsection{Per-view band-pass scoring}

Each candidate view $v$ is scored by
$$
 a_v=\exp\!\left(-\tfrac{1}{2}\left(\tfrac{r_v-r_0}{\sigma}\right)^2\right),
 \qquad r_v=\frac{|M_v|}{|\Omega_v|},
$$
with $r_0=0.30$ and $\sigma=0.15$.  The score peaks at intermediate
occupancy rather than increasing with mask area, because occupancy is not
monotone with usefulness for semantic lifting.

At very small $r_v$, the target occupies few pixels, so a fixed
image-space boundary error represents a larger fraction of the silhouette.
These views also contain fewer masked sparse tracks, making $Q_v$ a weaker
summary of the target's reconstructed support. At very large $r_v$, the
candidate is often a close view with limited context or target truncation at
the image boundary. Intermediate occupancy therefore provides a useful
operating region for semantic lifting.

The hard containment tests complement the smooth occupancy score by detecting
truncation directly. Masks are rejected when they touch more than one image
border, approach the frame too closely, or nearly fill either image dimension.
Together, the soft band-pass score and hard containment tests favor views that
show a substantial, well-contained target with enough surrounding context for
multiview discrimination.

Concretely, we retain candidates whose raw occupancy lies in $[0.02,0.60]$,
allow at most one touched image border, require a normalized bounding-box
margin of at least $0.03$, and reject masks whose normalized bounding box
exceeds $0.90$ in either image dimension. We then apply the mask-level
dominance filter over reliability, border margin, and truncation, giving the
reduced pool $\mathcal{V}_r$ used for subset enumeration. Two fallbacks
keep this deterministic. If the dominance filter would leave fewer than $k$
candidates, it is skipped and every eligible mask is retained; at $k=3$ this
happens on \emph{century}, \emph{red brick build}, and \emph{picnic}. If the
selector still produces no subset, the containment tests are disabled and the
run repeats with the candidate pool capped at 60 views. The cap sorts
candidates by decreasing band-pass score $a_v$, keeps the first 60, and
breaks ties by the order in which views appear in the COLMAP model, so it
depends only on the scores and that ordering; at $k=3$ this fallback runs on
\emph{century}, \emph{red brick build}, and \emph{table}, and the cap removes
views only on \emph{century} and \emph{red brick build}. The primary path enumerates the
reduced pool uncapped.

\paragraph{Candidate counts.}
Across the 18 scenes, the selector scores a median of 301 candidate masks
(range 268--358). A median of 72.5 remain after the containment tests (0--258)
and 10.5 after the dominance filter (3--268), so the enumerated pool has a
median of 10.5 views (3--60). This gives a median of 142.5 scored triples
(1--34{,}220) and a Pareto front of 13 subsets (1--472). The Pareto front in
the main text is therefore exact over this reduced pool.

\subsection{The four subset objectives}

\paragraph{Coverage $A(S)$.}
The minimum over the subset provides a weakest-link reliability criterion.
A heavily truncated or under-segmented view can contribute systematically
incomplete evidence, while a mean can hide it behind two strong views. Using
the minimum requires every selected view to clear the same basic semantic bar
before it influences 3D filtering.

\paragraph{Baseline diversity $B(S)$.}
Multiview localization is governed by parallax, which vanishes when cameras
are nearly co-located.  Two views may both be semantically clean and visually
distinct and still provide nearly identical 3D evidence if their centers are
close. In that regime small image-space or mask errors produce large
uncertainty in the lifted support.  Larger baselines make the corresponding
rays intersect more stably in 3D~\cite{hartley2004multiple}.

\paragraph{Viewing-angle diversity $T(S)$.}
Baseline alone does not guarantee directional diversity.  Cameras can move
appreciably while still viewing the target from nearly the same direction,
especially under forward motion or along shallow arcs, leaving the subset
sampling a narrow portion of the target's appearance and surface support.
Angular diversity reduces the chance that the selected views all miss the same
regions or repeat the same occlusion pattern~\cite{schoenberger2016mvs}.

\paragraph{Masked-support complementarity $O(S)$.}
Structure-from-motion and multi-view stereo usually favor large track overlap
because shared observations support matching and joint optimization
\cite{schoenberger2016sfm}. Here calibration has already been established and
all masks refer to the same target, so subset selection can instead reward
complementary support. The Jaccard dissimilarity $O(S)$ favors pairs whose
masked track sets extend one another, reducing the chance that multiple votes
repeat essentially the same reconstructed evidence.

\paragraph{Normalization and scalarization.}
Each objective is min--max normalized across \emph{all} enumerated
$k$-subsets of $\mathcal{V}_r$, so the normalized values do not depend on 
which subsets turn out to be non-dominated. The weighted Chebyshev criterion 
is then evaluated on the normalized objectives of the front members. When an 
objective is constant across all subsets, its normalized column is set to one.
Thus, it contributes no shortfall to the Chebyshev criterion and the remaining 
objectives determine the selection.

\subsection{Candidate reduction, the pool cap, and direct scalarization}
\label{sec:supp_reduction}

\paragraph{Direct scalarization.}
Lemma~\ref{lem:supp_dominance} implies that the Chebyshev optimum lies on the Pareto front, so the
front need not be constructed to obtain $S^\star$: minimizing the criterion
directly over all enumerated $k$-subsets returns the same subset. In
practice, a direct argmin returns the selected subset on 16 of 18 scenes. The remaining
two, \emph{gardens} and \emph{tractor}, have exact ties in the Chebyshev value
(3 and 5 tied subsets), where the first tied index can be a dominated subset.
Removing dominated subsets within the tie set before taking the first index
recovers the selection on all 18 scenes, and on 15 further scenes outside the
benchmark. Direct minimization takes under $0.1$\,ms at $k=3$, whereas the
all-pairs front costs about 51\,s on the two scenes whose pool reaches the cap
of 60 views and under $0.1$\,s elsewhere. We keep the explicit front because
it makes the trade-offs among the objectives visible and supports the
reweighting analysis, but the pipeline does not need it. The $k\ge4$ arms in
Section~\ref{sec:supp_frontier} use the direct argmin.

\paragraph{Pool cap and dominance filter.}
Only \emph{century} (178 eligible candidates) and \emph{red brick build} (268)
have more than 60 eligible candidates, so the cap affects only these two scenes. Table~\ref{tab:supp_cap} removes or relaxes the cap
on these two scenes using the direct argmin. The effect is scene dependent and
has no consistent sign. Disabling the mask-level dominance filter on the six
scenes where it shrinks the pool most (e.g.\ \emph{Horse}, 169 to 11 views;
\emph{tractor}, 258 to 19) changes standard PSNR by $-0.40$ to $+0.30$\,dB
and widened PSNR by $-0.22$ to $+2.36$\,dB, and lowers leakage on all six
scenes ($-0.02$ to $-0.14$). The dominance filter therefore mainly keeps the
explicit front tractable; the direct argmin does not need it.

\begin{table}[t]
  \centering
  \caption{\textbf{Effect of the 60-view pool cap.}
  Standard (captured-framing) PSNR and leakage on the two scenes with more than
  60 eligible candidates; $\Delta$ is relative to the main cap of 60.}
  \label{tab:supp_cap}
  \setlength{\tabcolsep}{4pt}
  \fitcol{%
  \begin{tabular}{lccc}
    \toprule
    Scene & Cap 60 (main) & Cap 120 & Uncapped \\
    \midrule
    century & 19.94\,dB, 0.391 & 18.98 ($-0.96$), 0.351 & 18.57 ($-1.37$), 0.347 \\
    red brick build & 16.34\,dB, 0.196 & 16.82 ($+0.48$), 0.201 & 17.63 ($+1.29$), 0.207 \\
    \bottomrule
  \end{tabular}}
\end{table}

\subsection{Relation to active view selection and mask confidence}

Active view-selection methods for radiance fields ask which new observation
would most improve a reconstruction. ActiveNeRF selects additional views using
estimated uncertainty \cite{supp_activenerf}, while FisherRF uses Fisher
information and expected information gain for active view selection and mapping
\cite{supp_fisherrf}. \method addresses a complementary problem: the scene is
already reconstructed and the candidate cameras already exist, so the selector
chooses which available masked views should provide semantic evidence for
post-hoc cleanup rather than which image should be acquired next. These methods
nevertheless suggest uncertainty- or information-based alternatives to the
geometric diversity terms studied here.

Mask confidence provides another natural alternative to occupancy-based
reliability. The original Segment Anything model predicts an estimated IoU
score for each candidate mask \cite{supp_sam}; such model-derived confidence
could be used directly or combined with $a_v$. The occupancy score used here
is intentionally model-agnostic and depends only on the final binary mask and
camera frame. Comparing learned mask confidence with this simple geometric
proxy is a useful direction for a dedicated view-quality study.

\section{Complexity}

Let $n_{\mathrm{all}}$ be the number of capture views, $n$ the candidates
surviving reduction, $k=3$ the subset size, $\bar q$ the mean number of
masked sparse tracks per view, $N_G$ the Gaussian count, and $H\times W$
the image resolution.

\begin{itemize}
  \item Mask statistics: $O(n_{\mathrm{all}}HW)$.
  \item Masked track sets: $O(\sum_v q_v)$.
  \item View-level dominance filter: $O(n_{\mathrm{all}}^2)$.
  \item Subset scoring: $O(\binom{n}{k}k^2\bar q)$.  The pairwise terms are
  $O(k^2)$ per subset, but $O(S)$ compares track sets rather than scalars,
  so each pair carries a factor $\bar q$.
  \item Exact Pareto filtering: $O(\binom{n}{k}^2)$ by all-pairs dominance.
  This term dominates the selector.
  \item Direct Chebyshev minimization over all subsets: $O(\binom{n}{k})$,
  which returns the same $S^\star$ (Section~\ref{sec:supp_reduction}).
  \item Multiview filtering: $O(k(N_G+HW))$, one rasterization per selected
  view rather than $k$ scalar projections per Gaussian.
  \item Export: $O(N_K)$ for the $N_K\le N_G$ retained rows.
\end{itemize}

At the $k=3$ operating point, candidate reduction keeps the
exact front practical. Across the 18 evaluated scenes, the median reduced pool
has $n=10.5$ views, giving 142.5 scored subsets and a median Pareto front of 13,
compared with the $\binom{60}{3}=34{,}220$ subsets allowed by the pool cap. 
The reduction itself keeps the subset count small on all but 2 scenes. For dense 
reconstructions, the $\bar q$ factor in the masked support objective can dominate subset-scoring time.

The exact Pareto front makes the trade-offs among the four objectives explicit
and provides a compact set of non-dominated alternatives that can be
rescalarized under different weight vectors. This supports the weight-sensitivity
analysis in Section~\ref{sec:supp_selector_ablations} while preserving the
multi-objective structure of the selector.

\subsection{Measured end-to-end cost}

Table~\ref{tab:supp_cost} reports the cost of the complete \method pipeline,
measured sequentially on one A100-80GB GPU. The filtering-and-export
time in Table~\ref{tab:efficiency} excludes view selection; including it, the median end-to-end time is 41\,s
(12.8--128\,s). The selector runs on the CPU with a peak resident memory of
165\,MB, and its time is dominated by reading and scoring masks rather than by
the Pareto front, except on the two scenes whose pool reaches the cap. With the
direct argmin, the end-to-end median is 40\,s (12.8--77\,s).

\begin{table}[t]
  \centering
  \caption{\textbf{End-to-end cost of \method} over 18 scenes on one
  A100-80GB GPU, run sequentially. Selection runs on the CPU.}
  \label{tab:supp_cost}
  \setlength{\tabcolsep}{4pt}
  \fitcol{%
  \begin{tabular}{lrr}
    \toprule
    Stage & Median & Range \\
    \midrule
    Selection, explicit front (main) & 21.0\,s & 2.5--101.9\,s \\
    \quad mask scoring, COLMAP, view data & $\sim$20\,s & 2.4--50\,s \\
    \quad explicit Pareto front & $<0.1$\,s & up to 51\,s \\
    Selection, direct argmin & 17.3\,s & 2.5--50.3\,s \\
    Filtering and export & 18.7\,s & 10.2--57.1\,s \\
    \midrule
    Total, explicit front & 41\,s & 12.8--128\,s \\
    Total, direct argmin & 40\,s & 12.8--77\,s \\
    \midrule
    Peak GPU memory, filtering & 1.9\,GB & 0.9--6.1\,GB \\
    Peak CPU memory, selector / filtering & 165\,MB / 2.0\,GB & \\
    \bottomrule
  \end{tabular}}
\end{table}

\section{Filtering and Metric Definitions}

\paragraph{Renderer visibility.}
The indicator $\operatorname{vis}_v(g_i)$ is read directly from the gsplat
rasterization of the full scene at camera $v$. A Gaussian counts as visible
when the screen-space radius reported by the rasterizer is strictly positive,
that is, it survives frustum culling and is assigned to at least one tile, and
when its projected center $\pi_v(\mu_i)$, truncated to integer pixel
coordinates, falls inside the image. Mask membership is then the value of the
binary mask at that pixel. Note that this is a contribution test, instead of an 
occlusion test: a Gaussian lying behind the target is still visible in this sense, 
which is the ambiguity discussed in the main text.

\paragraph{Masked SSIM}
We compute the full-image SSIM map between render and reference with an
$11\times11$ Gaussian window and data range $1$, average it over the
three colour channels, and take its mask-weighted mean,
$\sum_p M_v(p)\,\mathrm{SSIM}(p)\big/\sum_p M_v(p)$. Windows that
straddle the silhouette therefore include some surrounding context, as they
would for any windowed structural measure.

\paragraph{Masked LPIPS}
We crop both images to the tight bounding box of $M_v$, zero the pixels
outside the mask within that crop, pad to at least $64\times64$ when the
box is smaller, and evaluate AlexNet LPIPS on the result. Cropping keeps the
comparison on the target rather than on a largely empty frame, which matters at
$6\times$ widening, where the target covers a small fraction of the image.

\paragraph{Widened-field scoring.}
At the widened framing, the reference RGB image and the target mask undergo the
same image-plane transformation as the camera. Reference radiance exists only
inside the original captured field of view, so masked PSNR, SSIM, and LPIPS are
computed only over target pixels with captured reference RGB. Leakage, in
contrast, is computed over all off-target pixels, including regions newly
exposed by widening, because it requires only the rendered opacity.

\paragraph{Leakage.}
All reported leakage values are the unthresholded mean rendered opacity over
off-target pixels,
$$
  \sum_p(1-M_v(p))A_v(p)/\sum_p(1-M_v(p)),
$$
averaged over evaluation views and then over scenes. The evaluator also computes the
fraction of off-target pixels with $A_v(p)>0.1$. At the captured framing this
thresholded variant is $1.000$ for the uncleaned scene, $0.201$ for
FlashSplat, $0.128$ for GaussianCut, $0.084$ for
$\text{GaussianCut}^{\dagger}$, and $0.269$ for \method, preserving the
ranking of the unthresholded measure.

\section{Footprint-Aware Mask Membership}
\label{sec:supp_footprint}

The main filtering rule tests mask membership at the projected Gaussian center.
Large anisotropic Gaussians can have screen-space footprints that extend beyond
the mask even when their centers lie inside it. We therefore evaluate a
footprint-aware variant: a visible Gaussian supports a view only if at least a
fraction $\tau$ of its projected 2D Gaussian mass lies inside the mask. The
mass is estimated from 64 fixed samples of the projected covariance, and
samples that fall outside the image count as outside the mask. The vote and
every other stage are unchanged. Table~\ref{tab:supp_footprint} reports the
results with the same selected masks.

\begin{table}[t]
  \centering
  \caption{\textbf{Footprint-aware mask membership} with the same three
  selected masks over 18 scenes. Std: captured framing; $6\times$: widened
  framing. $p$ compares each row with center lookup (two-sided Wilcoxon).}
  \label{tab:supp_footprint}
  \setlength{\tabcolsep}{3pt}
  \fitcol{%
  \begin{tabular}{lccccc}
    \toprule
    Membership, vote & Leak (std) & PSNR (std) & Leak ($6\times$) & PSNR ($6\times$) & Retained \\
    \midrule
    Center, 2-of-3 (main) & 0.206 & 19.03 & 0.066 & 17.83 & 22.8\% \\
    $\tau=0.5$, 2-of-3 & 0.194 & 19.03 & 0.064 & 17.85 & 22.9\% \\
    $\tau=0.75$, 2-of-3 & 0.169 & 18.56 & 0.052 & 17.70 & 22.5\% \\
    $\tau=0.9$, 2-of-3 & 0.148 & 17.96 & 0.044 & 17.32 & 22.1\% \\
    \midrule
    $\tau=0.9$, 3-of-3 & 0.085 & 15.45 & 0.024 & 15.26 & 17.7\% \\
    $\tau=0.9$, 3-of-4 & 0.109 & 16.15 & 0.026 & 15.72 & 19.3\% \\
    $\tau=0.9$, 3-of-5 & 0.130 & 17.97 & 0.032 & 17.40 & 21.6\% \\
    $\tau=0.75$, 3-of-5 & 0.150 & 18.56 & 0.039 & 17.77 & 22.0\% \\
    \bottomrule
  \end{tabular}}
\end{table}

At $\tau=0.5$, the footprint test lowers leakage from $0.206$ to
$0.194$ ($p=0.009$) while standard PSNR stays at $19.03$\,dB, so it
directly reduces boundary leakage caused by wide footprints. Stricter thresholds provide a controlled fidelity--cleanup knob:
$\tau=0.75$ and $\tau=0.9$ lower leakage to $0.169$ and $0.148$ at a
cost of $0.48$ and $1.08$\,dB in standard PSNR (both $p<0.001$).
Gaussian counts decrease monotonically with $\tau$ on all 18 scenes.
Section~\ref{sec:supp_frontier} places these operating points on the
fidelity--cleanup frontier.

\section{Per-Scene Results}

Tables~\ref{tab:supp_standard} and~\ref{tab:supp_zoomed} report per-scene
photometric results at the captured and widened framings, respectively, while
Table~\ref{tab:supp_cleanup} reports cleanup diagnostics. The GaussianCut
columns correspond to the all-mask $\text{GaussianCut}^{\dagger}$
configuration from the main text. The FlashSplat columns use the same three
selected masks as \method. Photometric macro
averages match those reported in the main text; leakage in
Table~\ref{tab:supp_cleanup} is measured under widened framing.

\begin{table*}[t]
  \centering
  \caption{\textbf{Per-scene standard-view photometrics.}
Results over the same 18 scenes as the main comparison.}
  \label{tab:supp_standard}
  \setlength{\tabcolsep}{3pt}
  \resizebox{\textwidth}{!}{%
  \begin{tabular}{ll r *{12}{c}}
    \toprule
    Dataset & Scene & Frames
    & \multicolumn{3}{c}{Uncleaned}
    & \multicolumn{3}{c}{FlashSplat}
    & \multicolumn{3}{c}{GaussianCut$^\dag$}
    & \multicolumn{3}{c}{CleanGS}
    \\
    \cmidrule(lr){4-6}\cmidrule(lr){7-9}\cmidrule(lr){10-12}\cmidrule(lr){13-15}
    & & & PSNR $\uparrow$ & SSIM $\uparrow$ & LPIPS $\downarrow$ & PSNR $\uparrow$ & SSIM $\uparrow$ & LPIPS $\downarrow$ & PSNR $\uparrow$ & SSIM $\uparrow$ & LPIPS $\downarrow$ & PSNR $\uparrow$ & SSIM $\uparrow$ & LPIPS $\downarrow$ \\
    \midrule
    DL3DV & car\_set2 & 43 & 18.61 & 0.704 & 0.225 & 17.51 & 0.681 & 0.249 & 17.85 & 0.687 & 0.241 & 18.39 & 0.715 & 0.204 \\
     & courthouses & 52 & 19.37 & 0.665 & 0.239 & 18.90 & 0.648 & 0.252 & 18.97 & 0.650 & 0.249 & 22.89 & 0.703 & 0.202 \\
     & fountain & 44 & 20.63 & 0.602 & 0.096 & 14.69 & 0.487 & 0.215 & 18.19 & 0.566 & 0.143 & 19.47 & 0.632 & 0.108 \\
     & gardens & 44 & 18.79 & 0.551 & 0.153 & 15.84 & 0.505 & 0.208 & 17.45 & 0.537 & 0.170 & 18.64 & 0.579 & 0.143 \\
     & red\_brick\_build & 39 & 20.05 & 0.566 & 0.237 & 16.45 & 0.514 & 0.293 & 17.13 & 0.529 & 0.277 & 16.34 & 0.541 & 0.278 \\
     & redcar & 41 & 18.24 & 0.688 & 0.222 & 17.29 & 0.669 & 0.247 & 17.57 & 0.673 & 0.239 & 19.15 & 0.725 & 0.184 \\
     & super\_markets & 47 & 10.91 & 0.396 & 0.579 & 7.50 & 0.322 & 0.659 & 4.26 & 0.179 & 0.788 & 10.76 & 0.412 & 0.565 \\
     & tractor & 41 & 20.66 & 0.736 & 0.133 & 18.40 & 0.683 & 0.170 & 17.13 & 0.675 & 0.184 & 19.91 & 0.760 & 0.130 \\
    NeRFBusters & car & 42 & 17.88 & 0.616 & 0.244 & 16.67 & 0.570 & 0.300 & 16.84 & 0.579 & 0.287 & 19.00 & 0.650 & 0.234 \\
     & century & 50 & 19.71 & 0.686 & 0.126 & 15.20 & 0.529 & 0.258 & 17.18 & 0.628 & 0.193 & 19.94 & 0.741 & 0.109 \\
     & flowers & 44 & 19.81 & 0.533 & 0.151 & 15.50 & 0.441 & 0.202 & 15.39 & 0.439 & 0.206 & 16.39 & 0.521 & 0.171 \\
     & garbage & 45 & 19.77 & 0.717 & 0.211 & 14.08 & 0.611 & 0.351 & 19.32 & 0.708 & 0.215 & 22.06 & 0.768 & 0.165 \\
     & picnic & 40 & 19.29 & 0.508 & 0.090 & 14.07 & 0.399 & 0.191 & 18.35 & 0.475 & 0.112 & 16.37 & 0.464 & 0.149 \\
     & roses & 46 & 19.46 & 0.519 & 0.105 & 18.01 & 0.492 & 0.138 & 18.93 & 0.504 & 0.117 & 20.06 & 0.593 & 0.081 \\
     & table & 36 & 26.11 & 0.909 & 0.076 & 24.42 & 0.900 & 0.083 & 20.03 & 0.774 & 0.143 & 19.94 & 0.880 & 0.105 \\
    TnT & Family & 42 & 21.17 & 0.606 & 0.262 & 20.12 & 0.584 & 0.273 & 20.32 & 0.588 & 0.272 & 23.80 & 0.689 & 0.230 \\
     & Francis & 42 & 18.44 & 0.510 & 0.192 & 14.38 & 0.415 & 0.340 & 17.32 & 0.481 & 0.210 & 18.60 & 0.499 & 0.188 \\
     & Horse & 42 & 19.39 & 0.647 & 0.158 & 16.91 & 0.578 & 0.203 & 18.66 & 0.624 & 0.171 & 20.89 & 0.707 & 0.144 \\
    \bottomrule
  \end{tabular}}
\end{table*}

\begin{table*}[t]
  \centering
  \caption{\textbf{Per-scene widened-field photometrics.}
Results over the same 18 scenes at $6\times$ widened framing.}
  \label{tab:supp_zoomed}
  \setlength{\tabcolsep}{3pt}
  \resizebox{\textwidth}{!}{%
  \begin{tabular}{ll r *{12}{c}}
    \toprule
    Dataset & Scene & Frames
    & \multicolumn{3}{c}{Uncleaned}
    & \multicolumn{3}{c}{FlashSplat}
    & \multicolumn{3}{c}{GaussianCut$^\dag$}
    & \multicolumn{3}{c}{CleanGS}
    \\
    \cmidrule(lr){4-6}\cmidrule(lr){7-9}\cmidrule(lr){10-12}\cmidrule(lr){13-15}
    & & & PSNR $\uparrow$ & SSIM $\uparrow$ & LPIPS $\downarrow$ & PSNR $\uparrow$ & SSIM $\uparrow$ & LPIPS $\downarrow$ & PSNR $\uparrow$ & SSIM $\uparrow$ & LPIPS $\downarrow$ & PSNR $\uparrow$ & SSIM $\uparrow$ & LPIPS $\downarrow$ \\
    \midrule
    DL3DV & car\_set2 & 43 & 15.31 & 0.445 & 0.313 & 17.50 & 0.508 & 0.211 & 17.78 & 0.508 & 0.196 & 18.45 & 0.577 & 0.169 \\
     & courthouses & 52 & 12.22 & 0.324 & 0.492 & 17.65 & 0.480 & 0.238 & 17.59 & 0.480 & 0.237 & 20.78 & 0.568 & 0.184 \\
     & fountain & 44 & 12.64 & 0.192 & 0.214 & 16.01 & 0.430 & 0.091 & 18.60 & 0.489 & 0.061 & 20.14 & 0.604 & 0.048 \\
     & gardens & 44 & 12.32 & 0.159 & 0.219 & 16.38 & 0.414 & 0.096 & 16.77 & 0.428 & 0.089 & 17.22 & 0.484 & 0.088 \\
     & red\_brick\_build & 39 & 19.54 & 0.353 & 0.295 & 16.55 & 0.315 & 0.324 & 17.16 & 0.331 & 0.313 & 16.50 & 0.337 & 0.312 \\
     & redcar & 41 & 12.77 & 0.359 & 0.419 & 17.25 & 0.482 & 0.198 & 17.45 & 0.484 & 0.190 & 18.99 & 0.592 & 0.145 \\
     & super\_markets & 47 & 6.22 & 0.109 & 0.807 & 7.28 & 0.223 & 0.643 & 4.25 & 0.112 & 0.758 & 8.30 & 0.275 & 0.609 \\
     & tractor & 41 & 9.25 & 0.208 & 0.382 & 17.64 & 0.513 & 0.122 & 16.58 & 0.492 & 0.145 & 18.71 & 0.650 & 0.094 \\
    NeRFBusters & car & 42 & 12.66 & 0.317 & 0.301 & 15.60 & 0.394 & 0.212 & 15.71 & 0.397 & 0.201 & 16.57 & 0.471 & 0.201 \\
     & century & 50 & 15.98 & 0.366 & 0.116 & 15.04 & 0.331 & 0.120 & 15.72 & 0.349 & 0.112 & 16.54 & 0.393 & 0.106 \\
     & flowers & 44 & 14.06 & 0.240 & 0.288 & 15.80 & 0.292 & 0.187 & 15.66 & 0.280 & 0.190 & 16.21 & 0.330 & 0.184 \\
     & garbage & 45 & 12.20 & 0.439 & 0.342 & 14.04 & 0.437 & 0.311 & 18.39 & 0.578 & 0.182 & 19.84 & 0.645 & 0.159 \\
     & picnic & 40 & 16.32 & 0.356 & 0.104 & 13.87 & 0.321 & 0.140 & 16.50 & 0.355 & 0.094 & 15.01 & 0.332 & 0.115 \\
     & roses & 46 & 15.29 & 0.249 & 0.240 & 16.89 & 0.323 & 0.096 & 17.13 & 0.318 & 0.096 & 18.22 & 0.420 & 0.074 \\
     & table & 36 & 13.28 & 0.687 & 0.143 & 24.48 & 0.773 & 0.071 & 20.38 & 0.642 & 0.151 & 20.48 & 0.728 & 0.093 \\
    TnT & Family & 42 & 11.31 & 0.185 & 0.457 & 19.33 & 0.485 & 0.172 & 19.44 & 0.498 & 0.159 & 22.15 & 0.629 & 0.128 \\
     & Francis & 42 & 12.99 & 0.253 & 0.427 & 14.71 & 0.287 & 0.318 & 17.55 & 0.393 & 0.189 & 18.90 & 0.429 & 0.174 \\
     & Horse & 42 & 6.69 & 0.150 & 0.333 & 17.18 & 0.388 & 0.137 & 18.06 & 0.447 & 0.119 & 17.96 & 0.555 & 0.119 \\
    \bottomrule
  \end{tabular}}
\end{table*}

\begin{table}[htbp]
  \centering
  \caption{\textbf{Per-scene cleanup diagnostics.}
Widened-field silhouette leakage and retained Gaussian fraction.}
  \label{tab:supp_cleanup}
  \setlength{\tabcolsep}{4pt}
  \resizebox{\columnwidth}{!}{%
  \begin{tabular}{ll ccc ccc}
    \toprule
    & & \multicolumn{3}{c}{Leakage $\downarrow$} & \multicolumn{3}{c}{Retained $\downarrow$} \\
    \cmidrule(lr){3-5}\cmidrule(lr){6-8}
    Dataset & Scene & FS & GCut$^\dag$ & Ours & FS & GCut$^\dag$ & Ours \\
    \midrule
    DL3DV & car\_set2 & 0.023 & 0.012 & 0.037 & 25.8\% & 32.9\% & 32.6\% \\
     & courthouses & 0.023 & 0.012 & 0.037 & 7.2\% & 8.2\% & 7.6\% \\
     & fountain & 0.046 & 0.000 & 0.047 & 1.1\% & 1.5\% & 1.8\% \\
     & gardens & 0.006 & 0.000 & 0.082 & 2.5\% & 3.7\% & 5.8\% \\
     & red\_brick\_build & 0.019 & 0.026 & 0.016 & 29.2\% & 41.9\% & 29.3\% \\
     & redcar & 0.008 & 0.001 & 0.020 & 24.2\% & 28.3\% & 31.6\% \\
     & super\_markets & 0.098 & 0.003 & 0.221 & 11.3\% & 4.6\% & 42.7\% \\
     & tractor & 0.002 & 0.001 & 0.017 & 18.0\% & 20.4\% & 21.0\% \\
    NeRFBusters & car & 0.062 & 0.013 & 0.105 & 3.7\% & 4.7\% & 9.0\% \\
     & century & 0.131 & 0.040 & 0.125 & 22.2\% & 33.3\% & 44.9\% \\
     & flowers & 0.076 & 0.003 & 0.038 & 30.8\% & 33.0\% & 31.9\% \\
     & garbage & 0.029 & 0.013 & 0.129 & 6.6\% & 15.8\% & 17.0\% \\
     & picnic & 0.035 & 0.012 & 0.099 & 4.4\% & 9.3\% & 15.7\% \\
     & roses & 0.026 & 0.001 & 0.023 & 31.8\% & 40.0\% & 38.6\% \\
     & table & 0.008 & 0.004 & 0.008 & 6.9\% & 9.0\% & 6.2\% \\
    TnT & Family & 0.116 & 0.002 & 0.013 & 9.3\% & 8.7\% & 10.0\% \\
     & Francis & 0.037 & 0.017 & 0.095 & 26.6\% & 57.9\% & 53.3\% \\
     & Horse & 0.003 & 0.000 & 0.081 & 7.1\% & 10.8\% & 12.1\% \\
    \bottomrule
  \end{tabular}}
\end{table}

Tables~\ref{tab:supp_standard} and~\ref{tab:supp_zoomed} show that the
aggregate improvements are consistent across scenes. At both framings, \method
achieves higher PSNR than FlashSplat on 16 of 18 scenes and than all-mask
$\text{GaussianCut}^{\dagger}$ on 15 of 18, and outperforms matched three-mask
GaussianCut on all 18 scenes.
Its mean PSNR is $19.03$\,dB at the captured framing and $17.83$\,dB after
widening, with per-scene standard deviations of $2.83$ and $2.93$\,dB,
respectively.

The comparison with the uncleaned reconstruction shows the intended trade-off
more clearly. \method improves PSNR on 9 of 18 scenes at the captured framing,
but on 16 of 18 under widened evaluation. This mirrors the aggregate results:
cleanup largely preserves the appearance of the original reconstruction in
the captured views while substantially improving robustness to reframing.

\paragraph{Median PSNR}
Table~\ref{tab:supp_median} reports median alongside mean PSNR. The ranking
of the methods is unchanged at both framings. Medians are less sensitive to
individual scenes such as \emph{super markets}, where
$\text{GaussianCut}^{\dagger}$ reaches only $4.26$\,dB. The mean margin of
\method over $\text{GaussianCut}^{\dagger}$ is $1.76/1.12$\,dB
(standard/widened) over all 18 scenes and $1.49/0.95$\,dB when
\emph{super markets} is excluded.

\begin{table}[t]
  \centering
  \caption{\textbf{Mean and median masked PSNR} over 18 scenes. Std: captured
  framing; $6\times$: widened framing.}
  \label{tab:supp_median}
  \setlength{\tabcolsep}{4pt}
  \fitcol{%
  \begin{tabular}{lcccc}
    \toprule
    & \multicolumn{2}{c}{Mean} & \multicolumn{2}{c}{Median} \\
    \cmidrule(lr){2-3}\cmidrule(lr){4-5}
    Method & Std & $6\times$ & Std & $6\times$ \\
    \midrule
    Uncleaned \gs & \textbf{19.35} & 12.84 & \textbf{19.43} & 12.71 \\
    FlashSplat & 16.44 & 16.29 & 16.56 & 16.47 \\
    GaussianCut & 15.28 & 15.10 & 16.40 & 16.02 \\
    GaussianCut$^\dagger$ & 17.27 & 16.71 & 17.71 & 17.30 \\
    \textbf{\method} & 19.03 & \textbf{17.83} & 19.31 & \textbf{18.33} \\
    \bottomrule
  \end{tabular}}
\end{table}

\section{Choice of Zoom Factor}

The widened protocol reduces focal length at a fixed pose and a fixed output
resolution, so the target occupies fewer pixels and the masked metrics are
computed over a smaller region. Averaged over the 18 scenes, the evaluated
target pixel count $Z_v$ is $5.8\times10^{5}$ at the captured framing and
$2.3\times10^{4}$ at $6\times$, a ratio of $24.6$. Pure area scaling
would predict $36$; the difference is that several targets are already
clipped by the captured frame, so $Z_v$ at $1\times$ is smaller than the
full silhouette would imply.

The widened-field protocol divides focal length by six. Table~\ref{tab:supp_zoom}
repeats it at $2\times$ and $4\times$ on the same 18 scenes and the same
exported assets, so the only thing that varies is the focal length.

\begin{table}[t]
  \centering
  \caption{\textbf{PSNR across widening factors.}
Masked PSNR at captured framing and at $2\times$, $4\times$, and $6\times$
widening over the same 18 exported assets. \method ranks first at each tested
widened setting.}
  \label{tab:supp_zoom}
  \setlength{\tabcolsep}{5pt}
  \fitcol{%
  \begin{tabular}{lcccc}
    \toprule
    Method & $1\times$ & $2\times$ & $4\times$ & $6\times$ \\
    \midrule
    Uncleaned \gs & \textbf{19.35} & 18.72 & 14.72 & 12.84 \\
    FlashSplat & 16.44 & 16.65 & 16.59 & 16.29 \\
    GaussianCut$^\dag$ & 17.27 & 17.43 & 17.15 & 16.71 \\
    \textbf{\method} & 19.03 & \textbf{19.31} & \textbf{18.73} & \textbf{17.83} \\
    \bottomrule
  \end{tabular}}
\end{table}

\method ranks first at every widened factor, while FlashSplat and
GaussianCut keep the same relative order at $2\times$, $4\times$, and
$6\times$. The result is therefore stable to the chosen widening factor. The
uncleaned scene follows the complementary trend: it leads at captured framing,
is overtaken by $2\times$, and ranks last by $4\times$ as retained
off-object structure progressively enters the widened view.

Table~\ref{tab:supp_zoom_leak} reports leakage at each widening factor.
Leakage falls with widening for every method because the target and the
retained structure near it occupy a smaller share of the widened image while
newly exposed regions are mostly empty. The ranking is stable across factors.
$\text{GaussianCut}^{\dagger}$ has the lowest leakage at every factor,
followed by matched GaussianCut and FlashSplat, while \method has the highest
leakage among the cleanup methods at every factor, consistent with its
appearance-preserving operating point (Section~\ref{sec:supp_frontier}).

\begin{table}[t]
  \centering
  \caption{\textbf{Leakage across widening factors.}
  Mean unthresholded silhouette leakage over the same 18 exported assets at
  captured framing ($1\times$) and at $2\times$, $4\times$, and
  $6\times$ widening.}
  \label{tab:supp_zoom_leak}
  \setlength{\tabcolsep}{5pt}
  \fitcol{%
  \begin{tabular}{lcccc}
    \toprule
    Method & $1\times$ & $2\times$ & $4\times$ & $6\times$ \\
    \midrule
    Uncleaned \gs & 0.983 & 0.896 & 0.813 & 0.803 \\
    FlashSplat & 0.141 & 0.085 & 0.049 & 0.042 \\
    GaussianCut & 0.089 & 0.047 & 0.022 & 0.016 \\
    GaussianCut$^\dag$ & \textbf{0.055} & \textbf{0.038} & \textbf{0.015} & \textbf{0.009} \\
    \method & 0.206 & 0.123 & 0.075 & 0.066 \\
    \bottomrule
  \end{tabular}}
\end{table}

\section{Baseline Evaluation Setup}
\label{sec:supp_baselines}

\paragraph{FlashSplat.}
We evaluate FlashSplat using its exported object with removal slackness $0$,
the default setting of the reference implementation. It receives the same three
selected masks as \method's 3D filtering stage and reaches $16.44/16.29$\,dB
standard/widened PSNR at leakage $0.141$ and $14.9\%$ retention. Its lower
leakage does not come from empty outputs: the mean rendered opacity inside the
target mask is $0.94$, against $0.96$ for \method, and no scene falls below
$0.86$. The PSNR gap instead comes from target surfaces that are partly
removed or smeared, most visibly on \emph{Francis}, \emph{garbage}, and
\emph{picnic}.

The removal-slackness sweep in Section~\ref{sec:supp_frontier} was run on a
different three-view subset of the same curated masks, where FlashSplat at
slackness $0$ reaches $17.34/16.79$\,dB, leakage $0.148$, and $17.4\%$
retention. On that subset, changing the slackness to $-0.5$ increases
retention to $61.7\%$ and leakage to $0.446$, which shows how sensitive
FlashSplat's retained asset is to this parameter.

\paragraph{GaussianCut.}
We evaluate the final graph-cut output under two supervision settings. The
matched configuration receives the same three Pareto-selected masks as
\method. The second configuration, denoted
$\text{GaussianCut}^{\dagger}$, receives all available capture masks,
including masks at scored cameras, and serves as an all-mask reference.
Both configurations are evaluated with the same protocol over the same
18 scenes.

\paragraph{Additional methods.}
We restrict the quantitative comparison to methods that produce a directly
renderable Gaussian subset under comparable preprocessing. LUDVIG
\cite{ludvig} applies a fixed Gaussian-budget pruning stage before semantic
uplifting, so its output is not directly comparable to cleanup applied to the
original pretrained scene. SAGOnline \cite{sagonline} is not included because
a directly comparable retained-asset result is not available under this
evaluation protocol. Gaussian Grouping and
Gaga represent object identity through learned or rasterized semantic labels
rather than an exported pruned Gaussian subset; in our Gaga evaluation, the
exported scene retains all input Gaussians. We therefore discuss these methods
as related approaches rather than include them in the retained-asset table.

Recent Gaussian-level segmentation methods also lift multiview 2D masks to
Gaussians, but under a different setting. Click-Gaussian \cite{clickgaussian}
trains contrastive per-Gaussian features from dense SAM masks of all training
views. Trace3D \cite{supp_trace3d} densifies and prunes Gaussians during training
to obtain consistent instance segmentation. SAGOnline \cite{sagonline}
propagates masks across rendered views with a video segmenter and fuses
per-Gaussian labels, with per-scene refinement. Specifically, none of them 
exports a subset of the original pretrained Gaussians from three masks without 
retraining. So, we treat them as related work rather than as baselines.

\paragraph{Masks.}
Target masks were produced once during data curation with SAM~3 and manual
prompts, checked only for coverage of the intended object, and reused
unchanged in every run and for every method. Importantly, no mask was edited for any
particular method. \method, matched GaussianCut, and FlashSplat use the same three
selected training-view masks. Evaluation masks are never available to the
selector or to the scalarization.

\section{Selected Views against Random Views}
\label{sec:supp_random}

Table~\ref{tab:selector} expands the selector comparison from the main text.
Both arms use the same two-of-three vote and evaluator; only the three input
views differ. The random arm samples three training views uniformly with five
seeds per scene and averages the seeds within each scene before pairing.

\begin{table}[t]
  \centering
  \caption{\textbf{Pareto-selected views against random views at the same budget.}
Both arms use the same vote and evaluator and differ only in the three input
views. Five seeds are averaged within each scene, with paired results reported
over the 17 scenes with defined outputs for both arms. ``Wins'' counts scenes favoring
selection, and $p$ is a two-sided Wilcoxon signed-rank test. Std: captured
framing; $6\times$: widened framing.}
  \label{tab:selector}
  \setlength{\tabcolsep}{3.5pt}
  \fitcol{%
  \begin{tabular}{lccrcc}
    \toprule
    & Pareto & Random & $\Delta$ & Wins & $p$ \\
    \midrule
    PSNR $\uparrow$                  & \textbf{19.52} & 17.96 & $+1.56$ & 15/17 & 0.0005 \\
    SSIM $\uparrow$                  & \textbf{0.657} & 0.620 & $+0.037$ & 15/17 & 0.0004 \\
    LPIPS $\downarrow$               & \textbf{0.166} & 0.202 & $-0.036$ & 15/17 & 0.0011 \\
    PSNR (6$\times$) $\uparrow$    & \textbf{18.39} & 17.39 & $+1.00$ & 14/17 & 0.0150 \\
    SSIM (6$\times$) $\uparrow$    & \textbf{0.514} & 0.481 & $+0.033$ & 14/17 & 0.0038 \\
    LPIPS (6$\times$) $\downarrow$ & \textbf{0.141} & 0.170 & $-0.029$ & 13/17 & 0.0129 \\
    \midrule
    Leakage $\downarrow$             & 0.197 & \underline{0.154} & $+0.043$ & 5/17 & 0.0093 \\
    Retained $\downarrow$            & 21.7\% & \underline{18.8\%} & $+2.9$ & 4/17 & 0.0067 \\
    \bottomrule
  \end{tabular}}
\end{table}

Across the 17 scenes with defined photometrics in both arms, selected views
improve all six image-quality metrics: $+1.56$\,dB standard PSNR and
$+1.00$\,dB widened PSNR, together with corresponding SSIM and LPIPS gains.
Random subsets produce lower leakage ($0.154$ versus $0.197$) and lower
retention ($18.8\%$ versus $21.7\%$); these compactness differences are also
significant ($p=0.0093$ and $p=0.0067$, respectively). The comparison
therefore shows a clear operating-point shift: selected views preserve more of
the target while random views prune more aggressively.

On \emph{super markets}, three of the five random subsets retain no Gaussians
and therefore have undefined photometrics, whereas the selected subset remains
non-empty. Our selected main configuration produces a non-empty asset on all 18
scenes.

\paragraph{Do the selected views help other filters?}
Table~\ref{tab:supp_selfilter} applies the same selected masks and the first
random three-view set (seed 0) to GaussianCut and FlashSplat. Neither baseline
benefits from the selected views. In contrast, \method gains $0.80$\,dB in standard PSNR from selection
($p=0.017$), while its leakage and retention rise. The selected masks are
therefore not uniformly better inputs for every lifting method. They are
useful specifically for the vote, where reliable views make agreement
meaningful. Selection also improves robustness: with random views, a strict
three-of-three vote empties \emph{car} and \emph{super markets} for two of three
seeds, whereas the selected views never produce an empty asset.

\begin{table}[t]
  \centering
  \caption{\textbf{Selected against random masks for each filter.}
  Each filter receives either the three selected masks or the seed-0 random
  three-view set. Std: captured framing; $6\times$: widened framing. $p$
  is a two-sided Wilcoxon test on standard PSNR. The \method rows cover 17
  scenes because the random set empties \emph{super markets}.}
  \label{tab:supp_selfilter}
  \setlength{\tabcolsep}{3pt}
  \fitcol{%
  \begin{tabular}{llccccc}
    \toprule
    Filter & Masks & PSNR (std) & PSNR ($6\times$) & Leak & Retained & $p$ \\
    \midrule
    GaussianCut & selected & 15.28 & 15.10 & 0.089 & 17.2\% & \multirow{2}{*}{0.20} \\
                & random   & 15.92 & 15.72 & 0.085 & 18.2\% & \\
    \midrule
    FlashSplat  & selected & 16.44 & 16.29 & 0.141 & 14.9\% & \multirow{2}{*}{0.15} \\
                & random   & 16.91 & 16.56 & 0.149 & 16.0\% & \\
    \midrule
    \method     & selected & 19.52 & 18.39 & 0.197 & 21.7\% & \multirow{2}{*}{0.017} \\
                & random   & 18.72 & 18.02 & 0.158 & 19.3\% & \\
    \bottomrule
  \end{tabular}}
\end{table}

\paragraph{Camera variation.}
Because the selector uses camera baseline and viewing-direction diversity, its
benefit could depend on how much the capture trajectory varies. We measure the
angular spread of each scene as the mean pairwise viewing-direction angle over
all training cameras with candidate masks, and the selection gain as the
paired PSNR difference between selected and random views from
Table~\ref{tab:selector}. The gain does not follow the spread (Spearman
$\rho=0.24$ at captured framing and $0.13$ widened). The only scene where
selection loses at both framings is \emph{red brick build}
($-0.67/-0.70$\,dB), which also has the narrowest spread ($31^\circ$).

\section{Role of the Pareto Front}

The dominance lemma guarantees that any coordinate-wise non-decreasing utility
attains an optimum on the exact Pareto front of the enumerated subsets of
$\mathcal{V}_r$. Restricting selection to that front therefore preserves the
optimal value of the weighted Chebyshev criterion over the search space the
selector actually uses, while keeping the trade-offs among $A$, $B$,
$O$, and $T$ explicit.

This is particularly useful in our setting because the goal is not to collect
as many observations as possible, but to identify a small subset that provides
strong and complementary evidence. With only three selected views, redundancy
is costly: geometrically or semantically similar observations spend the limited
view budget repeating much of the same evidence. The multi-objective
formulation instead allocates this budget deliberately across mask reliability,
camera baseline, viewing-angle diversity, and masked-support complementarity
before a single operating point is chosen. The comparison with random
three-view subsets in Table~\ref{tab:selector} shows the practical benefit of
selecting the three-view evidence rather than choosing it arbitrarily.

The front also keeps the non-dominated trade-offs available when the relative
importance of the objectives changes. Table~\ref{tab:supp_weights} shows that
reweighting the same objective space moves the selected operating point while
maintaining similar overall fidelity. Exact Pareto construction becomes more
expensive as $k$ increases because the number of feasible subsets grows
combinatorially, as reflected in Table~\ref{tab:supp_budget}. We therefore use
$k=3$, where exact Pareto selection remains practical while providing the
redundancy required for two-of-three voting. Because the direct argmin returns
the same subset (Section~\ref{sec:supp_reduction}), larger budgets do not
require the explicit front.

\paragraph{Pareto selection against simpler selectors.}
Table~\ref{tab:supp_simple} compares the Pareto selector with four simpler
selectors that use the same reduced pool, vote, and evaluator: a weighted sum of
the same normalized objectives and weights, a greedy selector maximizing the
same weighted objective, the three most reliable views, and the most reliable
view followed by farthest-point sampling of camera centers. They choose a
different subset from Pareto selection on 11, 16, 11, and 11 scenes,
respectively. All four stay within $0.08$\,dB of Pareto selection in
standard PSNR. Pareto selection has lower leakage than
top-three reliability ($p=0.016$), whereas reliability with farthest-point
sampling has higher widened PSNR ($p=0.044$) and lower leakage ($p=0.001$)
than Pareto selection on these scenes. It is worth noting however that the 
Pareto formulation adds an explicit, tunable trade-off among reliability, 
baseline, and support, instead of a single hard-coded heuristic, which, as 
we show, does not generalize well.

As an exploratory check outside the 18 benchmark scenes, we ran the selectors
on 15 further scenes from the same benchmarks that have all required inputs.
There, reliability with farthest-point sampling nearly empties one scene
(\emph{m60}: 1{,}111 retained Gaussians and $6.0$\,dB, against 539{,}557
Gaussians for Pareto selection), because it ignores masked support.
Replacing the mean pairwise baseline in $B(S)$ by the smallest pairwise
baseline (a max--min form) changes the selection on 9 of 15 scenes and raises
standard PSNR by $0.13$\,dB ($p=0.012$) by removing near-duplicate views.
We report this as an indication of how the objectives can be tuned.

\begin{table}[t]
  \centering
  \caption{\textbf{Pareto selection against simpler selectors} on 18 scenes,
  with the same reduced pool, two-of-three vote, and evaluator. Std: captured
  framing; $6\times$: widened framing.}
  \label{tab:supp_simple}
  \setlength{\tabcolsep}{3pt}
  \fitcol{%
  \begin{tabular}{lccccc}
    \toprule
    Selector & PSNR (std) & PSNR ($6\times$) & Leak (std) & Leak ($6\times$) & Retained \\
    \midrule
    Pareto (main) & 19.03 & 17.83 & 0.206 & 0.066 & 22.8\% \\
    Weighted sum & 19.00 & 17.71 & 0.193 & 0.060 & 21.3\% \\
    Greedy & 18.96 & 17.72 & 0.194 & 0.060 & 21.5\% \\
    Top-3 reliability & 19.07 & 17.70 & 0.224 & 0.077 & 23.1\% \\
    Reliability + farthest point & 19.02 & 18.15 & 0.166 & 0.032 & 20.6\% \\
    \bottomrule
  \end{tabular}}
\end{table}

\paragraph{Geometry and complementarity of the selected views.}
Selected views are geometrically closer to one another than random triplets:
their median smallest pairwise viewing angle is $7.6^\circ$ against
$34^\circ$ for random views ($p=0.014$), and in 10 of 18 scenes two of the
three selected views lie within $10^\circ$. This follows from the weighting,
in which the weakest-link reliability $A$ dominates and $B$ and $T$ are
normalized within a small reduced pool. In terms of target evidence, the
selected views remain complementary in absolute terms: in the median scene,
$90\%$ of the in-mask SfM points are seen by only one of the three views, and
whichever view is added third still contributes at least $22\%$ new points.
They are not more complementary than random triplets, however (median $O$ of
$0.943$ against $0.987$), and \emph{Francis}, \emph{garbage}, and
\emph{picnic} are near-redundant. We therefore describe the selected views as
reliable views that each contribute non-redundant target evidence.

\section{Fidelity--Cleanup Frontier}
\label{sec:supp_frontier}

\begin{figure*}[t]
  \centering
  \includegraphics[width=\textwidth]{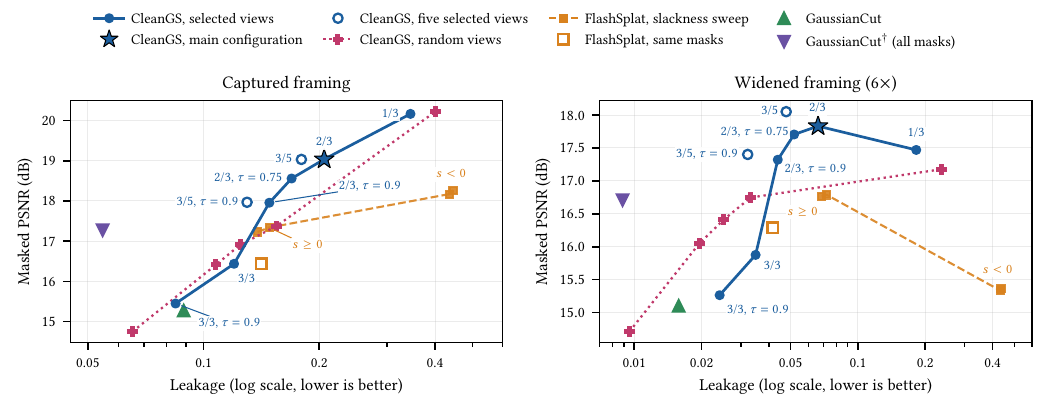}
  \caption{\textbf{Leakage--PSNR frontier} over 18 scenes at captured (left)
  and $6\times$ widened (right) framing, with leakage on a log scale. Labels
  give the vote ($k$-of-$n$) and the footprint threshold $\tau$. The
  FlashSplat sweep varies the removal slackness $s$ on a different three-view
  subset; the hollow square is FlashSplat at $s=0$ with our selected masks.
  Random-view points use the same filters and average seeds 0--2.}
  \Description{Two plots of masked PSNR against silhouette leakage on a log
  scale, with curves for CleanGS with selected views, CleanGS with random
  views, and the FlashSplat slackness sweep, and single points for FlashSplat
  with the selected masks and for GaussianCut.}
  \label{fig:supp_frontier}
\end{figure*}

Masked PSNR rewards keeping the target, while leakage and retention reward
pruning, so a single operating point cannot show whether a method improves the
fidelity--cleanup trade-off or only moves along it.
Figure~\ref{fig:supp_frontier} therefore plots PSNR against leakage. With the
same three selected masks, we vary \method through the vote threshold (one-,
two-, or three-of-three) and the footprint threshold $\tau$
(Section~\ref{sec:supp_footprint}); hollow circles use five selected masks. We
vary FlashSplat through its removal slackness, and run random-view \method with
the same filters on random three-view subsets (seeds 0--2). All points are
means over the 18 scenes, and the $k\ge4$ selections use the direct argmin.

At captured framing and the same three-mask budget, the \method curve lies
above every matched-budget baseline point. At the leakage of FlashSplat's
slackness-$0$ point on the sweep subset ($0.148$), it reaches
$17.96$\,dB, $0.62$\,dB higher, and FlashSplat with our selected masks
($0.141$, $16.44$\,dB) lies further below the curve. At low leakage,
$\tau=0.9$ with a three-of-three vote dominates matched GaussianCut
($0.085/15.45$ against $0.089/15.28$). Only the all-mask
$\text{GaussianCut}^{\dagger}$, which uses masks at the scored cameras, stays
above the curve at low leakage. Under widened framing, $\tau=0.9$ with
two-of-three voting has lower leakage ($0.044$) and at least $0.5$\,dB
higher PSNR than every FlashSplat slackness setting, and FlashSplat with our
selected masks is $1.0$\,dB lower at similar leakage ($0.042$). With five
masks, three-of-five voting matches the standard PSNR of the main configuration
($19.03$\,dB) at lower leakage ($0.180$ against $0.206$), and with
$\tau=0.9$ it reaches $17.97$\,dB at $0.130$.

The random-view curve isolates the contribution of selection under the same
filter. Selected views are $0.6$--$1.1$\,dB higher than random views for
leakage between $0.15$ and $0.35$, while below about $0.13$ random views
are up to $0.3$\,dB higher; the curves cross near $0.14$. Selection
therefore lifts the frontier in the moderate-leakage region that contains the
main operating point, rather than only moving to a more retentive point, while
under very aggressive cleanup the difference is mainly a shift in operating
point. Paired on the same footprint filter, selection still raises PSNR
together with leakage ($+1.54$\,dB at $\tau=0.9$, $p=0.001$).

\section{Selector Ablations}
\label{sec:supp_selector_ablations}

The following ablations vary one component of the selector while keeping the
remaining pipeline fixed. Each arm is paired against a rerun of the main
$k{=}3$, 2-of-3 configuration inside the same ablation family. That
reference covers 17 of the 18 benchmark scenes; the NeRFBusters \emph{car}
scene is excluded because its ablation runs did not complete across the arms.
Its leakage on those 17 scenes is $0.204$, measured at the captured
($1\times$) framing on the full evaluation masks, the same protocol as
Table~\ref{tab:main}, whose $0.206$ is the corresponding 18-scene mean.
Arms that did not complete on all 17 scenes are reported on their own completed
subsets.

\emph{Reliability aggregation.} The reliability objective $A(S)$ can
aggregate the per-view scores of a subset by their minimum, as in the main
text, or by their mean. We report the ablations for both settings.
Tables~\ref{tab:supp_objectives}--\ref{tab:supp_budget} use the mean, and each
arm is paired with a reference run that also uses the mean.
Tables~\ref{tab:supp_sigma} and~\ref{tab:supp_budget_min} use the minimum and
report the sensitivity to the occupancy width $\sigma$ and the budget and
threshold study on all 18 scenes.

\begin{table}[t]
  \centering
  \caption{\textbf{Sensitivity to the occupancy width $\sigma$} with minimum
  aggregation over 18 scenes; all other settings are fixed. ``Changed'' counts
  the scenes whose three selected views differ from those at the main
  $\sigma=0.15$; the other scenes keep the main result. Std: captured
  framing; $6\times$: widened framing.}
  \label{tab:supp_sigma}
  \setlength{\tabcolsep}{4pt}
  \fitcol{%
  \begin{tabular}{lccccc}
    \toprule
    $\sigma$ & Changed & PSNR (std) & PSNR ($6\times$) & Leak (std) & Leak ($6\times$) \\
    \midrule
    0.05 & 5/18 & 19.14 & 17.76 & 0.206 & 0.073 \\
    0.10 & 1/18 & 19.05 & 17.84 & 0.207 & 0.066 \\
    0.15 (main) & -- & 19.03 & 17.83 & 0.206 & 0.066 \\
    0.20 & 2/18 & 19.05 & 17.70 & 0.205 & 0.072 \\
    0.30 & 3/18 & 19.04 & 17.96 & 0.206 & 0.065 \\
    \bottomrule
  \end{tabular}}
\end{table}

Across $\sigma\in[0.05,0.30]$, the selected views differ from the main
selection on at most 5 of 18 scenes
and mean standard PSNR changes by at most $0.11$\,dB, so the selection is
stable over this range.

\begin{table}[t]
  \centering
  \caption{\textbf{Sensitivity to zeroing individual scalarization weights.}
Each row sets one Chebyshev weight to zero and renormalizes the remaining
weights while retaining the four-objective Pareto front, thereby measuring
the influence of each weight. Mean aggregation. Std: captured framing; $6\times$: widened framing.}
  \label{tab:supp_objectives}
  \setlength{\tabcolsep}{3.5pt}
  \fitcol{%
  \begin{tabular}{lrrrr}
  \toprule
  Weight zeroed &
  $\Delta$PSNR &
  $\Delta$PSNR$_{6\times}$ &
  Median $\Delta_{6\times}$ &
  Leakage $\downarrow$ \\
  \midrule
  Main        & --    & --    & --    & 0.204 \\
  without $A$ & -0.48 & -0.07 & +0.01 & 0.163 \\
  without $B$ & +0.13 & +0.09 & +0.00 & 0.204 \\
  without $O$ & -0.08 & -0.16 & -0.05 & 0.218 \\
  without $T$ & -0.05 & -0.22 & -0.07 & 0.211 \\
  \bottomrule
  \end{tabular}}
\end{table}

Table~\ref{tab:supp_objectives} examines how much each scalarization weight
influences selection at the main weighting. The Pareto front is unchanged in
every row, so these are weight ablations and not objective ablations. Zeroing
the weight on $A$ reduces standard PSNR by $0.48$\,dB and changes the
selected subset on 14 of 17 scenes, making mask reliability the strongest
individual contributor at this operating point. Zeroing $B$ changes 10 of 17
selections, with a small $+0.13$\,dB change in standard PSNR. Zeroing
$O$ or $T$ carries a smaller but consistent cost: standard PSNR falls by
$0.08$ and $0.05$\,dB and widened PSNR by $0.16$ and
$0.22$\,dB, with negative medians in both cases, while leakage rises from
$0.204$ to $0.218$ and $0.211$, respectively. The diversity terms
therefore contribute least at the main weighting;
their effect is largest under widened evaluation, where redundant supporting
views are most exposed.

The main weighting deliberately emphasizes $A$ and $B$. Mask reliability
directly controls the quality of the semantic evidence entering the vote,
while camera baseline provides a strong geometric cue for avoiding redundant
observations. The support-overlap and viewing-angle objectives provide
additional notions of diversity, but receive smaller weights because their
informativeness depends more strongly on the available sparse tracks and
capture trajectory.

\begin{table}[t]
  \centering
  \caption{\textbf{Chebyshev weight sensitivity.}
The main weights are $(A,B,O,T)=(0.45,0.30,0.15,0.10)$. Each variant changes
the relative emphasis placed on the four objectives. Mean aggregation. Std: captured framing; $6\times$: widened framing.}
  \label{tab:supp_weights}
  \setlength{\tabcolsep}{3.5pt}
  \fitcol{%
  \begin{tabular}{lrrrrrr}
    \toprule
    Weights & $n$ & $\Delta$PSNR & $\Delta$PSNR$_{6\times}$ & Med. $\Delta_{6\times}$ & Leak $\downarrow$ & Wins \\
    \midrule
    Main & 17 & -- & -- & -- & 0.204 & -- \\
    \midrule
    equal, $0.25$ each & 17 & $+0.02$ & $-0.03$ & $+0.00$ & 0.192 & 5/17 \\
    $B$ heavy, $0.40$ & 17 & $+0.14$ & $+0.17$ & $+0.08$ & 0.188 & 9/17 \\
    $O$ heavy, $0.40$ & 17 & $-0.02$ & $+0.07$ & $+0.00$ & 0.188 & 5/17 \\
    $T$ heavy, $0.40$ & 17 & $+0.13$ & $+0.06$ & $+0.00$ & 0.191 & 7/17 \\
    $O,T$ heavy, $0.35$ & 17 & $+0.09$ & $+0.48$ & $+0.00$ & 0.180 & 8/17 \\
    \bottomrule
  \end{tabular}}
\end{table}

Table~\ref{tab:supp_weights} shows that performance remains stable across a
broad range of weightings: standard PSNR varies by at most $0.14$\,dB across
the tested settings. Increasing the emphasis on $O$ and $T$ changes the
selected subsets and shifts the macro operating point under widened evaluation:
the overlap-and-angle-heavy setting improves widened PSNR by $0.48$\,dB and
lowers leakage from $0.204$ to $0.180$. That gain is not uniform across
scenes, however: the same arm improves widened PSNR on 8 of the 17 scenes, with
a median per-scene change of $0.00$\,dB. Thus, $A$ and $B$ determine
most selections at the main weights, while larger weights on $O$ and $T$
expose alternative operating points with different fidelity--cleanup
trade-offs, indicating room for further tuning.

\begin{table}[t]
  \centering
  \caption{\textbf{Per-view reliability score.}
The main configuration uses a Gaussian-shaped occupancy score with
$r_0{=}0.30$ and $\sigma{=}0.15$; each variant changes the occupancy target
or decay profile while leaving the remaining pipeline fixed. Mean aggregation. Std: captured framing; $6\times$: widened framing.}
  \label{tab:supp_coverage}
  \setlength{\tabcolsep}{3.5pt}
  \fitcol{%
  \begin{tabular}{lrrrrrr}
    \toprule
    Reliability score & $n$ & $\Delta$PSNR & $\Delta$PSNR$_{6\times}$ & Med. $\Delta_{6\times}$ & Leak $\downarrow$ & Wins \\
    \midrule
    Main & 17 & -- & -- & -- & 0.204 & -- \\
    \midrule
    target $0.05$ & 17 & $-1.69$ & $-1.43$ & $-0.18$ & 0.164 & 5/17 \\
    target $0.75$ & 17 & $-0.18$ & $-0.20$ & $+0.00$ & 0.215 & 4/17 \\
    linear, target $1.0$ & 17 & $-0.27$ & $-0.26$ & $+0.00$ & 0.215 & 4/17 \\
    linear decay & 17 & $-0.01$ & $+0.39$ & $+0.00$ & 0.183 & 7/17 \\
    step decay & 17 & $+0.19$ & $+0.56$ & $+0.18$ & 0.155 & 9/17 \\
    \bottomrule
  \end{tabular}}
\end{table}

Table~\ref{tab:supp_coverage} examines the per-view reliability score. Moving
the occupancy target to $r_0=0.05$ reduces standard PSNR by
$1.69$\,dB, while favoring very large masks through $r_0=0.75$ or raw-area
scoring reduces it by $0.18$--$0.27$\,dB. The degradation at both extremes
supports the band-pass design: useful views should contain enough target area
to provide stable support without becoming near-full-frame close-ups.

Performance is less sensitive to the precise decay profile. Linear decay
remains close to the Gaussian score, while a step profile improves the macro
widened PSNR by $0.56$\,dB and lowers leakage, winning on 9 of the 17 scenes
with a median per-scene change of $+0.18$\,dB. The main benefit therefore
comes from favoring an intermediate occupancy range, while the precise shape of
the reliability function offers room for further refinement.

\begin{table}[t]
  \centering
  \caption{\textbf{Selected-view budget and vote threshold.}
Each arm is paired against the main $k{=}3$, 2-of-3 configuration on its
completed scene set. ``Median'' is the median per-scene change in widened PSNR,
and ``Wins'' counts scenes where the arm performs better. Exact Pareto
construction is all-pairs dominance over $\binom{n}{k}$ subsets, which is
$3.4\times10^{4}$ at $k{=}3$ but $4.9\times10^{5}$ at $k{=}4$ and
$5.5\times10^{6}$ at $k{=}5$ for a capped pool of 60. Selection does not depend on the vote threshold, so the
two $k{=}4$ rows differ in $n$ for run-level resource reasons alone and
are not directly comparable with each other. Mean aggregation. Std: captured framing; $6\times$: widened framing.}
  \label{tab:supp_budget}
  \setlength{\tabcolsep}{3.5pt}
  \fitcol{%
  \begin{tabular}{lrrrrrr}
    \toprule
    Configuration & $n$ & $\Delta$PSNR & $\Delta$PSNR$_{6\times}$ & Med. $\Delta_{6\times}$ & Leak $\downarrow$ & Wins \\
    \midrule
    Main & 17 & -- & -- & -- & 0.204 & -- \\
    \midrule
    $k{=}2$, 2-of-2 & 17 & $-2.28$ & $-1.52$ & $-0.18$ & 0.115 & 6/17 \\
    $k{=}4$, 2-of-4 & 15 & $+0.52$ & $+0.21$ & $+0.04$ & 0.220 & 10/15 \\
    $k{=}5$, 2-of-5 & 13 & $+0.57$ & $-0.18$ & $+0.01$ & 0.250 & 9/13 \\
    $k{=}4$, 3-of-4 & 17 & $-1.19$ & $-0.66$ & $-0.03$ & 0.144 & 8/17 \\
    $k{=}5$, 3-of-5 & 13 & $+0.10$ & $+0.54$ & $+0.13$ & 0.147 & 10/13 \\
    \bottomrule
  \end{tabular}}
\end{table}

Table~\ref{tab:supp_budget} studies the interaction between view budget and
vote threshold. The $k{=}2$ arm uses unanimity, so it changes both the view
budget and the relative strictness of the vote; it represents the natural
two-view operating point rather than a budget-only control. Relative to the
main 2-of-3 configuration, it reduces standard and widened PSNR by $2.28$ and
$1.52$\,dB, respectively.

Increasing the number of views while keeping a threshold of two provides only
small median widened gains and increases leakage. Raising the threshold does
not help uniformly: 3-of-4 loses $1.19$ and $0.66$\,dB in standard and
widened PSNR on its 17 scenes. The 3-of-5 configuration is the strongest
larger-budget setting among the completed runs, improving standard PSNR by
$0.10$\,dB and widened PSNR by $0.54$\,dB on its 13-scene subset while
reducing leakage to $0.147$.

Table~\ref{tab:supp_budget_min} reports the same study with minimum aggregation
on all 18 scenes. Arms with $k\ge4$ use the direct argmin. Three-of-five 
voting matches the standard PSNR of the main configuration ($-0.004$\,dB), 
improves widened PSNR by $0.22$\,dB, and lowers leakage at both framings ($0.206$ to $0.180$
captured, $0.066$ to $0.048$ widened). On average, it is therefore a better
operating point, although each arm wins on about half of the scenes. Raising the threshold at a
fixed budget, or using unanimity at $k=2$, lowers leakage at a substantial
cost in fidelity.

\begin{table}[t]
  \centering
  \caption{\textbf{Selected-view budget and vote threshold with minimum
  aggregation} on all 18 scenes. Std: captured framing; $6\times$: widened
  framing.}
  \label{tab:supp_budget_min}
  \setlength{\tabcolsep}{3pt}
  \fitcol{%
  \begin{tabular}{lccccc}
    \toprule
    Configuration & PSNR (std) & PSNR ($6\times$) & Leak (std) & Leak ($6\times$) & Retained \\
    \midrule
    $k{=}3$, 1-of-3 & 20.16 & 17.47 & 0.345 & 0.182 & 29.1\% \\
    $k{=}3$, 2-of-3 (main) & 19.03 & 17.83 & 0.206 & 0.066 & 22.8\% \\
    $k{=}3$, 3-of-3 & 16.44 & 15.88 & 0.120 & 0.035 & 18.6\% \\
    $k{=}2$, 2-of-2 & 16.87 & 16.25 & 0.132 & 0.036 & 19.3\% \\
    $k{=}4$, 3-of-4 & 16.96 & 16.19 & 0.155 & 0.040 & 20.1\% \\
    $k{=}5$, 3-of-5 & 19.03 & 18.05 & 0.180 & 0.048 & 22.4\% \\
    \bottomrule
  \end{tabular}}
\end{table}

The $k{=}3$, 2-of-3 configuration therefore provides a practical operating
point: it is the smallest setting with voting redundancy and supports exact
Pareto selection across all 18 scenes. The 3-of-5 result shows that additional
views with a correspondingly higher threshold can improve the fidelity--cleanup
trade-off.

\section{Per-Scene Effect of Clustering}
\label{sec:supp_cluster}

\begin{table}[t]
  \centering
  \caption{\textbf{Largest-component geometric refinement.}
Macro averages over 18 scenes. Both endpoints use the same eroded evaluation
masks, isolating the effect of the clustering stage; their absolute values are
therefore not directly comparable with the Table~\ref{tab:main} endpoint, which is
evaluated on the full masks. Std: captured framing; $6\times$: widened framing.}
  \label{tab:ablation_cluster}
  \setlength{\tabcolsep}{4pt}
  \fitcol{%
  \begin{tabular}{l ccc ccc c}
    \toprule
    & \multicolumn{3}{c}{Standard} & \multicolumn{3}{c}{Widened} & Leak \\
    \cmidrule(lr){2-4}\cmidrule(lr){5-7}
    Endpoint & PSNR & SSIM & LPIPS & PSNR & SSIM & LPIPS & $\downarrow$ \\
    \midrule
    Vote only & \textbf{19.86} & \textbf{0.658} & \textbf{0.170} & \textbf{18.25} & \textbf{0.516} & \textbf{0.158} & 0.221 \\
    \quad + clustering & 16.02 & 0.478 & 0.281 & 14.97 & 0.370 & 0.303 & \underline{0.115} \\
    \bottomrule
  \end{tabular}}
\end{table}

Table~\ref{tab:ablation_cluster} shows that spatial connectivity provides
useful cleanup signal: retaining the largest HDBSCAN component reduces
silhouette leakage from $0.221$ to $0.115$. The corresponding loss in
photometric fidelity is highly scene dependent. Six scenes lose more than
$3$\,dB in standard PSNR, while the remaining twelve lose at most
$0.48$\,dB.

\begin{table}[htbp]
  \centering
  \caption{\textbf{Per-scene effect of largest-component clustering.}
Scenes are ordered by standard-PSNR change. ``Cl.'' is the number of HDBSCAN
clusters in the voted set, and ``top'' is the fraction of voted Gaussians in
the largest cluster. Values use the same eroded evaluation masks as
Table~\ref{tab:ablation_cluster}. Std: captured framing; $6\times$: widened framing.}
  \label{tab:supp_ablation}
  \setlength{\tabcolsep}{3.5pt}
  \fitcol{%
  \begin{tabular}{l rr r rr rr}
    \toprule
    & \multicolumn{2}{c}{Std PSNR} & & \multicolumn{2}{c}{Leakage} & \multicolumn{2}{c}{Clustering} \\
    \cmidrule(lr){2-3}\cmidrule(lr){5-6}\cmidrule(lr){7-8}
    Scene & vote & +cl. & $\Delta$ & vote & +cl. & Cl. & top \\
    \midrule
    table & 22.29 & 2.72 & -19.57 & 0.226 & 0.012 & 112 & 16\% \\
    tractor & 20.56 & 8.97 & -11.58 & 0.223 & 0.002 & 92 & 9\% \\
    redcar & 19.33 & 8.58 & -10.75 & 0.272 & 0.004 & 39 & 13\% \\
    Horse & 22.32 & 11.69 & -10.63 & 0.070 & 0.005 & 22 & 47\% \\
    car\_set2 & 18.56 & 8.50 & -10.06 & 0.372 & 0.004 & 35 & 16\% \\
    red\_brick\_build & 16.51 & 11.69 & -4.81 & 0.200 & 0.031 & 7 & 88\% \\
    fountain & 21.46 & 20.98 & -0.48 & 0.132 & 0.019 & 34 & 74\% \\
    super\_markets & 10.84 & 10.56 & -0.28 & 0.357 & 0.218 & 7 & 67\% \\
    gardens & 19.90 & 19.67 & -0.23 & 0.081 & 0.005 & 56 & 57\% \\
    Family & 24.76 & 24.61 & -0.16 & 0.062 & 0.038 & 2 & 96\% \\
    garbage & 22.40 & 22.25 & -0.15 & 0.345 & 0.308 & 3 & 97\% \\
    car & 19.14 & 19.06 & -0.08 & 0.224 & 0.193 & 4 & 98\% \\
    courthouses & 23.36 & 23.29 & -0.07 & 0.054 & 0.044 & 2 & 100\% \\
    flowers & 17.20 & 17.14 & -0.06 & 0.261 & 0.186 & 2 & 100\% \\
    century & 21.07 & 21.03 & -0.04 & 0.456 & 0.411 & 7 & 95\% \\
    Francis & 19.26 & 19.23 & -0.03 & 0.182 & 0.137 & 10 & 94\% \\
    picnic & 17.94 & 17.91 & -0.03 & 0.256 & 0.250 & 3 & 98\% \\
    roses & 20.54 & 20.54 & -0.00 & 0.208 & 0.203 & 4 & 98\% \\
    \bottomrule
  \end{tabular}}
\end{table}

Table~\ref{tab:supp_ablation} shows that the largest fidelity losses are
concentrated in scenes where the voted target is distributed across multiple
density components. For example, the largest component contains only
$9\%$, $13\%$, and $16\%$ of the retained Gaussians on \emph{tractor},
\emph{redcar}, and \emph{table}, respectively, so retaining a single component
removes substantial target structure. In contrast, clustering has little
effect on many scenes in which most of the retained target lies in one
component.

These results show that connectivity can remove residual structure, but that
component size alone is not a reliable criterion for identifying the target.
A geometry-aware refinement should therefore use spatial structure together
with semantic evidence rather than assume that the largest density component
is the object.

\section{Mask Erosion}
\label{sec:supp_erosion}

\begin{table}[t]
  \centering
  \caption{\textbf{Effect of mask erosion.}
Rows B and C isolate 4\,px supervision-mask erosion using the same full
evaluation masks. Row A shows the corresponding no-erosion result when
evaluation masks are also eroded. Std: captured framing; $6\times$: widened framing.}
  \label{tab:erosion}
  \setlength{\tabcolsep}{4pt}
  \resizebox{\columnwidth}{!}{%
  \begin{tabular}{l l ccc c}
    \toprule
    & Configuration & Std PSNR & Std LPIPS & Zoom PSNR & Leak \\
    \midrule
    A & no erosion, eval on eroded masks & 19.86 & 0.170 & 18.25 & 0.221 \\
    B & no erosion, eval on full masks   & \textbf{19.03} & \textbf{0.188} & \textbf{17.83} & 0.206 \\
    C & 4\,px erosion, eval on full masks & 18.37 & 0.201 & 17.63 & \textbf{0.196} \\
    \bottomrule
  \end{tabular}}
\end{table}

Table~\ref{tab:erosion} examines whether shrinking the supervision masks can
suppress residual structure near object boundaries. Comparing Rows B and C,
4\,px erosion reduces leakage from $0.206$ to $0.196$, but decreases
standard PSNR from $19.03$ to $18.37$\,dB and widened PSNR from
$17.83$ to $17.63$\,dB. The main configuration therefore uses the
uneroded masks, which better preserves target appearance for a small increase
in leakage.

This trade-off is consistent with Gaussian rendering. A rendered silhouette
arises from accumulated opacity rather than a sharp geometric boundary, so
Gaussians whose centers lie near the 2D mask boundary can still contribute
substantially to the target appearance. Erosion removes support from these
boundary primitives first. In contrast, target-adjacent surfaces such as a
ground slab or shelving surface can remain well inside the silhouette from
multiple views and therefore survive even after the mask is eroded. Mask
erosion consequently suppresses some boundary support without directly
addressing the residual structure that is consistently aligned with the target.

\section{Target-Adjacent Radiance and Leakage}
\label{sec:supp_roses}

\begin{figure*}[t]
  \centering
  \setlength{\tabcolsep}{2pt}
  \begin{tabular}{@{}cccc@{}}
    \scriptsize GT & \scriptsize Uncleaned 3DGS &
    \scriptsize GaussianCut & \scriptsize Ours \\
    \includegraphics[width=.235\textwidth]{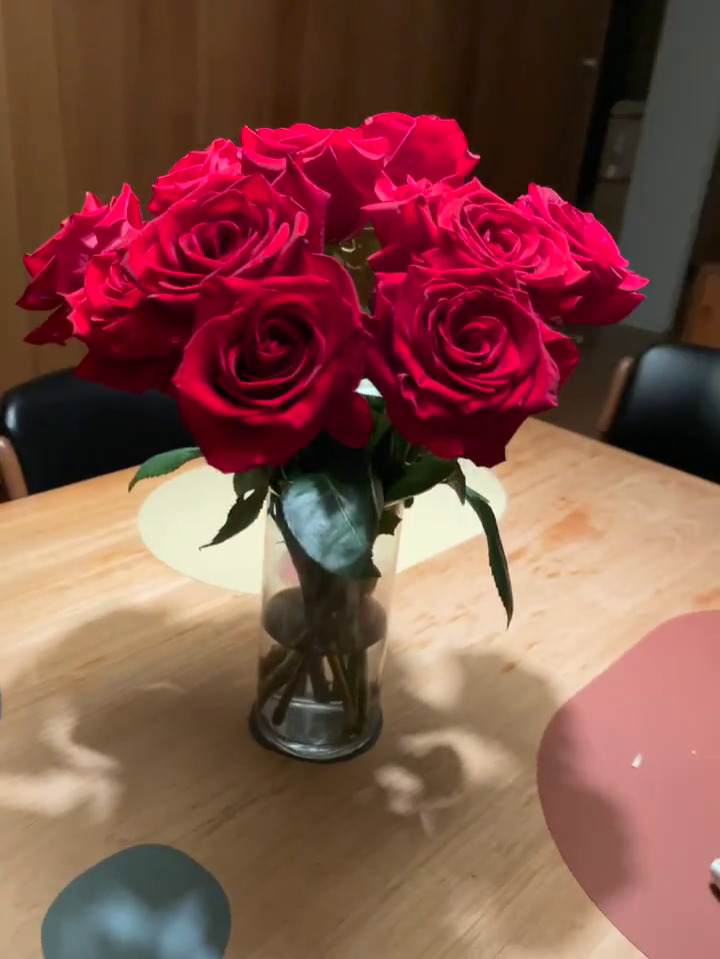} &
    \includegraphics[width=.235\textwidth]{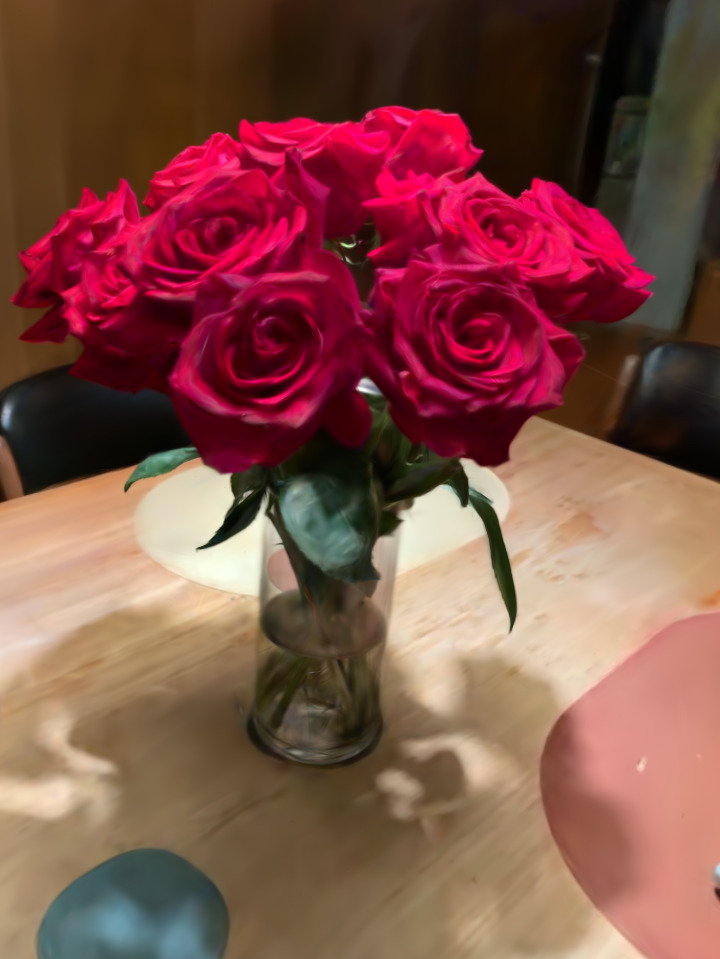} &
    \includegraphics[width=.235\textwidth]{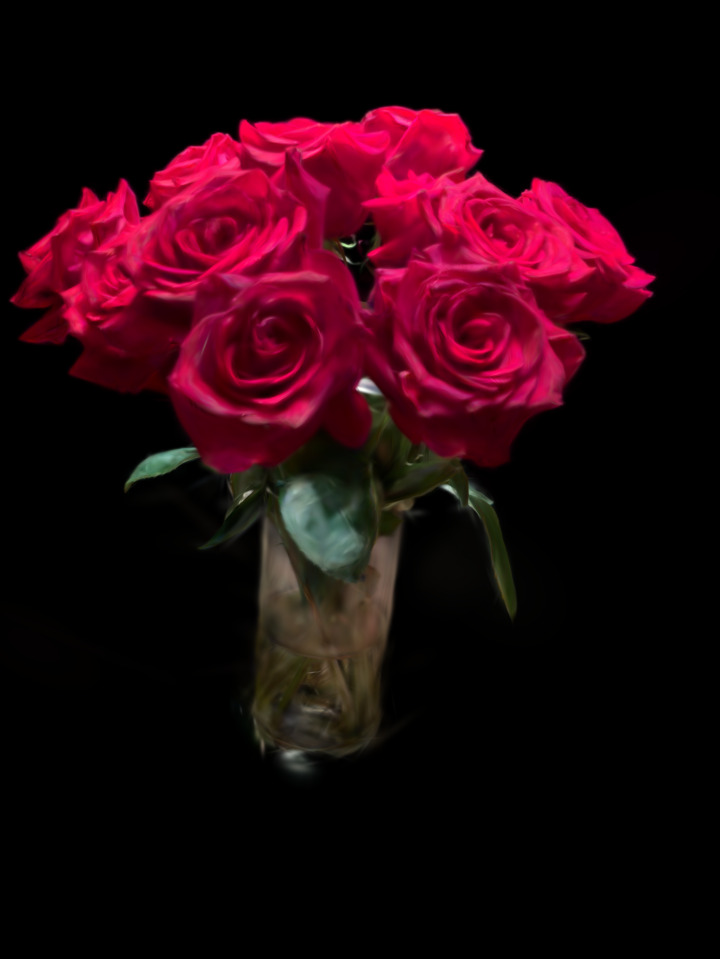} &
    \includegraphics[width=.235\textwidth]{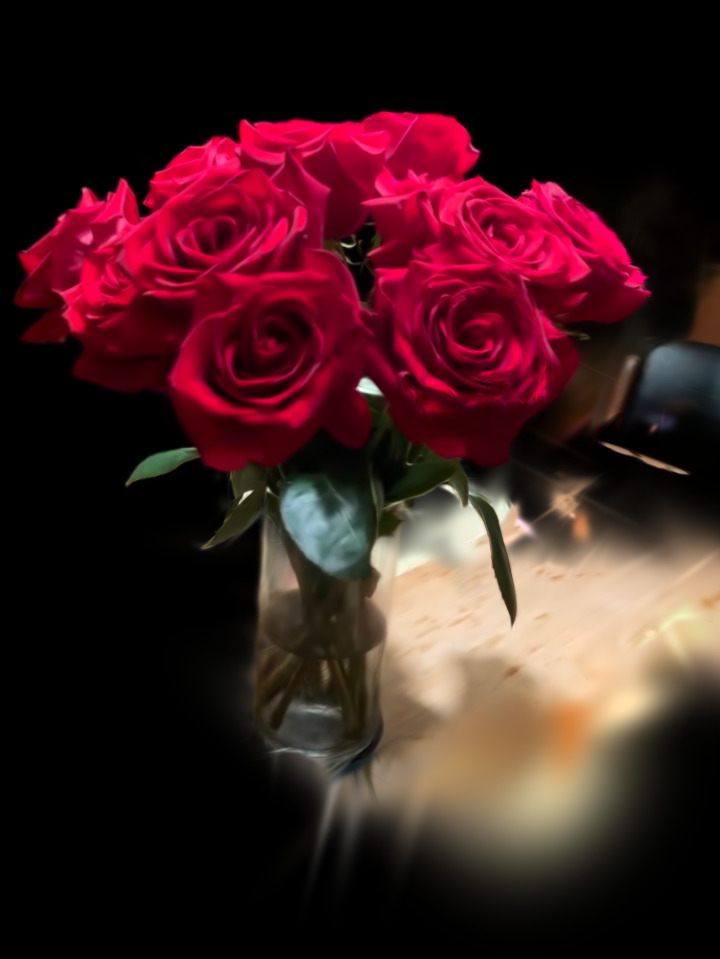} \\
    \includegraphics[width=.235\textwidth]{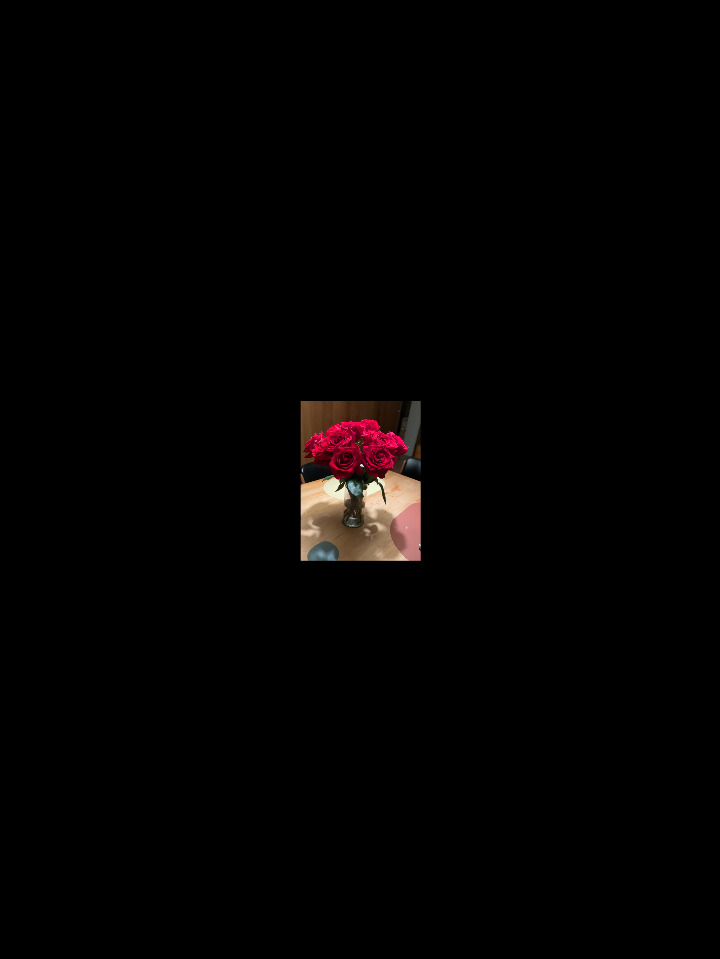} &
    \includegraphics[width=.235\textwidth]{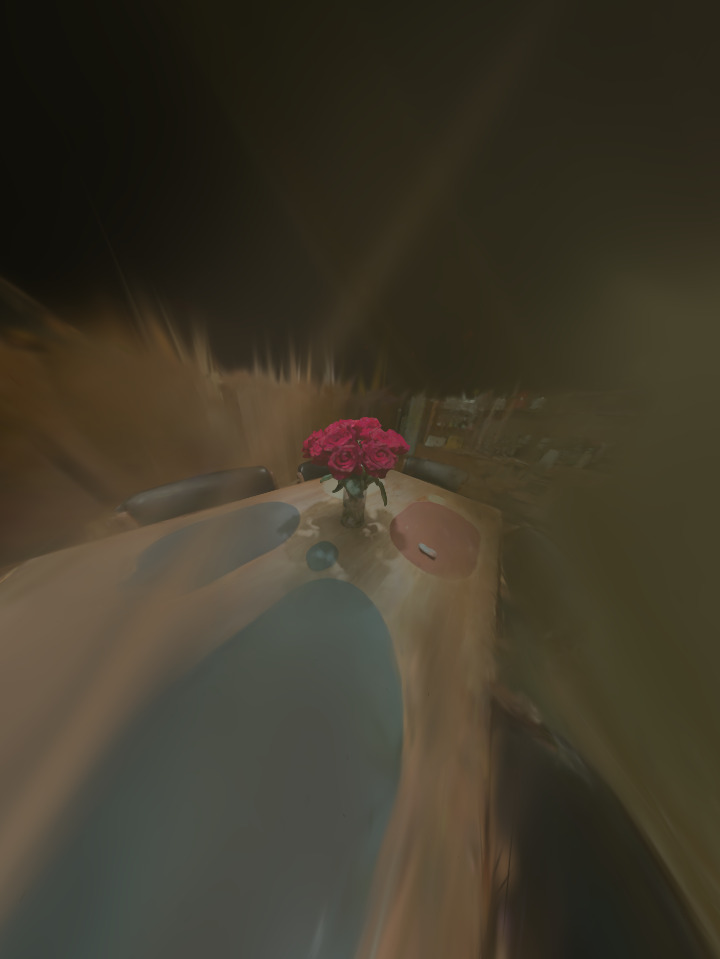} &
    \includegraphics[width=.235\textwidth]{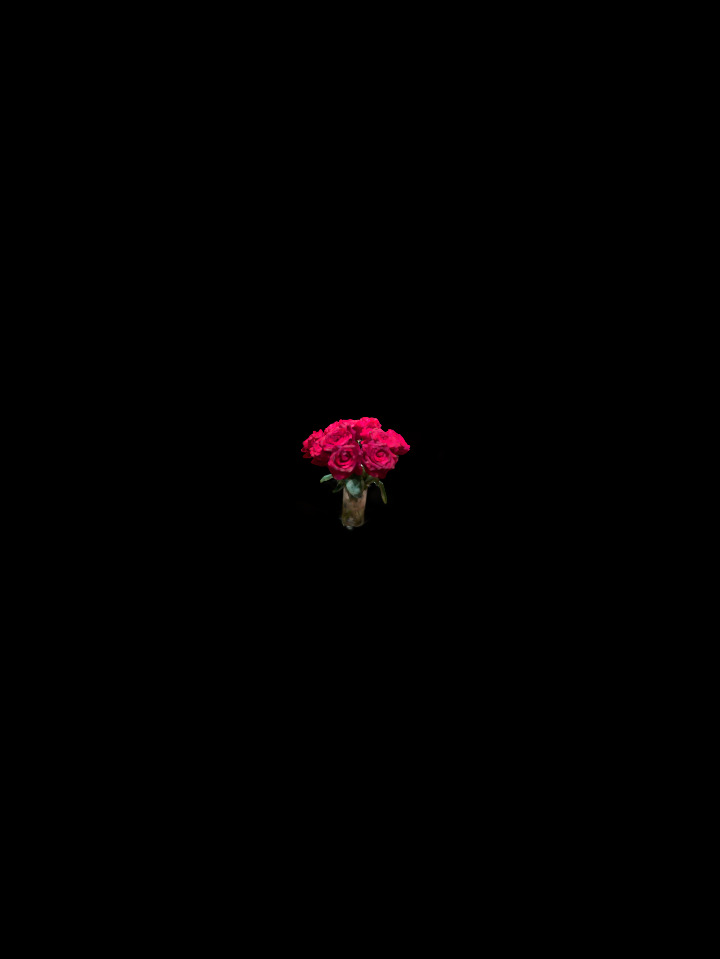} &
    \includegraphics[width=.235\textwidth]{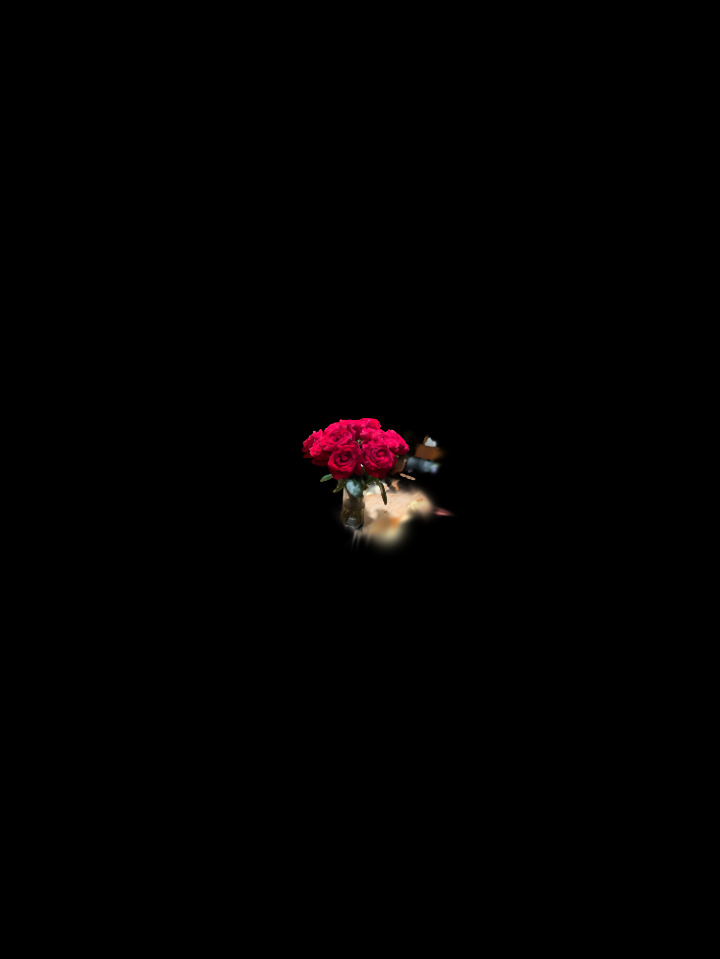}
  \end{tabular}
  \caption{\textbf{Target-adjacent radiance and silhouette leakage.}
  The \emph{roses} scene at captured framing (top) and $6\times$ widened
  framing (bottom). GaussianCut removes the illuminated tabletop beneath the
  bouquet and darkens the glass vase, while \method preserves the vase, water,
  and nearby highlight. \method is $1.61$\,dB better in widened PSNR on the
  displayed frame, although the retained tabletop region lies outside the
  binary target mask and therefore contributes to leakage.}
  \Description{Ground truth, uncleaned, GaussianCut and CleanGS renderings of
  a rose bouquet at ordinary and widened fields of view.}
  \label{fig:supp_roses}
\end{figure*}

Figure~\ref{fig:supp_roses} illustrates the complementary information provided
by leakage and target photometrics. On \emph{roses}, \method achieves higher
PSNR than GaussianCut on 44 of 46 frames, with an average advantage of
$1.13$\,dB. The widened rendering also retains part of the illuminated
tabletop immediately beneath the vase.

Because this region lies outside the bouquet-and-vase mask, it contributes to
silhouette leakage. At the same time, the nearby radiance is visually coupled
to the appearance of the transparent vase and its contact with the supporting
surface; in the displayed frame, its removal coincides with a darker vase and
a $1.61$\,dB lower widened PSNR for GaussianCut. Thus, off-mask opacity can
capture both unwanted scene structure and target-adjacent appearance effects.
Reporting leakage together with photometric fidelity distinguishes these
different cleanup outcomes.

\section{Effect of Additional GaussianCut Supervision}
\label{sec:supp_gcut_supervision}

\begin{figure}[htbp]
  \centering
  \setlength{\tabcolsep}{1pt}
  \newcommand{\pl}[1]{\rotatebox{90}{\makebox[.155\columnwidth][c]{\tiny #1}}}
  \begin{tabular}{@{}c@{\hspace{1pt}}cccc@{}}
    & \scriptsize GCut, 3 masks & \scriptsize GCut, 3 masks (6$\times$)
    & \scriptsize GCut, all masks & \scriptsize GCut, all masks (6$\times$) \\
    \pl{Family}
    & \includegraphics[width=.225\columnwidth]{assets/q_Family_gcut.jpg}
    & \includegraphics[width=.225\columnwidth]{assets/q_Family_gcut_z6.jpg}
    & \includegraphics[width=.225\columnwidth]{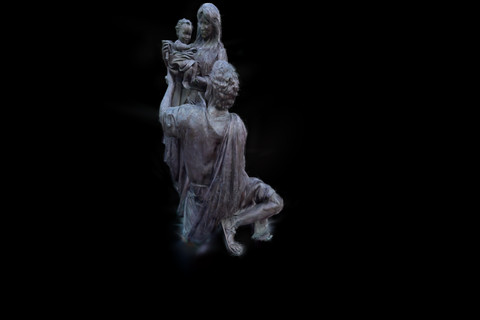}
    & \includegraphics[width=.225\columnwidth]{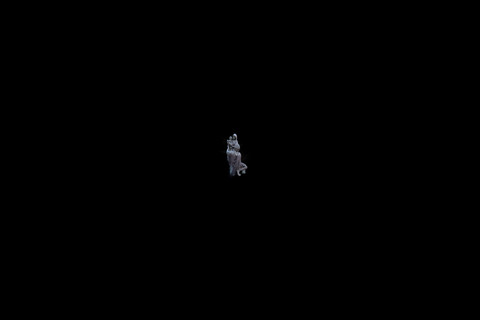} \\
    \pl{courthouses}
    & \includegraphics[width=.225\columnwidth]{assets/q_dl3dv_courthouses_gcut.jpg}
    & \includegraphics[width=.225\columnwidth]{assets/q_dl3dv_courthouses_gcut_z6.jpg}
    & \includegraphics[width=.225\columnwidth]{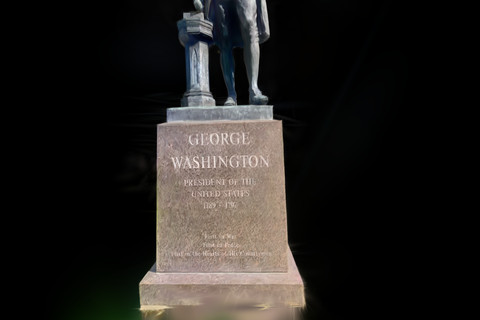}
    & \includegraphics[width=.225\columnwidth]{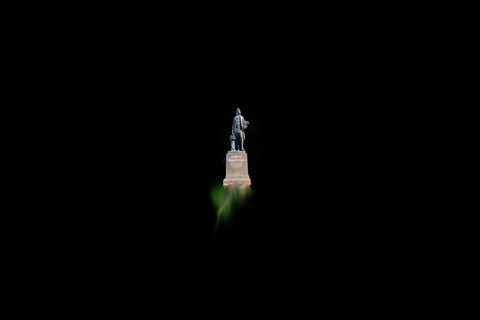} \\
    \pl{garbage}
    & \includegraphics[width=.225\columnwidth]{assets/q_garbage_gcut.jpg}
    & \includegraphics[width=.225\columnwidth]{assets/q_garbage_gcut_z6.jpg}
    & \includegraphics[width=.225\columnwidth]{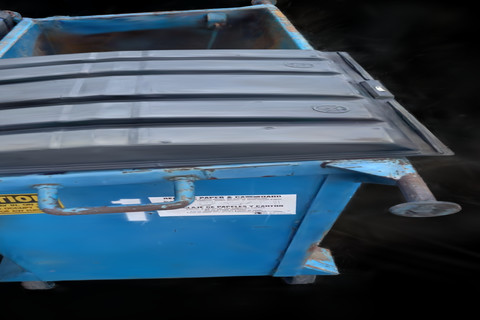}
    & \includegraphics[width=.225\columnwidth]{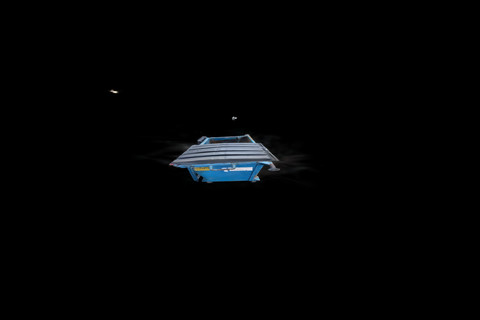} \\
    \pl{fountain}
    & \includegraphics[width=.225\columnwidth]{assets/q_dl3dv_fountain_gcut.jpg}
    & \includegraphics[width=.225\columnwidth]{assets/q_dl3dv_fountain_gcut_z6.jpg}
    & \includegraphics[width=.225\columnwidth]{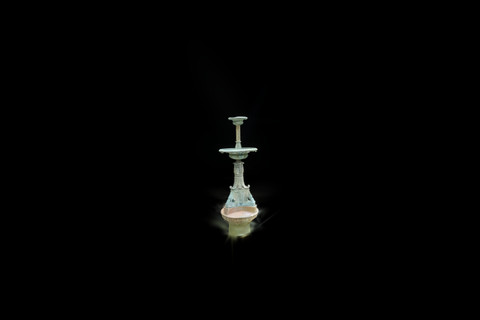}
    & \includegraphics[width=.225\columnwidth]{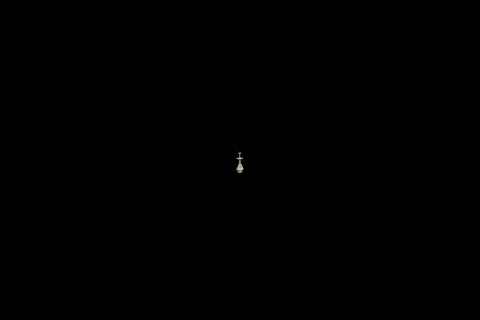}
  \end{tabular}
  \caption{\textbf{Effect of additional GaussianCut supervision.}
  GaussianCut is shown with the same three selected masks used by \method and
  with all available capture masks. Denser supervision removes additional
  residual structure and produces tighter silhouettes across these examples.
  The all-mask setting includes masks at scored cameras and corresponds to the
  daggered GaussianCut reference in the main text.}
  \label{fig:gcut_supervision}
  \Description{Four rows comparing GaussianCut given three masks against
  GaussianCut given every capture mask, at both fields of view.}
\end{figure}

Figure~\ref{fig:gcut_supervision} shows the effect of increasing the supervision
available to GaussianCut. With three selected masks, GaussianCut operates at
the same filtering budget as \method. Supplying all capture masks provides
denser multiview evidence and removes additional residual structure in the
examples shown. We therefore report the three-mask configuration as the
matched-input comparison and the all-mask configuration,
$\text{GaussianCut}^{\dagger}$, as an additional reference with substantially
more supervision.

\section{Qualitative Results for All Scenes}

\begin{figure*}[p]
  \centering
  \setlength{\tabcolsep}{1pt}
  \begin{tabular}{@{}c@{\hspace{1pt}}cccccccc@{}}
    \qhead
      \rotatebox{90}{\makebox[.108\textwidth][c]{\tiny car set2}} &
  \includegraphics[width=.116\textwidth]{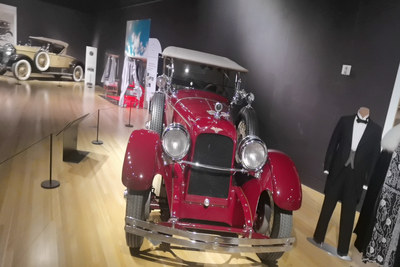} &
  \includegraphics[width=.116\textwidth]{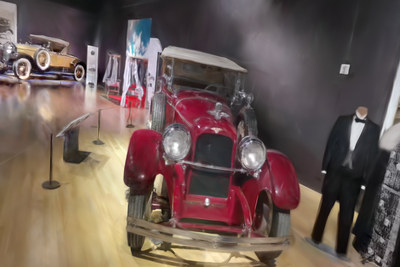} &
  \includegraphics[width=.116\textwidth]{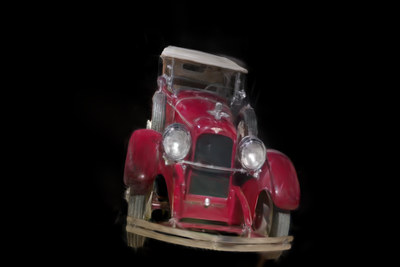} &
  \includegraphics[width=.116\textwidth]{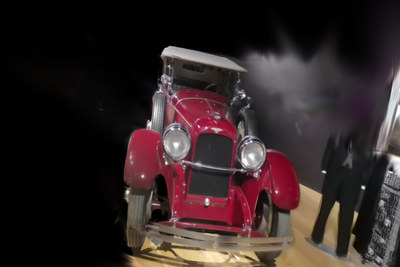} &
  \includegraphics[width=.116\textwidth]{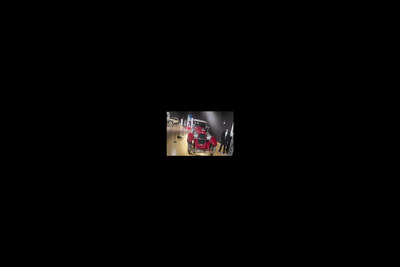} &
  \includegraphics[width=.116\textwidth]{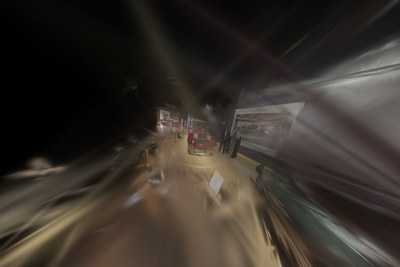} &
  \includegraphics[width=.116\textwidth]{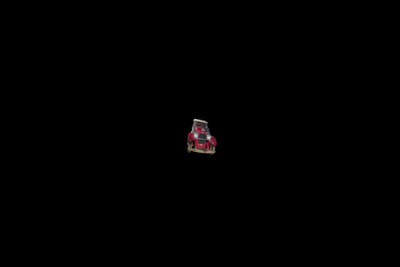} &
  \includegraphics[width=.116\textwidth]{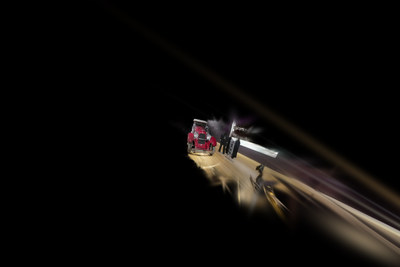} \\
      \rotatebox{90}{\makebox[.108\textwidth][c]{\tiny courthouses}} &
  \includegraphics[width=.116\textwidth]{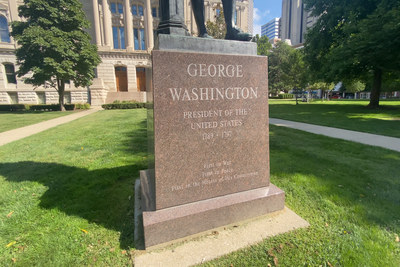} &
  \includegraphics[width=.116\textwidth]{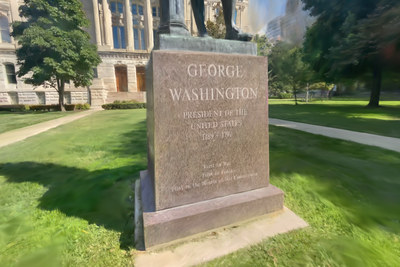} &
  \includegraphics[width=.116\textwidth]{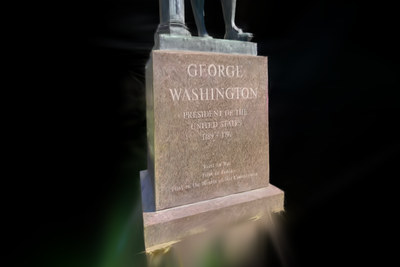} &
  \includegraphics[width=.116\textwidth]{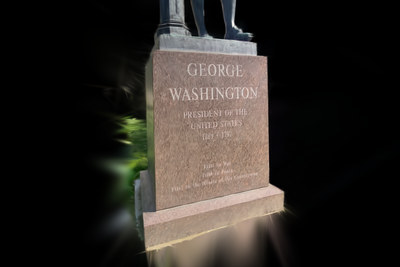} &
  \includegraphics[width=.116\textwidth]{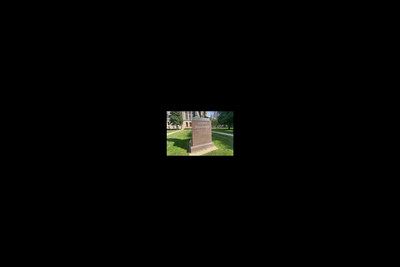} &
  \includegraphics[width=.116\textwidth]{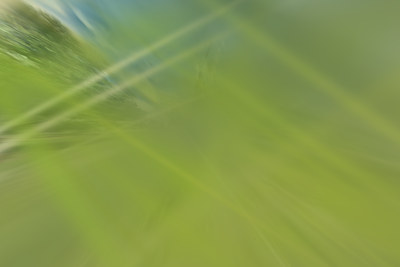} &
  \includegraphics[width=.116\textwidth]{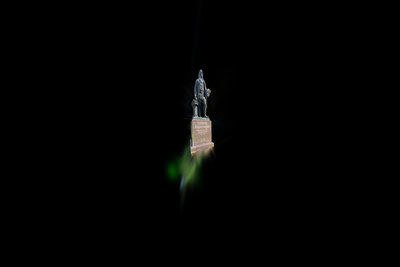} &
  \includegraphics[width=.116\textwidth]{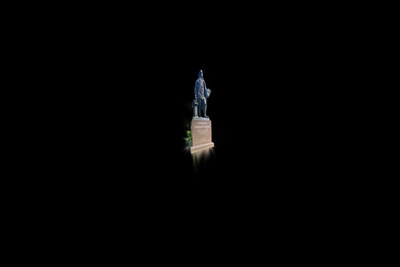} \\
      \rotatebox{90}{\makebox[.108\textwidth][c]{\tiny fountain}} &
  \includegraphics[width=.116\textwidth]{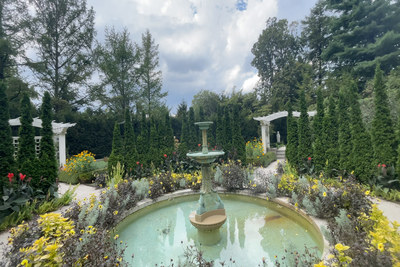} &
  \includegraphics[width=.116\textwidth]{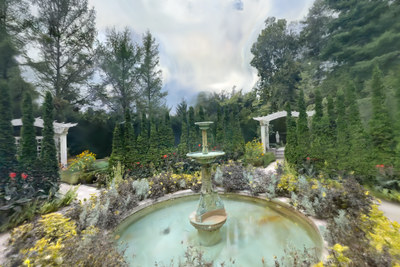} &
  \includegraphics[width=.116\textwidth]{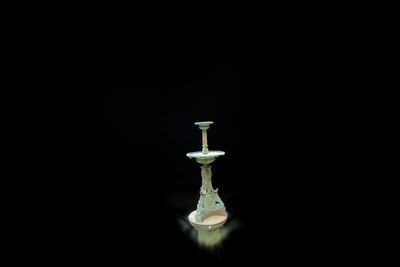} &
  \includegraphics[width=.116\textwidth]{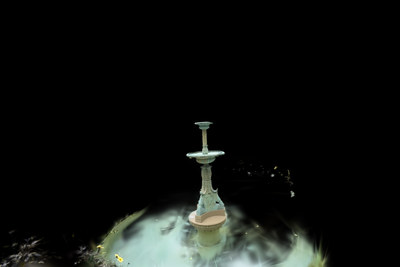} &
  \includegraphics[width=.116\textwidth]{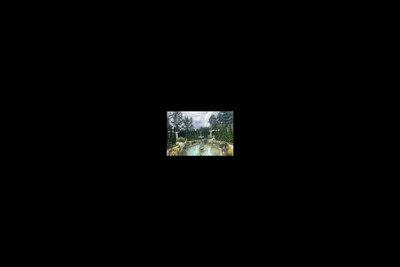} &
  \includegraphics[width=.116\textwidth]{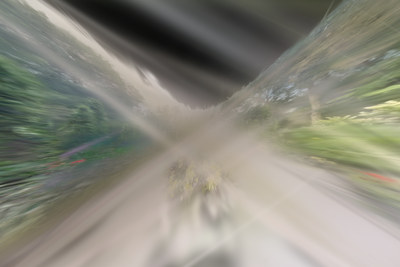} &
  \includegraphics[width=.116\textwidth]{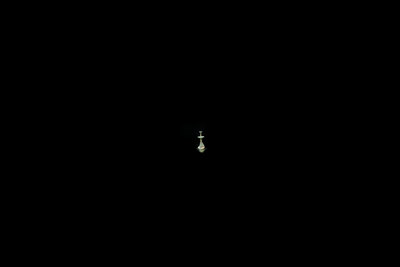} &
  \includegraphics[width=.116\textwidth]{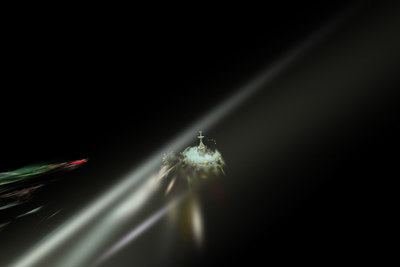} \\
      \rotatebox{90}{\makebox[.108\textwidth][c]{\tiny gardens}} &
  \includegraphics[width=.116\textwidth]{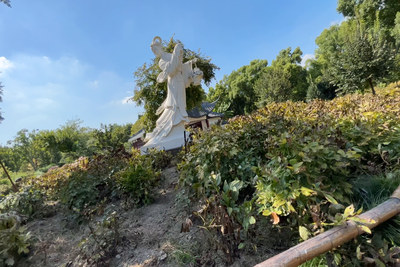} &
  \includegraphics[width=.116\textwidth]{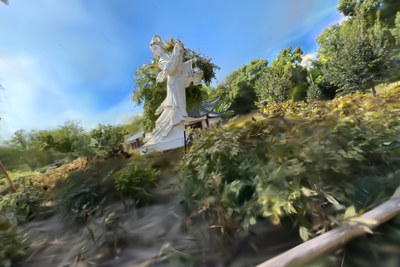} &
  \includegraphics[width=.116\textwidth]{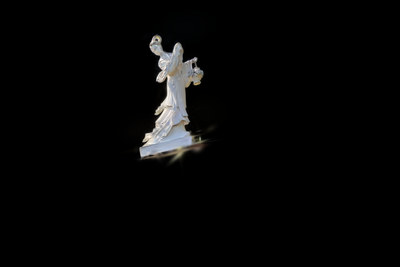} &
  \includegraphics[width=.116\textwidth]{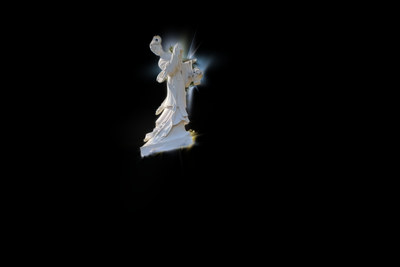} &
  \includegraphics[width=.116\textwidth]{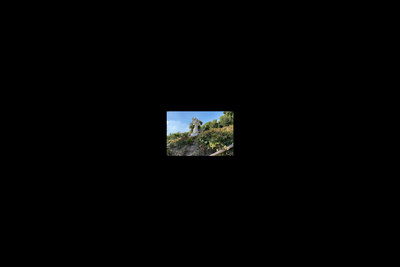} &
  \includegraphics[width=.116\textwidth]{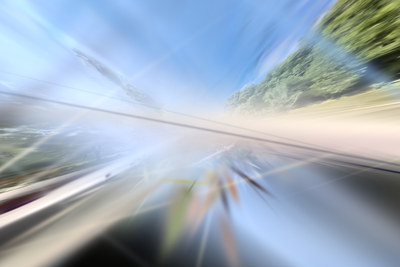} &
  \includegraphics[width=.116\textwidth]{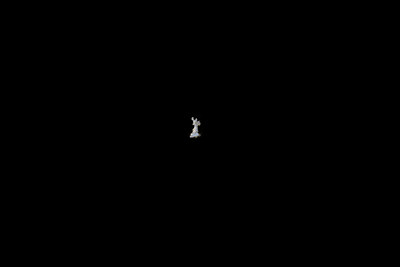} &
  \includegraphics[width=.116\textwidth]{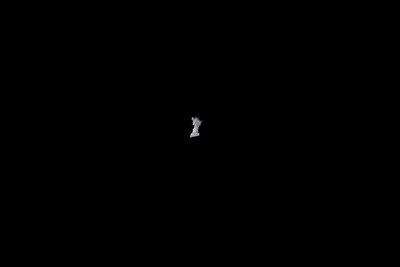} \\
      \rotatebox{90}{\makebox[.108\textwidth][c]{\tiny red brick build}} &
  \includegraphics[width=.116\textwidth]{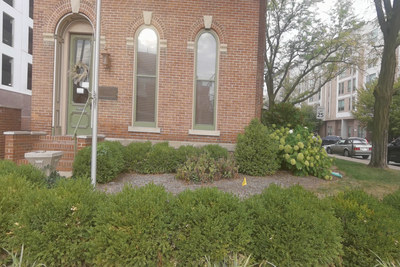} &
  \includegraphics[width=.116\textwidth]{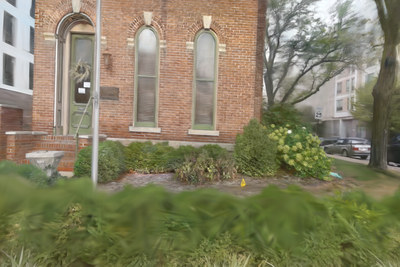} &
  \includegraphics[width=.116\textwidth]{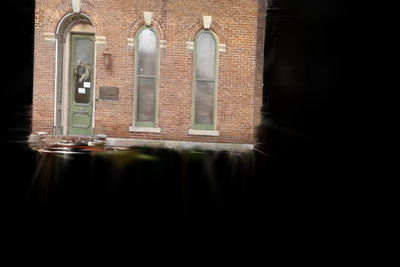} &
  \includegraphics[width=.116\textwidth]{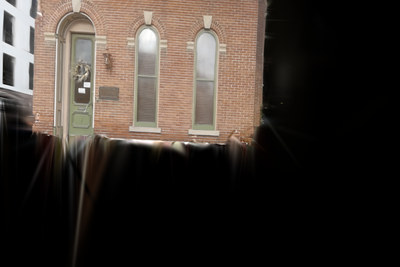} &
  \includegraphics[width=.116\textwidth]{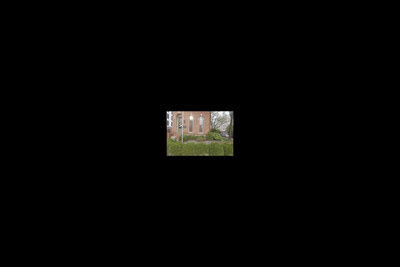} &
  \includegraphics[width=.116\textwidth]{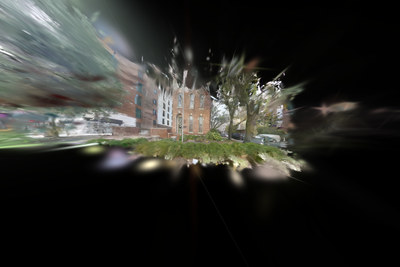} &
  \includegraphics[width=.116\textwidth]{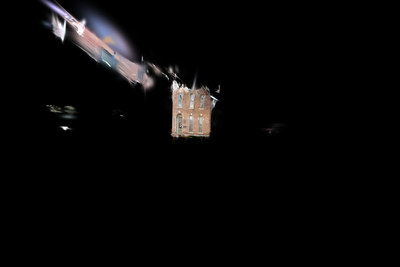} &
  \includegraphics[width=.116\textwidth]{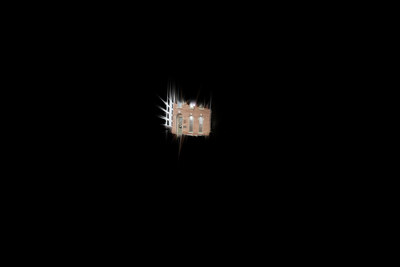} \\
      \rotatebox{90}{\makebox[.108\textwidth][c]{\tiny redcar}} &
  \includegraphics[width=.116\textwidth]{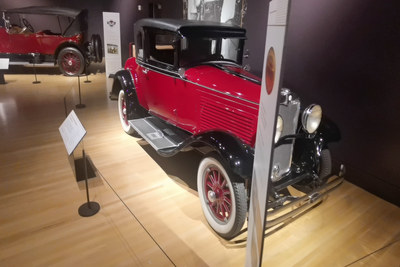} &
  \includegraphics[width=.116\textwidth]{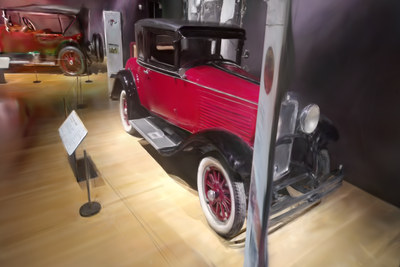} &
  \includegraphics[width=.116\textwidth]{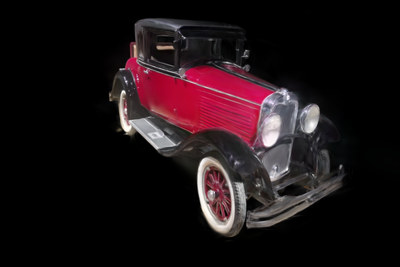} &
  \includegraphics[width=.116\textwidth]{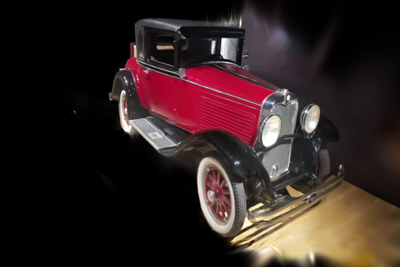} &
  \includegraphics[width=.116\textwidth]{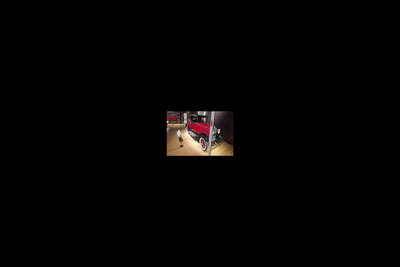} &
  \includegraphics[width=.116\textwidth]{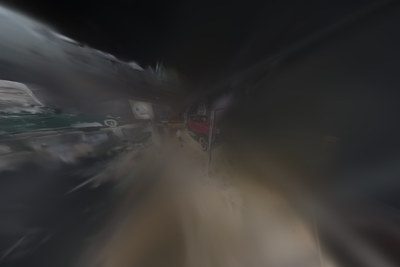} &
  \includegraphics[width=.116\textwidth]{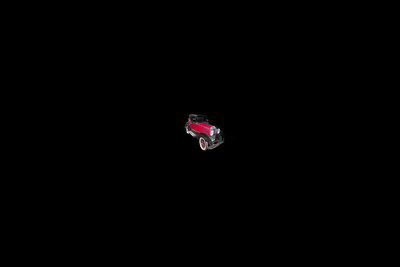} &
  \includegraphics[width=.116\textwidth]{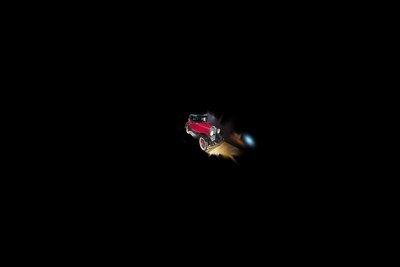} \\
      \rotatebox{90}{\makebox[.108\textwidth][c]{\tiny super markets}} &
  \includegraphics[width=.116\textwidth]{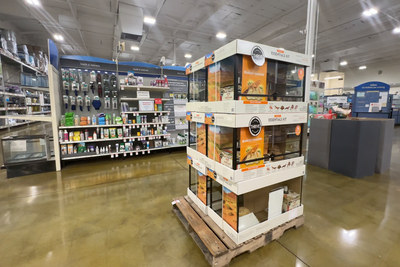} &
  \includegraphics[width=.116\textwidth]{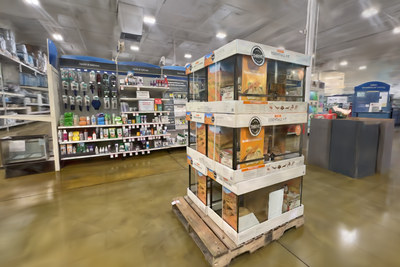} &
  \includegraphics[width=.116\textwidth]{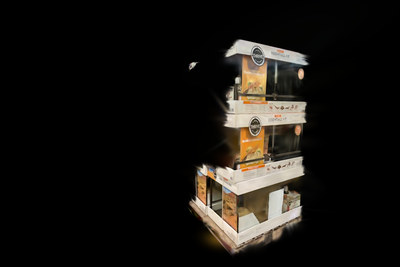} &
  \includegraphics[width=.116\textwidth]{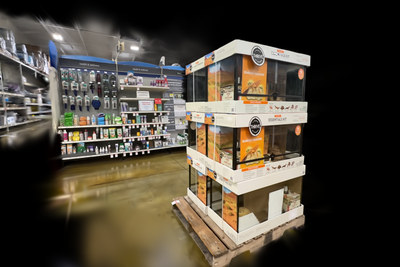} &
  \includegraphics[width=.116\textwidth]{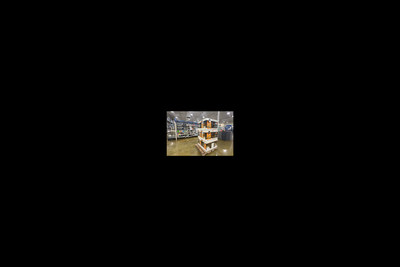} &
  \includegraphics[width=.116\textwidth]{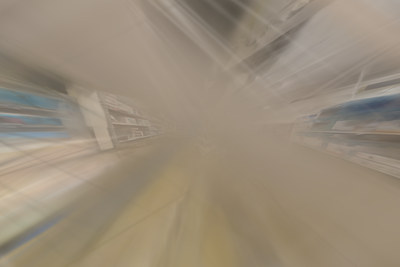} &
  \includegraphics[width=.116\textwidth]{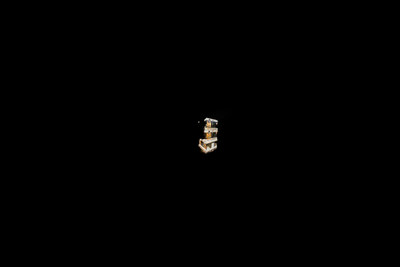} &
  \includegraphics[width=.116\textwidth]{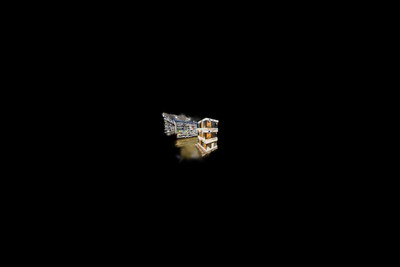} \\
      \rotatebox{90}{\makebox[.108\textwidth][c]{\tiny tractor}} &
  \includegraphics[width=.116\textwidth]{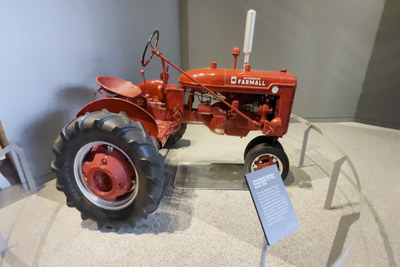} &
  \includegraphics[width=.116\textwidth]{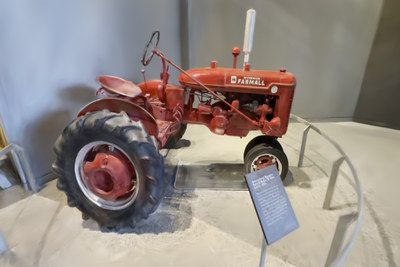} &
  \includegraphics[width=.116\textwidth]{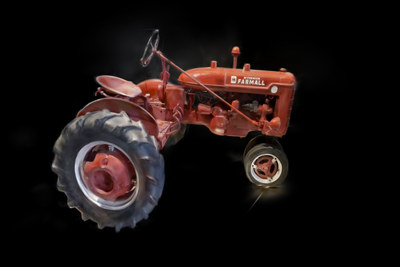} &
  \includegraphics[width=.116\textwidth]{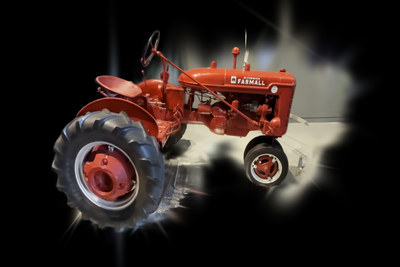} &
  \includegraphics[width=.116\textwidth]{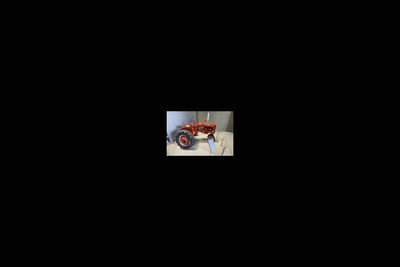} &
  \includegraphics[width=.116\textwidth]{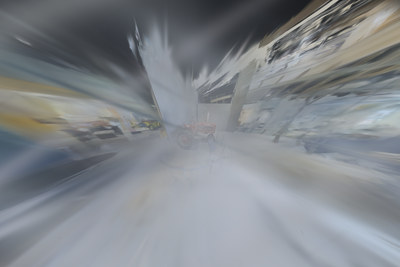} &
  \includegraphics[width=.116\textwidth]{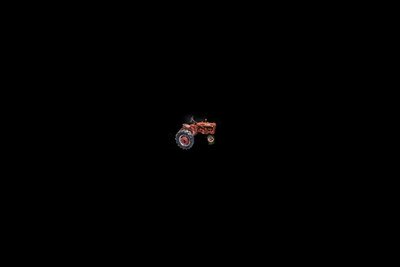} &
  \includegraphics[width=.116\textwidth]{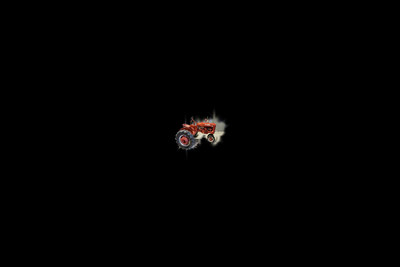} \\
      \rotatebox{90}{\makebox[.108\textwidth][c]{\tiny car}} &
  \includegraphics[width=.116\textwidth]{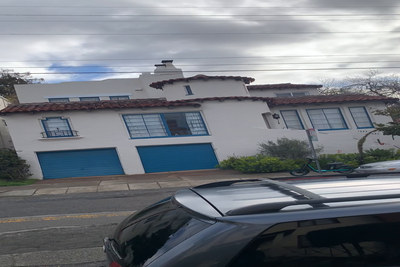} &
  \includegraphics[width=.116\textwidth]{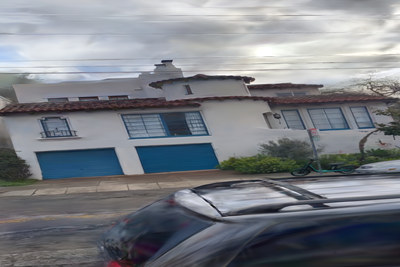} &
  \includegraphics[width=.116\textwidth]{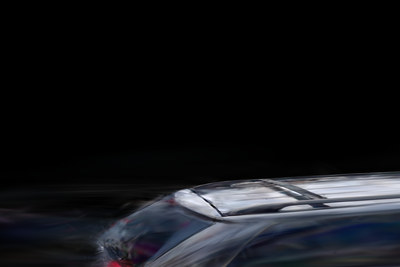} &
  \includegraphics[width=.116\textwidth]{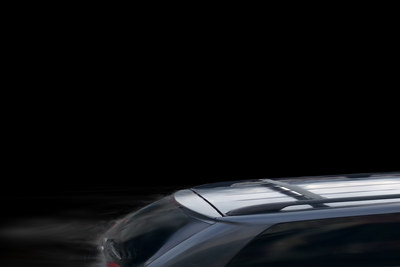} &
  \includegraphics[width=.116\textwidth]{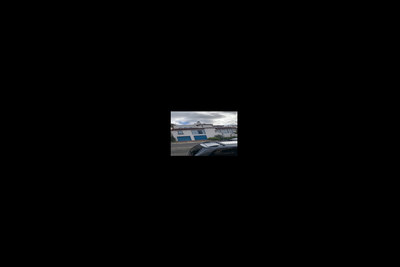} &
  \includegraphics[width=.116\textwidth]{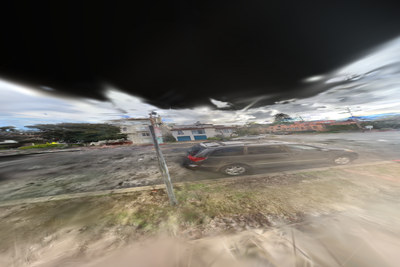} &
  \includegraphics[width=.116\textwidth]{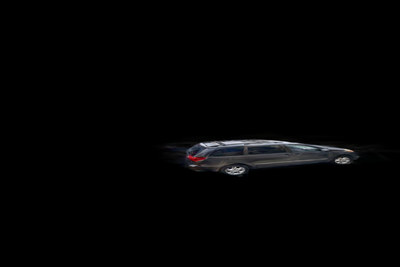} &
  \includegraphics[width=.116\textwidth]{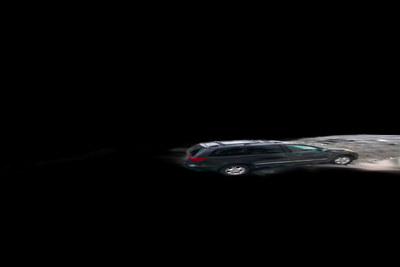}
  \end{tabular}
  \caption{\textbf{Qualitative results across DL3DV and NeRFBusters scenes.}
  All eight DL3DV scenes and the NeRFBusters \emph{car} scene are shown at
  captured and $6\times$ widened framing. Widening preserves camera pose and output
  resolution while reducing focal length, exposing additional content retained
  in each exported asset. GaussianCut is shown using its all-mask
  $\text{GaussianCut}^{\dagger}$ configuration.}
  \Description{A nine-row grid of renderings for nine scenes, with four
  columns at captured framing and four at widened field of view.}
  \label{fig:supp_qual_a}
\end{figure*}

\begin{figure*}[p]
  \centering
  \setlength{\tabcolsep}{1pt}
  \begin{tabular}{@{}c@{\hspace{1pt}}cccccccc@{}}
    \qhead
      \rotatebox{90}{\makebox[.108\textwidth][c]{\tiny century}} &
  \includegraphics[width=.116\textwidth]{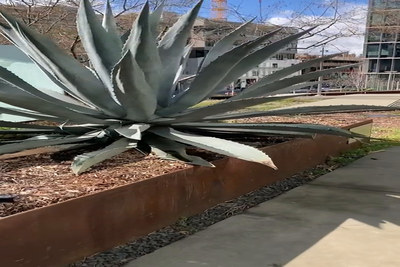} &
  \includegraphics[width=.116\textwidth]{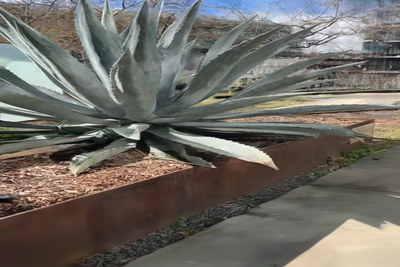} &
  \includegraphics[width=.116\textwidth]{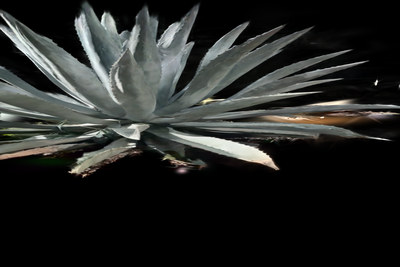} &
  \includegraphics[width=.116\textwidth]{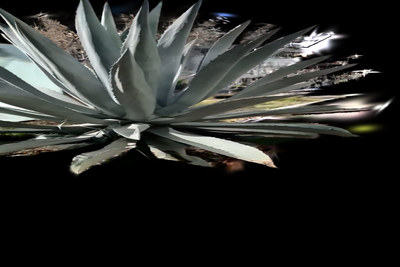} &
  \includegraphics[width=.116\textwidth]{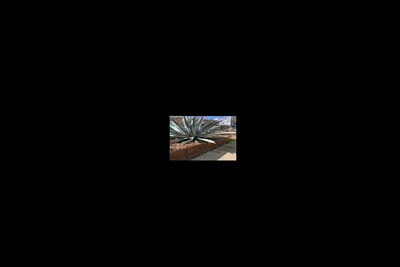} &
  \includegraphics[width=.116\textwidth]{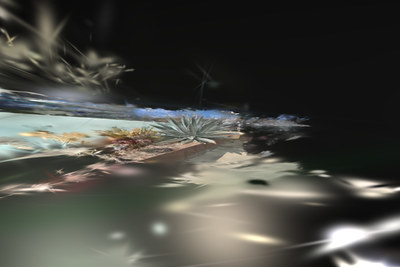} &
  \includegraphics[width=.116\textwidth]{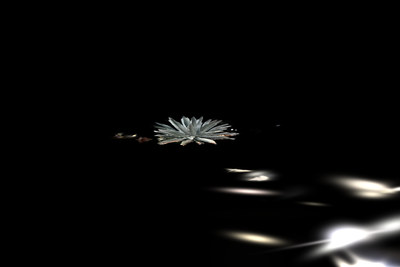} &
  \includegraphics[width=.116\textwidth]{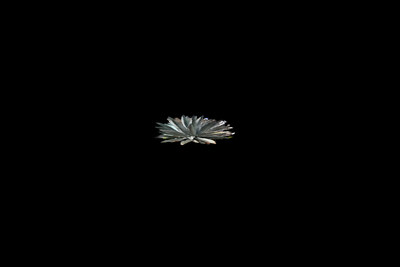} \\
      \rotatebox{90}{\makebox[.108\textwidth][c]{\tiny flowers}} &
  \includegraphics[width=.116\textwidth]{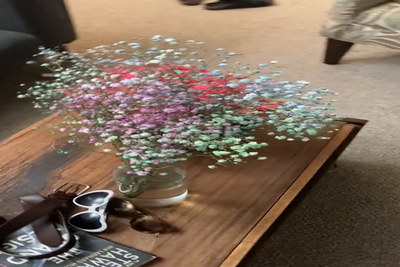} &
  \includegraphics[width=.116\textwidth]{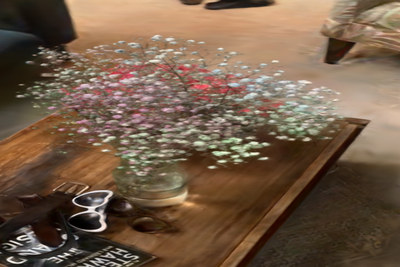} &
  \includegraphics[width=.116\textwidth]{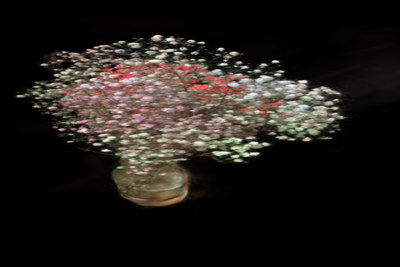} &
  \includegraphics[width=.116\textwidth]{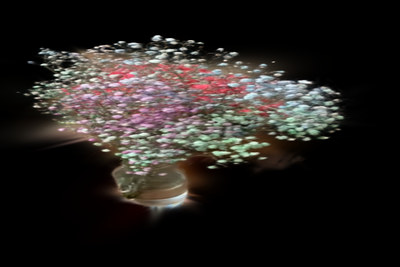} &
  \includegraphics[width=.116\textwidth]{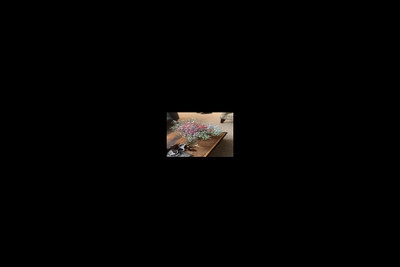} &
  \includegraphics[width=.116\textwidth]{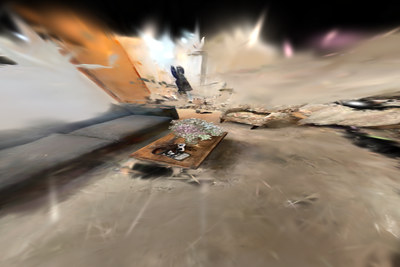} &
  \includegraphics[width=.116\textwidth]{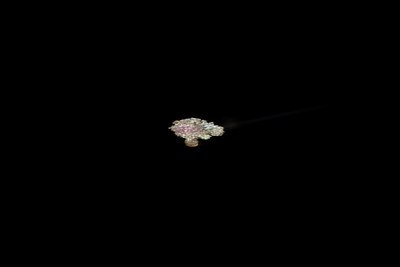} &
  \includegraphics[width=.116\textwidth]{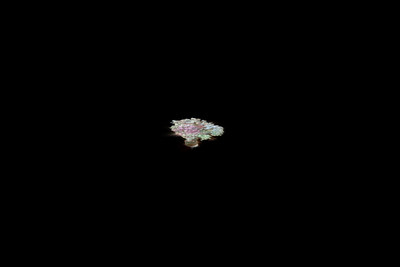} \\
      \rotatebox{90}{\makebox[.108\textwidth][c]{\tiny garbage}} &
  \includegraphics[width=.116\textwidth]{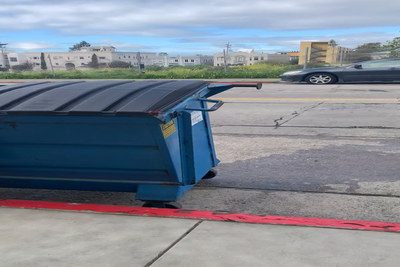} &
  \includegraphics[width=.116\textwidth]{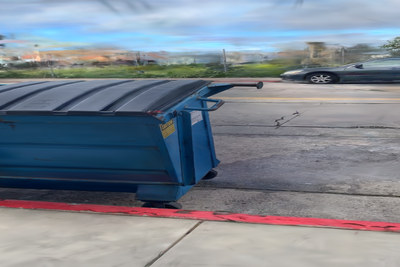} &
  \includegraphics[width=.116\textwidth]{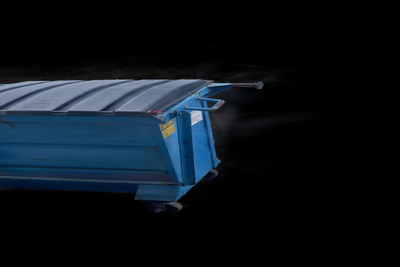} &
  \includegraphics[width=.116\textwidth]{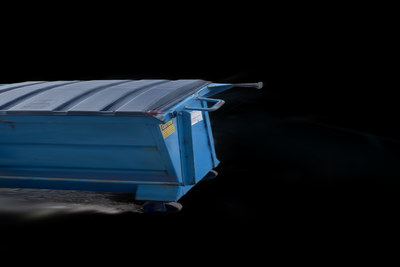} &
  \includegraphics[width=.116\textwidth]{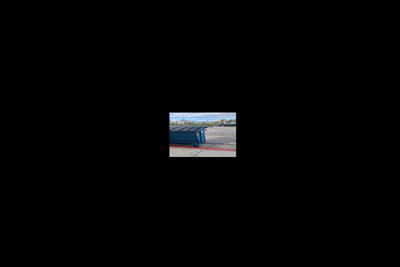} &
  \includegraphics[width=.116\textwidth]{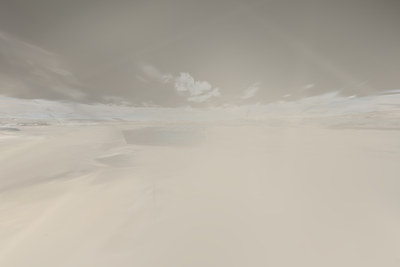} &
  \includegraphics[width=.116\textwidth]{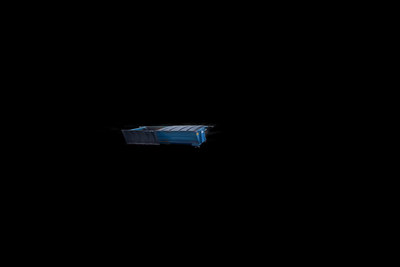} &
  \includegraphics[width=.116\textwidth]{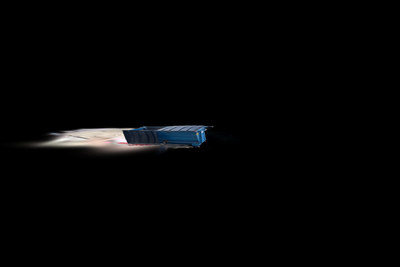} \\
      \rotatebox{90}{\makebox[.108\textwidth][c]{\tiny picnic}} &
  \includegraphics[width=.116\textwidth]{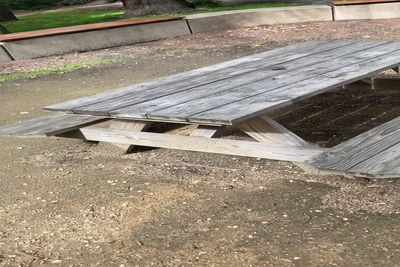} &
  \includegraphics[width=.116\textwidth]{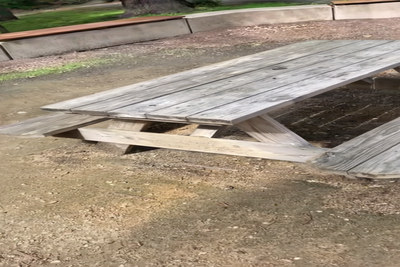} &
  \includegraphics[width=.116\textwidth]{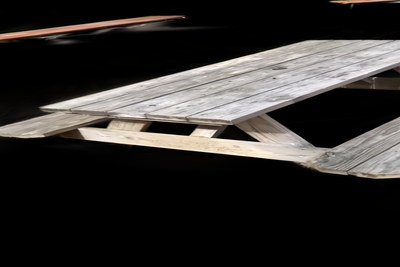} &
  \includegraphics[width=.116\textwidth]{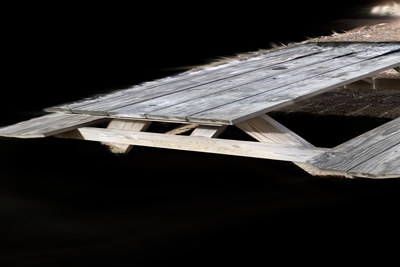} &
  \includegraphics[width=.116\textwidth]{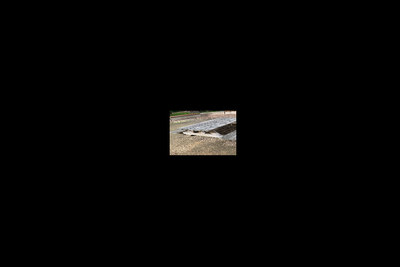} &
  \includegraphics[width=.116\textwidth]{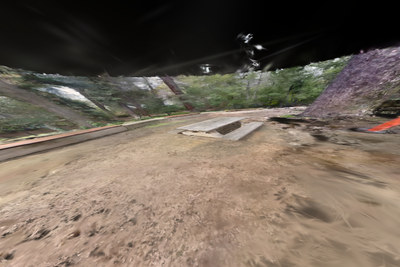} &
  \includegraphics[width=.116\textwidth]{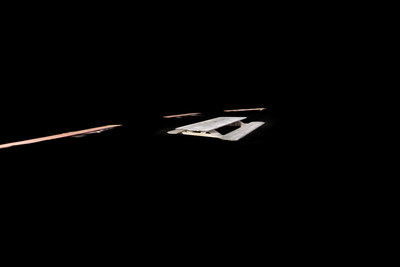} &
  \includegraphics[width=.116\textwidth]{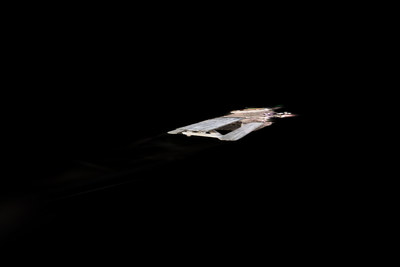} \\
      \rotatebox{90}{\makebox[.108\textwidth][c]{\tiny roses}} &
  \includegraphics[width=.116\textwidth]{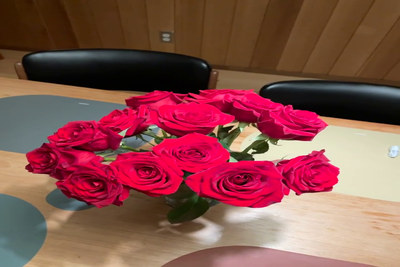} &
  \includegraphics[width=.116\textwidth]{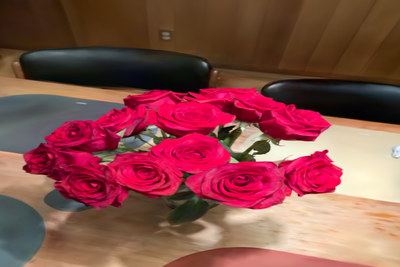} &
  \includegraphics[width=.116\textwidth]{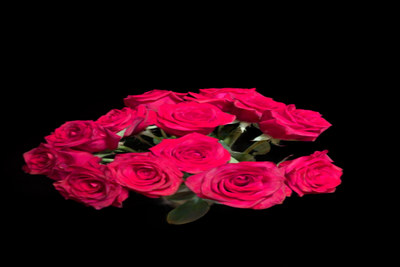} &
  \includegraphics[width=.116\textwidth]{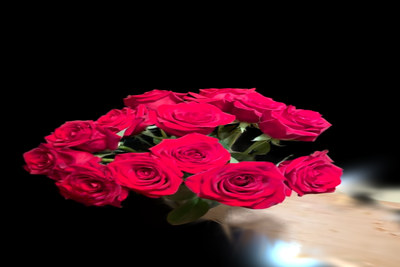} &
  \includegraphics[width=.116\textwidth]{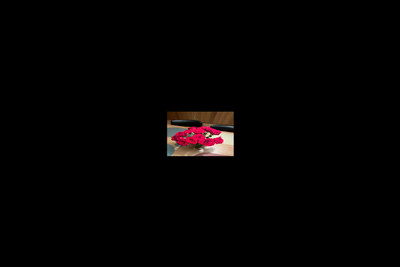} &
  \includegraphics[width=.116\textwidth]{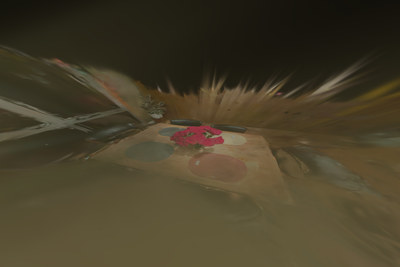} &
  \includegraphics[width=.116\textwidth]{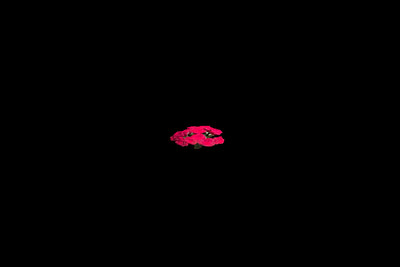} &
  \includegraphics[width=.116\textwidth]{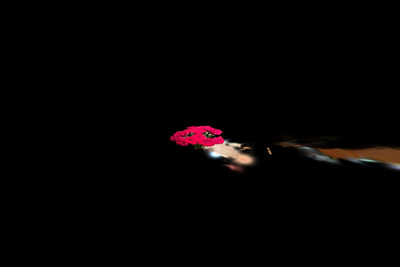} \\
      \rotatebox{90}{\makebox[.108\textwidth][c]{\tiny table}} &
  \includegraphics[width=.116\textwidth]{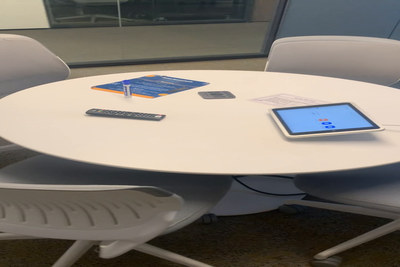} &
  \includegraphics[width=.116\textwidth]{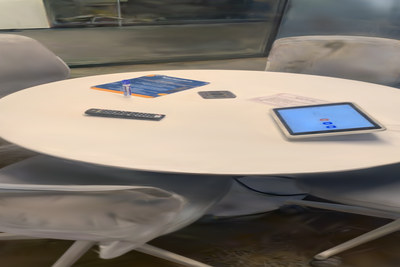} &
  \includegraphics[width=.116\textwidth]{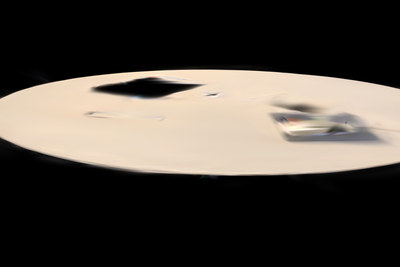} &
  \includegraphics[width=.116\textwidth]{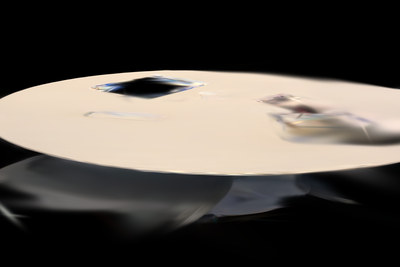} &
  \includegraphics[width=.116\textwidth]{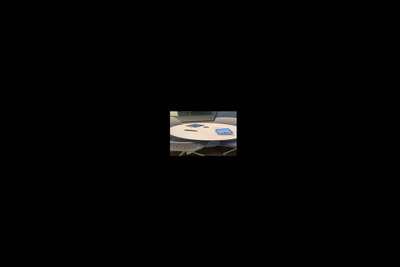} &
  \includegraphics[width=.116\textwidth]{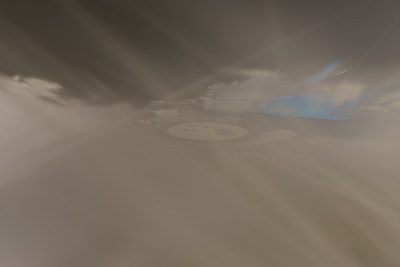} &
  \includegraphics[width=.116\textwidth]{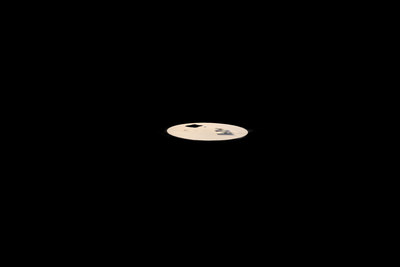} &
  \includegraphics[width=.116\textwidth]{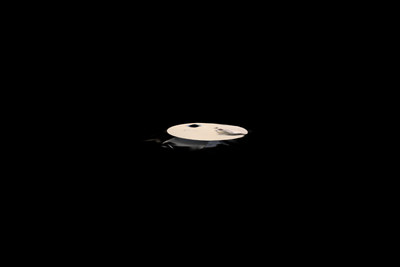} \\
      \rotatebox{90}{\makebox[.108\textwidth][c]{\tiny Family}} &
  \includegraphics[width=.116\textwidth]{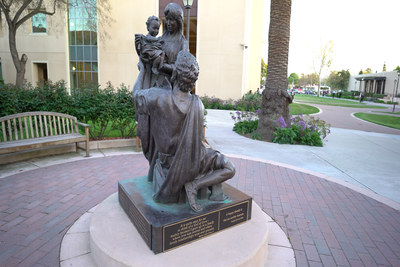} &
  \includegraphics[width=.116\textwidth]{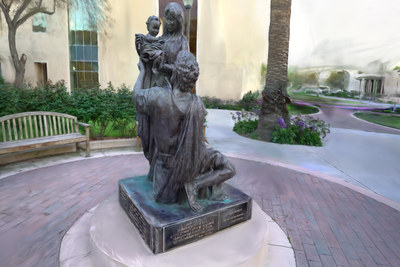} &
  \includegraphics[width=.116\textwidth]{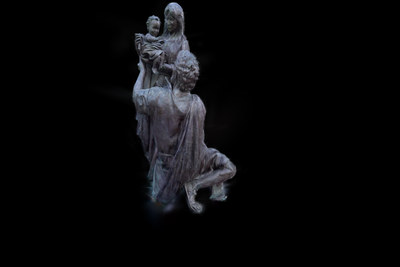} &
  \includegraphics[width=.116\textwidth]{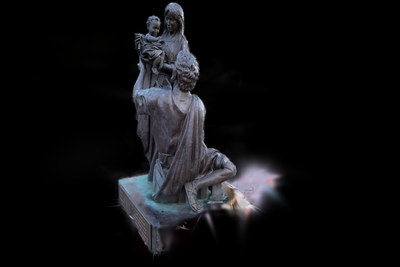} &
  \includegraphics[width=.116\textwidth]{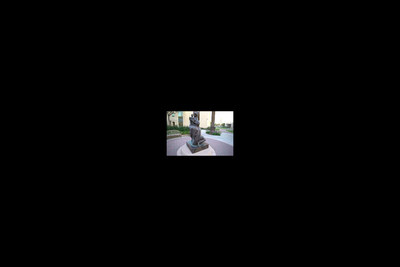} &
  \includegraphics[width=.116\textwidth]{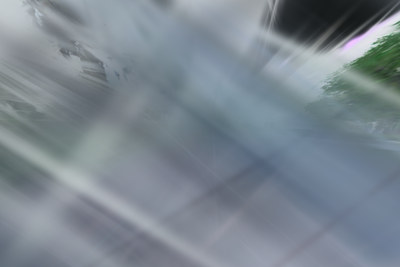} &
  \includegraphics[width=.116\textwidth]{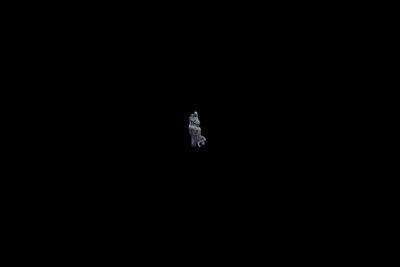} &
  \includegraphics[width=.116\textwidth]{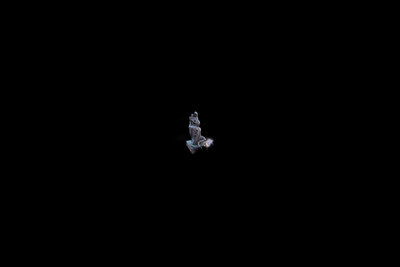} \\
      \rotatebox{90}{\makebox[.108\textwidth][c]{\tiny Francis}} &
  \includegraphics[width=.116\textwidth]{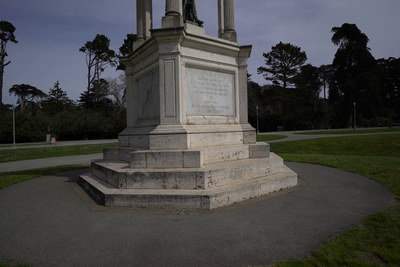} &
  \includegraphics[width=.116\textwidth]{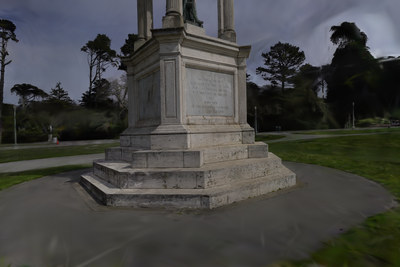} &
  \includegraphics[width=.116\textwidth]{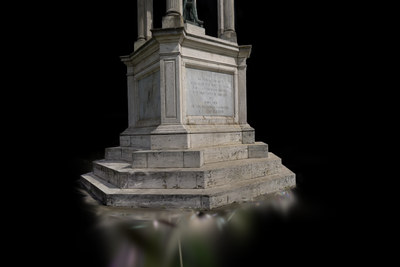} &
  \includegraphics[width=.116\textwidth]{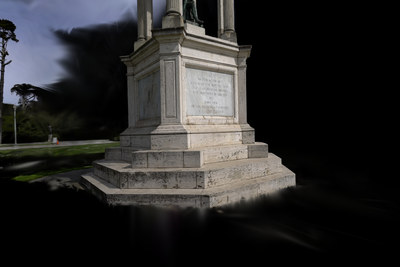} &
  \includegraphics[width=.116\textwidth]{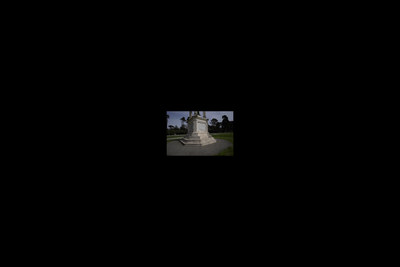} &
  \includegraphics[width=.116\textwidth]{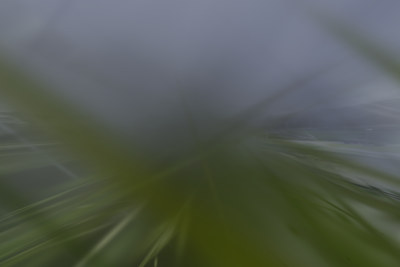} &
  \includegraphics[width=.116\textwidth]{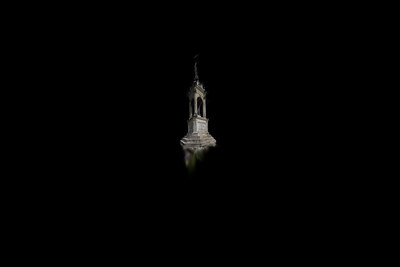} &
  \includegraphics[width=.116\textwidth]{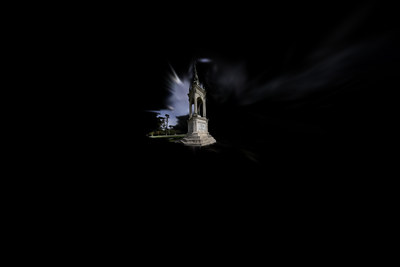} \\
      \rotatebox{90}{\makebox[.108\textwidth][c]{\tiny Horse}} &
  \includegraphics[width=.116\textwidth]{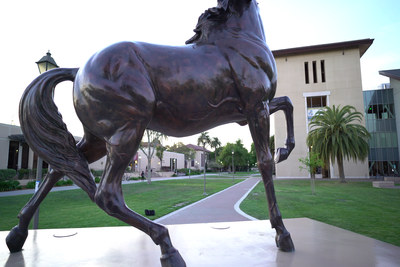} &
  \includegraphics[width=.116\textwidth]{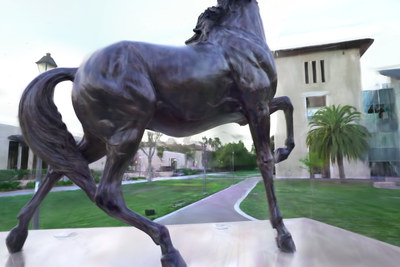} &
  \includegraphics[width=.116\textwidth]{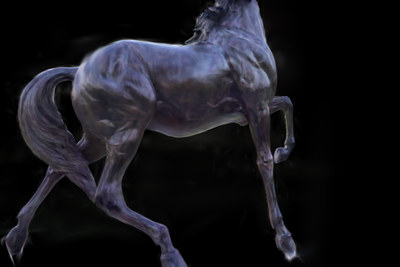} &
  \includegraphics[width=.116\textwidth]{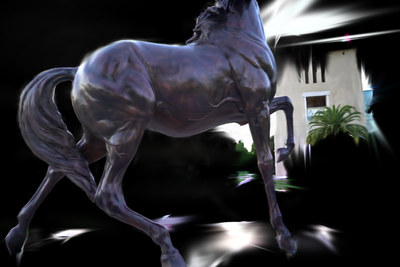} &
  \includegraphics[width=.116\textwidth]{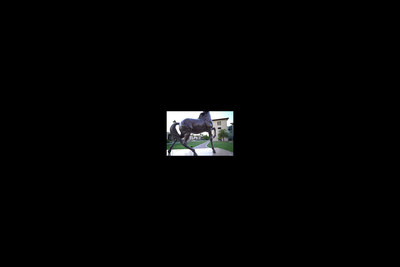} &
  \includegraphics[width=.116\textwidth]{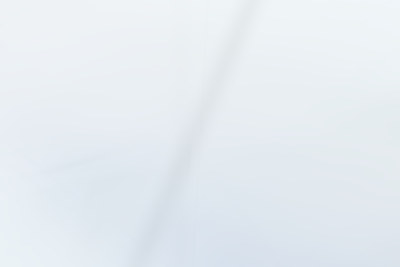} &
  \includegraphics[width=.116\textwidth]{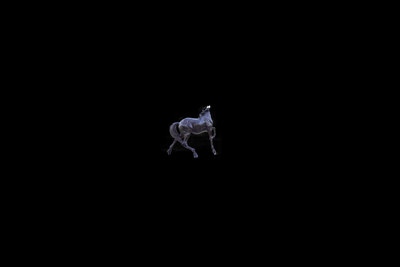} &
  \includegraphics[width=.116\textwidth]{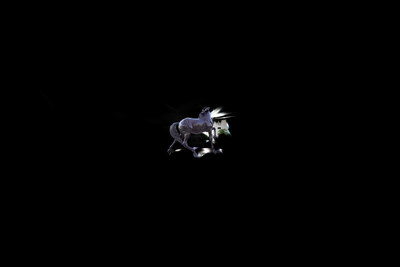}
  \end{tabular}
  \caption{\textbf{Qualitative results across NeRFBusters and Tanks and
  Temples scenes.}
  Six additional NeRFBusters scenes and all three Tanks and Temples scenes
  are shown at captured and $6\times$ widened framing. The widened views
  preserve camera pose and output resolution while exposing Gaussian content
  retained outside the original framing. GaussianCut is shown using its
  all-mask $\text{GaussianCut}^{\dagger}$ configuration.}
  \Description{A nine-row grid of renderings for nine scenes, with four
  columns at captured framing and four at widened field of view.}
  \label{fig:supp_qual_b}
\end{figure*}

Figures~\ref{fig:supp_qual_a} and~\ref{fig:supp_qual_b} show one evaluation
frame for every scene in the 18-scene benchmark at both captured framing and
$6\times$ widened field of view. Together, they provide the scene-level
counterpart to the aggregate photometric and cleanup results, illustrating
both target preservation and the residual structure exposed by reframing.

The widened reference is obtained by applying the same intrinsics
transformation used for evaluation to the captured image. Because reference
radiance is available only within the original captured field of view, the
transformed ground truth occupies the corresponding central region of the
widened image. Camera pose and output resolution remain fixed throughout.

\clearpage

\end{document}